\documentclass[runningheads]{llncs}
\usepackage{subfigure}
\usepackage{xcolor}

\usepackage[T1]{fontenc}
\usepackage{amsmath,amssymb,amsfonts}
\usepackage{hyperref}
\usepackage{graphicx}
\begin{document}
\title{Swin Transformer for Robust CGI Images Detection: Intra- and Inter-Dataset Analysis across Multiple Color Spaces}
%
%
\author{Preeti Mehta\inst{1}\orcidID{0000-0002-6153-2145} \and
Aman Sagar\inst{2} \and
Suchi Kumari\inst{2}}
\authorrunning{M. Preeti et al.}
%
\institute{Computer Science and Engineering, IILM University, Gurugram, India\\
\email{preeti.mehta@iilm.edu}\\
\and
Shiv Nadar Institute of Eminence, Delhi-NCR, India\\
\email{suchi.singh24@gmail.com}}
\maketitle              
\begin{abstract}
\textbf{Purpose}
This study aims to address the growing challenge of distinguishing computer-generated imagery (CGI) from authentic digital images across three different color spaces; RGB, YCbCr, and HSV. Given the limitations of existing classification methods in handling the complexity and variability of CGI, this research proposes a Swin Transformer based model for accurate differentiation between natural and synthetic images.

\textbf{Methods}
The proposed model leverages the Swin Transformer's hierarchical architecture to capture local and global features for distinguishing CGI from natural images. Its performance was assessed through intra- and inter-dataset testing across three datasets: CiFAKE, JSSSTU, and Columbia. The model was evaluated individually on each dataset (D1, D2, D3) and on the combined datasets (D1+D2+D3) to test its robustness and domain generalization. To address dataset imbalance, data augmentation techniques were applied. Additionally, t-SNE visualization was used to demonstrate the feature separability achieved by the Swin Transformer across the selected color spaces.

\textbf{Results}
The model's performance was tested individually on all datasets across all color schemes, with the RGB color scheme yielding the highest accuracy for each dataset. As a result, RGB was selected for domain generalization analysis and compared with other CNN-based models, VGG-19 and ResNet-50. The comparative results demonstrate the proposed model's effectiveness in detecting CGI, highlighting its robustness and reliability in both intra-dataset and inter-dataset evaluations.

\textbf{Conclusion}
The findings of this study highlight the Swin Transformer model's potential as an advanced tool for digital image forensics, particularly in distinguishing CGI from natural images. The model's strong performance across multiple datasets indicates its capability for domain generalization, making it a valuable asset in scenarios requiring precise and reliable image classification.

\end{abstract}

\section{Introduction}
\label{introduction}

\begin{figure}
	\centering
	\includegraphics[width =\columnwidth]{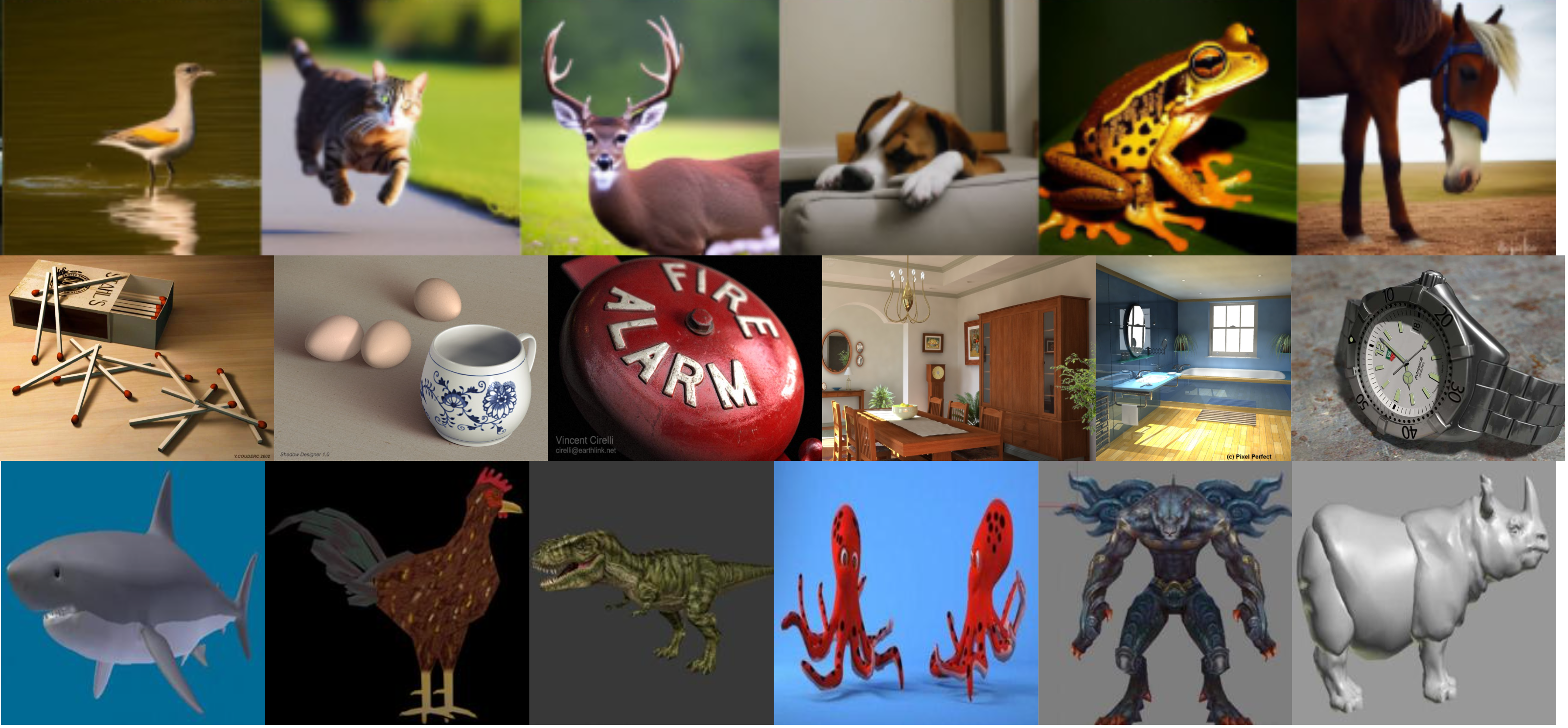}
	
	\caption{Example of few Computer Generated (CG) images from dataset CIFAKE-10, columbia RCGI and JSSSTU datasets (from top to bottom row), respectively. The images shows the difficulty in distingushing between two classes of images with naked eye. Also, the variation in computer generated images}
	\label{fig1}
	
\end{figure}

The rapid proliferation of computer-generated imagery (CGI) has posed a significant challenge in differentiating these synthetic images from natural images (NIs) captured by digital cameras. As advancements in computer graphics achieve increasingly photorealistic results, the visual distinction between CGI and NIs becomes more complex, making it a critical concern in digital forensics. The primary challenge lies in the goal of computer graphics, which is to create synthetic images that closely mimic the authenticity of natural scenes. Consequently, some CGI can appear indistinguishable from real photographs, as illustrated in Figure \ref{fig1} with examples from the CIFAKE-10 \cite{bird2024cifake}, columbia RCGI \cite{ng2005columbia} and JSSSTU\cite{kumar2022dataset} datasets.

Traditional methods for addressing this problem have typically been divided into subjective and objective approaches. Subjective methods often rely on human judgment through psychophysical experiments, whereas objective methods analyze images' statistical and intrinsic properties. Conventional objective techniques usually involve crafting sophisticated and discriminative features and applying classifiers such as support vector machines (SVM) or ensemble models \cite{lyu2005realistic,chen2007identifying,ng2005physics}. Various image editing platforms are examined using three techniques: cut-paste, copy-move, and erase-fill. The findings reveals that the accuracy of detecting forged images is influenced by several factors, including the number of images, the tampering method, image resolution, and many more. While these approaches may perform adequately on more straightforward datasets, they frequently must improve when confronted with complex datasets that include images from diverse sources.

The recent rise of neural networks and vision transformers has transformed the field of computer vision, offering powerful alternatives to feature-based methods. Convolutional Neural Networks (CNNs) have demonstrated exceptional ability in learning multi-level representations from data in an end-to-end manner, making them particularly suitable for complex classification tasks. Inspired by the success of CNNs, there has been growing interest in applying these models to multimedia security and digital forensics \cite{ni2019evaluation,carvalho2017exposing,quan2018distinguishing,yao2022cgnet}.

This research introduces a novel methodology for distinguishing between computer-generated and real digital images, leveraging the capabilities of Swin Transformers. Unlike traditional approaches, Swin Transformers eliminate the reliance on handcrafted feature extraction by directly processing raw pixel data, making them particularly effective for this classification task. Our method involves both intra-dataset and inter-dataset testing, utilizing RGB, YCbCr and HSV color spaces data to enhance classification accuracy. The approach is rigorously evaluated on three diverse datasets; CiFAKE, JSSSTU, and Columbia, demonstrating its robustness and generalization across different domains.
Our contributions are summarized as follows:
 
\begin{itemize} 
	\item We propose a Swin Transformer-based framework for differentiating CGI from real digital images, emphasizing intra-dataset and inter-dataset testing to evaluate robustness. 
	\item The model incorporates advanced preprocessing techniques to enhance classification performance. 
	\item The proposed model was tested for different color spaces, namely, RGB, YCbCr and HSV.
	\item The comparative analysis with different CNN and transfomer baseline models shows the effectiveness of our proposed model.
	\item Our approach achieves state-of-the-art accuracy, consistently between 97-99\% across multiple datasets, demonstrating its efficacy in CGI detection. 
\end{itemize}

The remainder of this paper is structured as follows: Section \ref{introduction} outlines the challenge of distinguishing CGI from real images, emphasizing its importance in digital image forensics and the contributions of the research work. Section \ref{relatedwork} reviews existing methods, from traditional techniques to recent profound learning advancements and the various dataset availablility. Section \ref{method} details our proposed approach, including preprocessing steps and the Swin Transformer architecture used for classification. Section \ref{result} presents the experimental results, describing dataset usage, experimental setups, comparative analysis and performance evaluation. Finally, Section \ref{conclusion} concludes the paper with a summary of findings, a discussion of limitations, and suggestions for future research directions.

\section{Related Work}
\label{relatedwork}

Various methodologies have been developed in computer-generated (CG) image detection, primarily focusing on feature extraction and classification. One approach is to extract abnormal statistical traces left by specific graphic generation modules and employ threshold-based evaluation for detection. Ng et al. (2005) \cite{ng2005physics} identified physical disparities between photographic and computer-generated images, achieving an 83.5\% classification accuracy by designing object geometry features. Wu et al. (2006) \cite{wu2006detecting} utilized visual features such as color, edge, saturation, and texture with the Gabor filter as discriminative features. Dehnie et al. (2006) \cite{dehnie2006digital} emphasized differences in image acquisition between digital cameras and generative algorithms, designing features based on residual images extracted by wavelet-based denoising filters. Texture-based methods have also been developed for CG and photographic (PG) classification. Li et al. (2014) \cite{li2014distinguishing} achieved 95.1\% accuracy using uniform gray-scale invariant local binary patterns. Peng et al. (2015) \cite{peng2015identification} combined statistical and textural features, enhancing performance with histogram and multi-fractal spectrum features. Despite their interpretability, hand-crafted feature-based methods are constrained by manual design limitations and feature description capacities.

In response to the limitations of hand-crafted features, recent research has leaned towards leveraging deep learning methods for improved detection performance. For instance,
Mo et al. (2018) \cite{mo2018fake} proposed a CNN-based method focusing on high-frequency components, achieving an average accuracy of over 98\%. Meena and Tyagi (2021) \cite{meena2021distinguishing} proposed a two-stream convolutional neural network, integrating a pre-trained VGG-19 network for trace learning and employing high-pass filters to emphasize noise-based features. Yao et al. (2022) \cite{yao2022cgnet} introduced a novel approach utilizing VGG-16 architecture combined with a Convolutional Block Attention Module, achieving an impressive accuracy of 96\% on the DSTok dataset. Furthermore, Chen et al. (2021) \cite{chen2021locally} presented an enhanced Xception model tailored for locally generated face detection in GANs. Liu et al. (2022) \cite{liu2022detecting} introduced a method focusing on authentic image noise patterns for detection. These recent advancements underscore the growing trend of applying sophisticated deep-learning architectures to enhance the detection accuracy of computer-generated images, addressing the challenges posed by increasingly realistic synthetic imagery. Fang et al. (2023) \cite{Fang2023} presented a robust blind quality metric for evaluating altered images from Depth-image-based rendering (DIBR), considering both local and global features of the images and tested on multiple benchmark datasets.

Some of the researchers have provided a combination of subjective and objective approaches to identify real and altered images.  Yue et al. (2024) \cite{yuesubjective2024} developed an approach for assessing the quality of retouched face images, combining subjective evaluations and a multi-task learning-based No-Reference Image Quality Assessment method. Min et al. (2024) \cite{minexploring2024} used a three-stage Convolutional vision Transformer (CvT) network and a multi-label training strategy to predict the quality of the wild images based on various subjective information.  
Li J and Wang K \cite{li2024detecting} (2024) presented a data-efficient approach for detecting computer-generated images using the CLIP image encoder and a limited set of real and proxy images, avoiding the need for extensive synthetic datasets. Employing proxy sample construction methods like frequency masking and color transfer achieves high accuracy and generalizes well across synthetic image types, such as GANs and diffusion models. The research underscores the value of proxy samples for improving detection robustness in data-limited scenarios.

The study in \cite{ladjevic2024detection} (2024) presents a lightweight CNN-based approach with eight convolutional layers to effectively classify AI-generated images, achieving a high accuracy of 97.32\% on the CIFAKE dataset, comparable to leading detection methods. The model's performance was evaluated on both standard and custom datasets, with results surpassing four state-of-the-art methods. Additionally, explainable AI was employed to reveal distinct activation patterns, showing that real images focus activations on primary objects, while synthetic images activate around edges or complex textures, advancing understanding in AI-generated image detection. Authors in \cite{anon2024detecting} (2024) presented a framework for distinguishing real images from those generated by advanced AI models, including DALL-E 3, MidJourney, and Stable Diffusion 3, through a hybrid Kolmogorov-Arnold Network (KAN) and MLP architecture. By leveraging a dataset specifically crafted for cutting-edge AI generators, this model identifies complex patterns in synthetic images, outperforming standard MLPs across diverse datasets with high F1 scores. The method shows promising application in digital forensics and content verification, although data collection costs from high-end AI generators present a scaling challenge.

\begin{table}[!htbp]
	\centering
	\caption{Summary of Computer-Generated Image Datasets}
	\label{tab:cgi_datasets}
	\resizebox{\columnwidth}{!}{%
		\begin{tabular}{|p{2cm}|p{1.5cm}|p{2cm}|p{4cm}|p{3cm}|p{1cm}|}
			\hline
			\textbf{Dataset}        & \textbf{Labelled} & \textbf{No. of Images (Real/Fake)} & \textbf{Generation Technique (CGI)}     & \textbf{URL}                                                                                     & \textbf{Paper Link}                                                                                     \\ \hline
			CiFAKE   \cite{bird2024cifake}                   & \checkmark                    & ~1,20,000       & 3D Rendering Engines                    & \href{https://www.kaggle.com/datasets/birdy654/cifake-real-and-ai-generated-synthetic-images}{CiFAKE Dataset}                                         & \href{https://paperswithcode.com/dataset/cifake-real-and-ai-generated-synthetic-images}{1}                                                 \\ \hline
			JSSSTU Dataset \cite{kumar2022dataset}             & \checkmark                    & ~5,000 each          & Blender, Maya, and other 3D modeling tools & \href{https://drive.google.com/drive/u/0/folders/132t7GHRnRFkve7H2oTIYqQrIt8WyZDmR}{JSSSTU Dataset}             & \href{http://doi.org/10.11591/ijai.v11.i1.pp137-147}{2}                     \\ \hline
			Columbia Dataset  \cite{ng2005physics}          & \checkmark                    & 800 each          & 3D Studio Max, rendering tools         & \href{https://www.ee.columbia.edu/~dvmmweb/dvmm/downloads/PIM_PRCG_dataset/dataset-download.htm}{Columbia Dataset}                 & \href{https://www.ee.columbia.edu/ln/dvmm/publications/05/ng_acmmm05.pdf}{3}                         \\ \hline
			FFHQ (Flickr-Faces-HQ)  \cite{karras2019style}          & \checkmark                   & ~70,000 each     & GAN (StyleGAN)                          & \href{https://github.com/NVlabs/stylegan}{StyleGAN Resources}                                    & \href{https://arxiv.org/abs/1812.04948}{4}                                               \\ \hline
			DeepFake Detection   \cite{dolhansky2020deepfake}       & \checkmark                    & ~3,500 videos   & DeepFake Algorithm                      & \href{https://www.kaggle.com/c/deepfake-detection-challenge/data}{DeepFake Challenge Dataset}    & \href{https://kaggle.com/competitions/deepfake-detection-challenge}{5}                                     \\ \hline
			Synthetic Face Datasets  \cite{beniaguev2022synthetic}   & \checkmark                    & ~425,000                             & GAN, Neural Rendering                   & \href{https://www.kaggle.com/datasets/selfishgene/synthetic-faces-high-quality-sfhq-part-1}{Synthetic Face Datasets}                                & \href{https://github.com/SelfishGene/SFHQ-dataset}{6}                                        \\ \hline
			
			BigGAN 128 Datasets  \cite{brock2018large}           & \checkmark                  & ~500 each & GAN (BigGAN)                            & \href{https://www.kaggle.com/code/cdeotte/big-gan-128-lb-12}{BigGAN Dataset}                                    & \href{https://openreview.net/forum?id=B1xsqj09Fm}{7}                                                 \\ \hline
		\end{tabular}%
	}
\end{table}

The evaluation of computer-generated image (CGI) detection techniques requires diverse datasets that include both real and synthetic images generated by different methods. Table \ref{tab:cgi_datasets} presents a comprehensive summary of prominent datasets in this field. Each dataset is characterized by its labeling status, the number of real and fake images, the generation technique employed for synthetic image creation, and relevant links to access the datasets and associated research papers. These datasets are critical for training and benchmarking algorithms in domains such as digital image forensics, domain generalization, and CGI detection.

\section{Proposed Methodology}
\label{method}
\label{methodology}
\begin{figure*}[!htbp]
	\centering
	
	\includegraphics[width=\columnwidth]{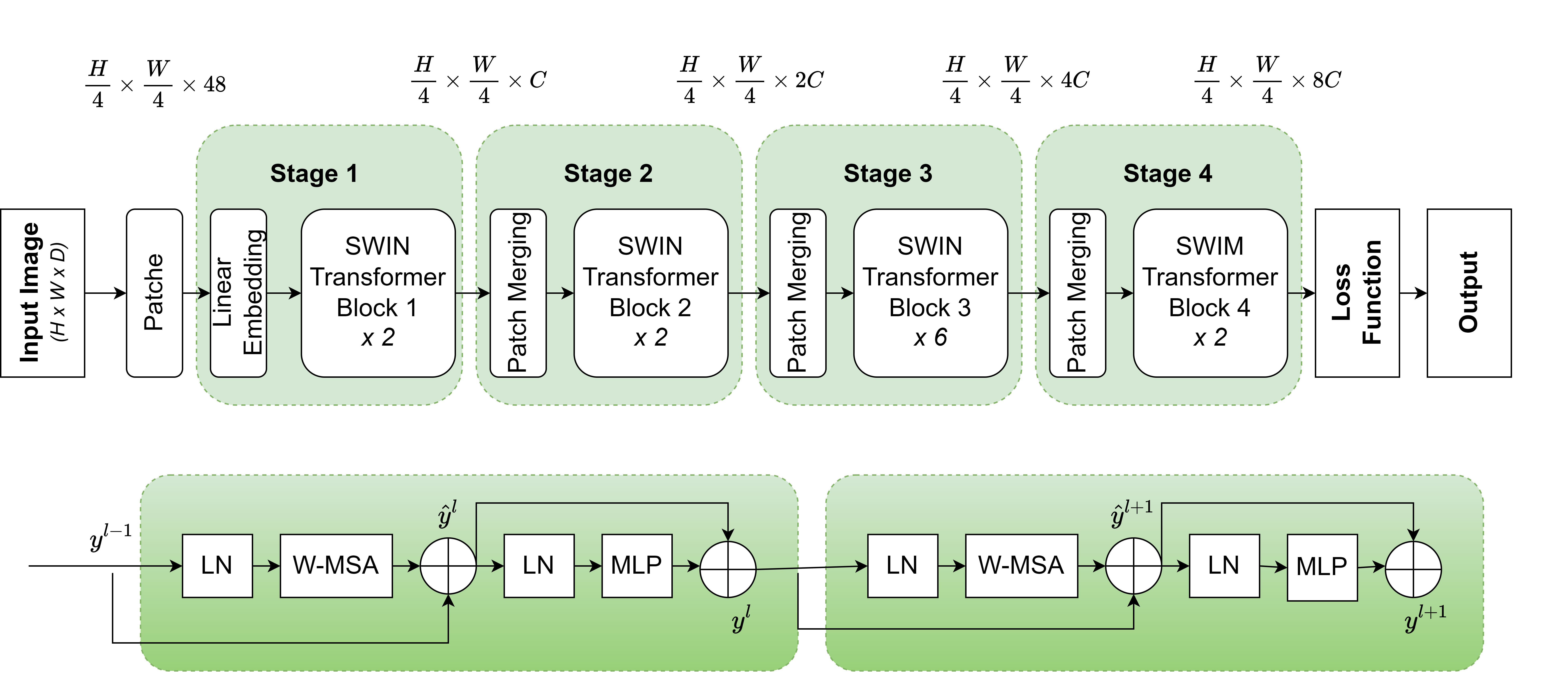}
	\caption{Architecture of the proposed Swin transformer and the expansion of Swin Transformer Block under it.}
	\label{fig2}
\end{figure*}

\subsection{Swin Transformer Architecture}
The Swin Transformer is a cutting-edge deep learning architecture renowned for effectively processing large-scale visual data with hierarchical representations. Unlike traditional convolutional neural networks (CNNs), the Swin Transformer adopts a hierarchical architecture that efficiently captures long-range spatial dependencies across the input images.

The architecture figure \ref{fig2} illustrates the hierarchical structure of the Swin Transformer, showcasing its ability to capture global context and local details simultaneously. By leveraging self-attention mechanisms and multi-layered processing, the Swin Transformer excels at learning complex patterns in visual data, making it ideal for image classification tasks.
It is characterized by self-attention mechanisms. Mathematically, the self-attention mechanism of the Swin Transformer can be represented as follow:

\[
\text{Att}(Q, K, V) = \text{softmax} \left( \frac{QK^T}{\sqrt{d_k}} \right) V
\]

where \( Q \), \( K \), and \( V \) represent the query, key, and value matrices, respectively, and \( d_k \) denotes the dimensionality of the key vectors.

Our study employed the Swin Transformer to distinguish between computer-generated images (CGI) and authentic images. We conducted color frame analysis to enhance the model's understanding of the distinct characteristics of CGI and authentic images. By analyzing the three different color spaces, namely, RGB, YCbCr and HSV color frames and extracting meaningful features, we aimed to provide the model with valuable insights into the color distribution and spatial arrangements between CGI and authentic images for all the three datasets.

\subsection{Feature Visualization with t-SNE}

To visualize the impact of color frame analysis on feature extraction, we employ t-Distributed Stochastic Neighbor Embedding (t-SNE) plots. Mathematically, t-SNE minimizes the Kullback-Leibler divergence between the distribution of high-dimensional feature vectors and their low-dimensional counterparts. The t-SNE algorithm can be represented as:

\[
p_{ij} = \frac{\exp(-\lVert x_i - x_j \rVert^2 / 2 \sigma_i^2)}{\sum_{k \neq l} \exp(-\lVert x_k - x_l \rVert^2 / 2 \sigma_k^2)}
\]
\[
q_{ij} = \frac{(1 + \lVert y_i - y_j \rVert^2)^{-1}}{\sum_{k \neq l} (1 + \lVert y_k - y_l \rVert^2)^{-1}}
\]
\[
C = \sum_i KL(P_i || Q_i) = \sum_{ij} p_{ij} \log \frac{p_{ij}}{q_{ij}}
\]

Here, \( p_{ij} \) represents the pairwise similarity between points \( x_i \) and \( x_j \) in the high-dimensional space, \( q_{ij} \) denotes the pairwise similarity between points \( y_i \) and \( y_j \) in the low-dimensional space, and \( C \) represents the Kullback-Leibler divergence.

\begin{figure}[!htbp]
	\centering
	\subfigure[\label{fig3a}]{\includegraphics[width=4cm,height=3cm]{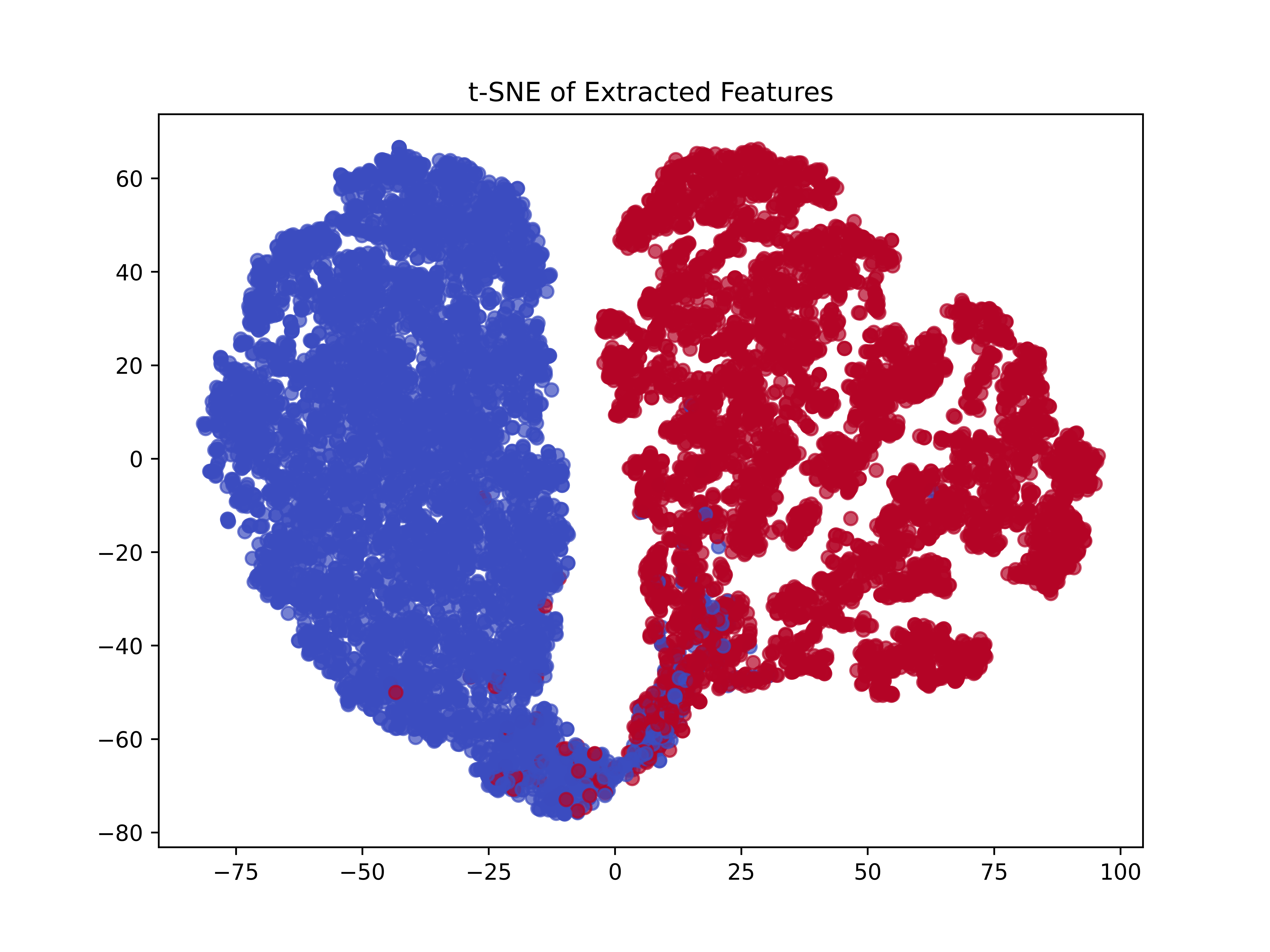}}
	\subfigure[\label{fig3b}]{\includegraphics[width=4cm,height=3cm]{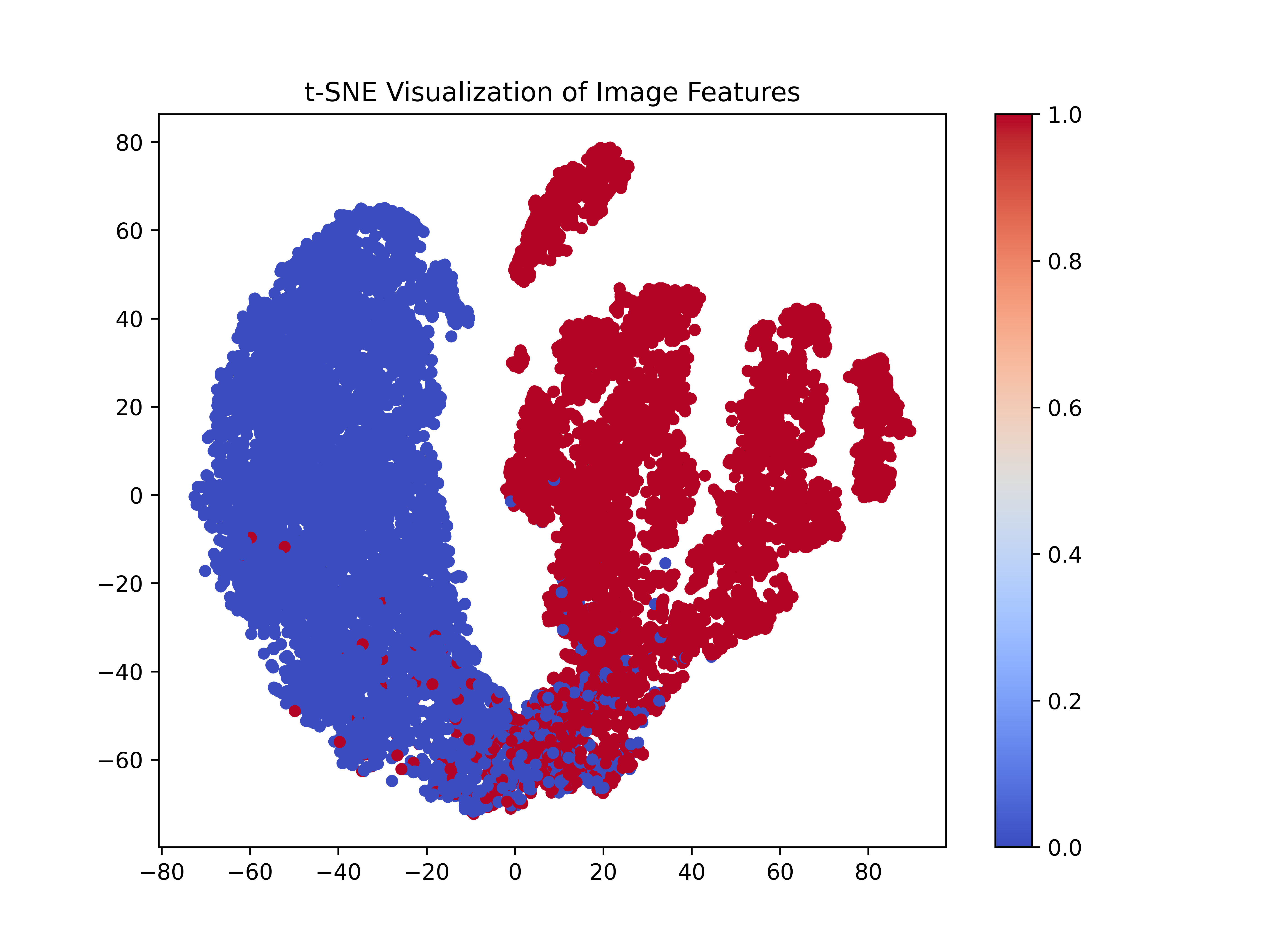}}
	\subfigure[\label{fig3c}]{\includegraphics[width=4cm,height=3cm]{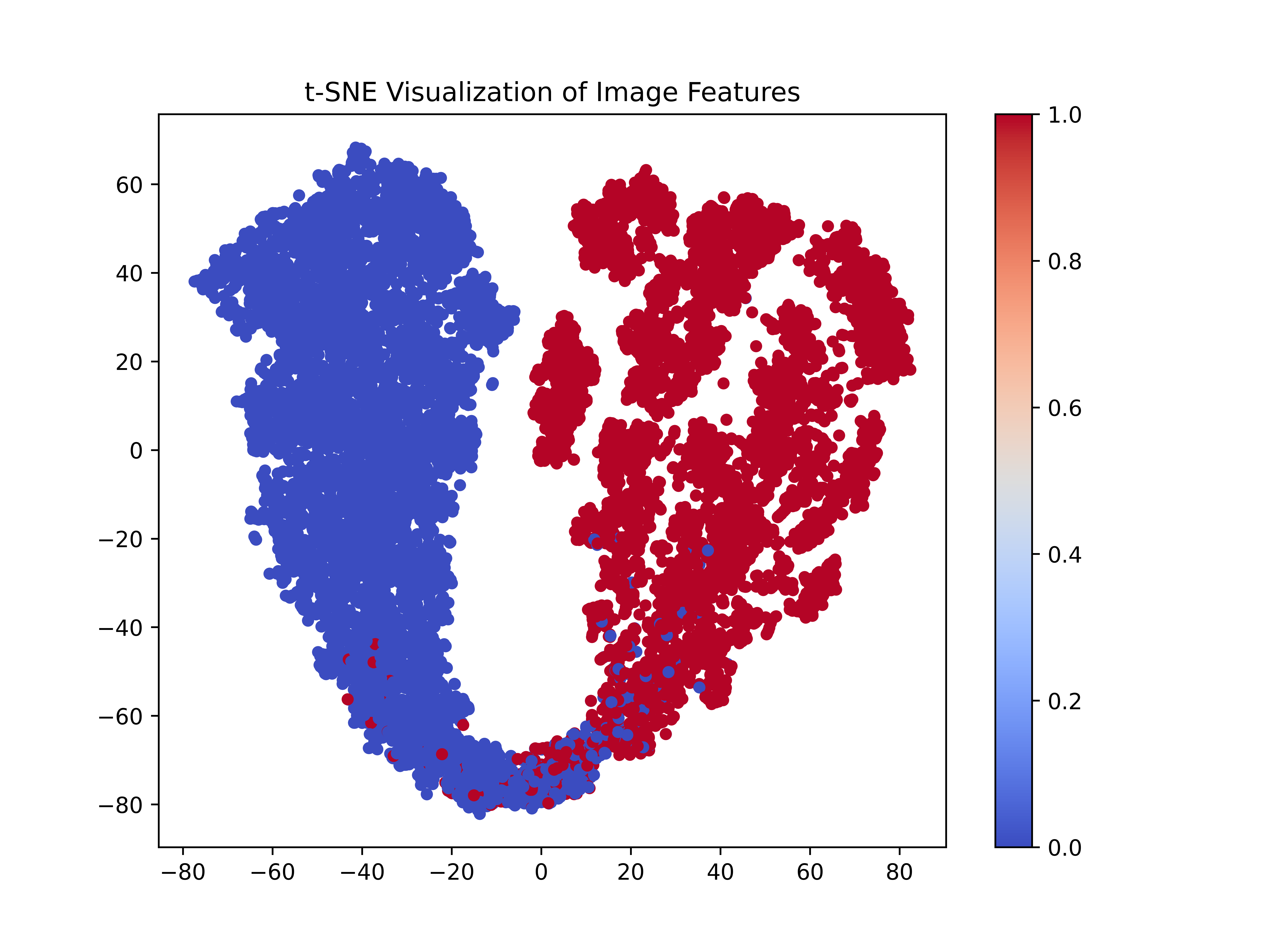}}
	
	\caption{The plot illustrate the t-SNE plots of the extracted features from the Swin Transformer for the CiFake dataset (D1) images for the RGB, YCbCr and HSV color space (from left to right)}
	\label{fig3}
\end{figure}

\begin{figure}[!htbp]
	\centering
	\subfigure[\label{fig4a}]{\includegraphics[width=4cm,height=3cm]{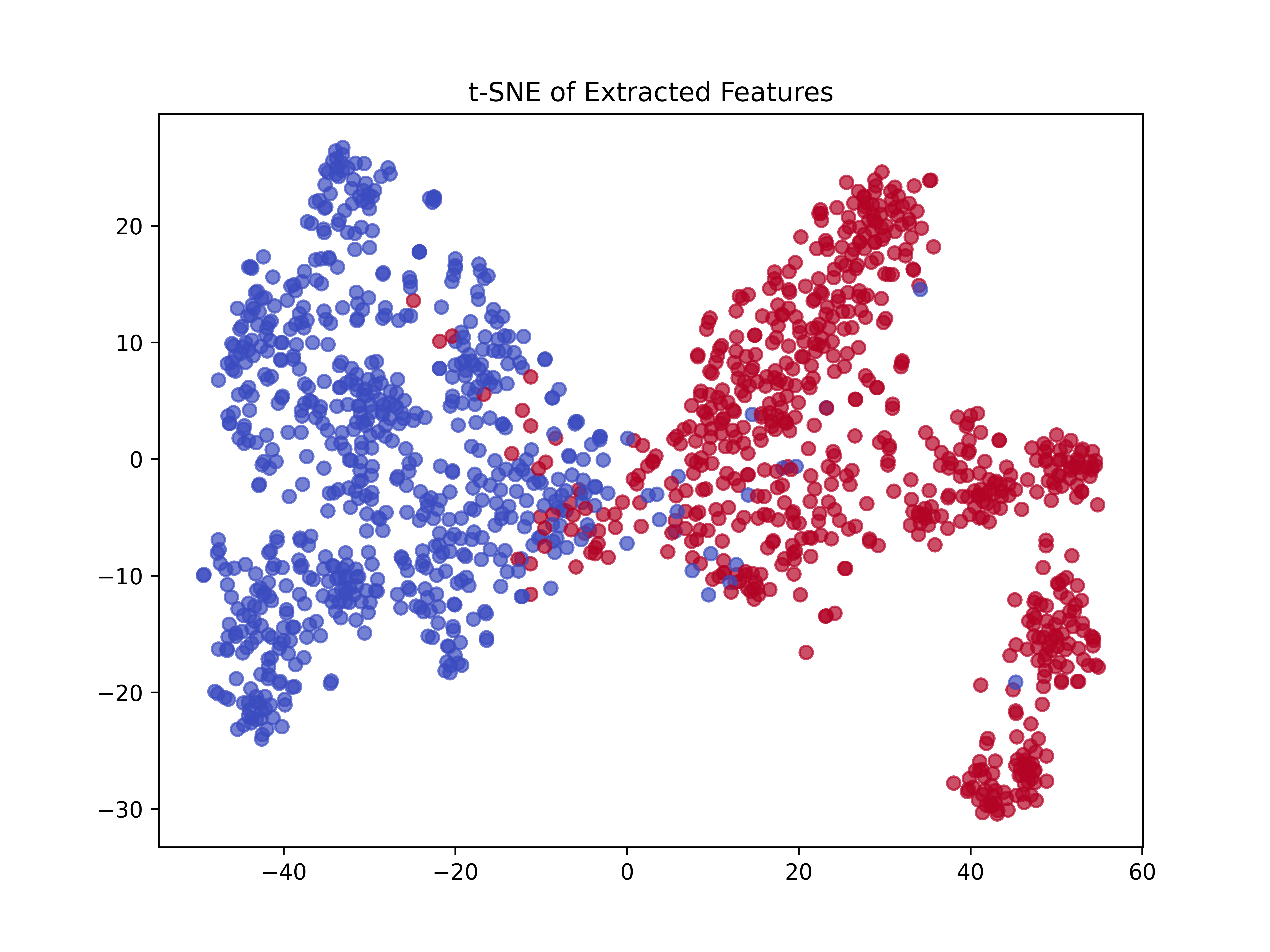}}
	\subfigure[\label{fig4b}]{\includegraphics[width=4cm,height=3cm]{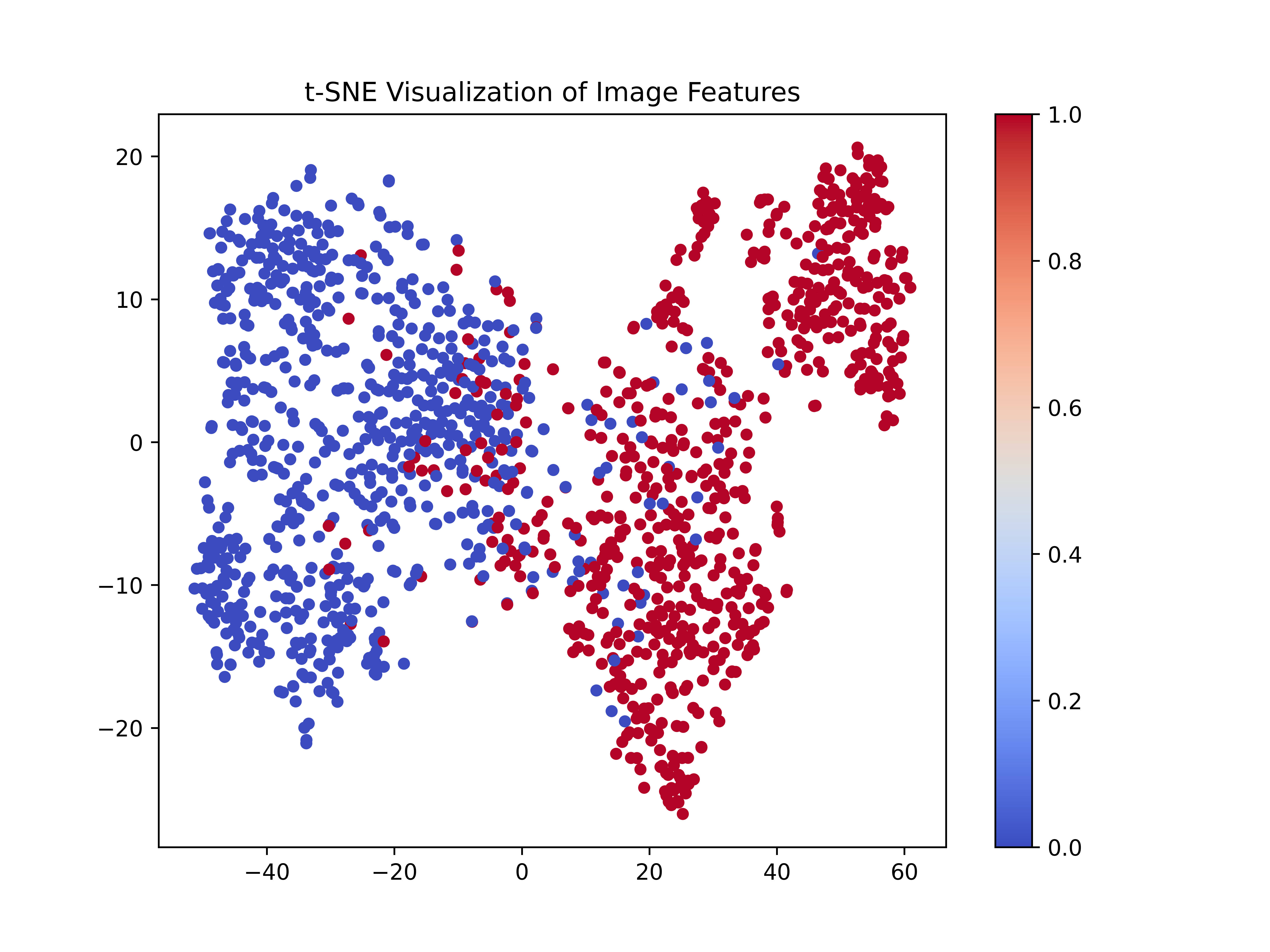}}
	\subfigure[\label{fig4c}]{\includegraphics[width=4cm,height=3cm]{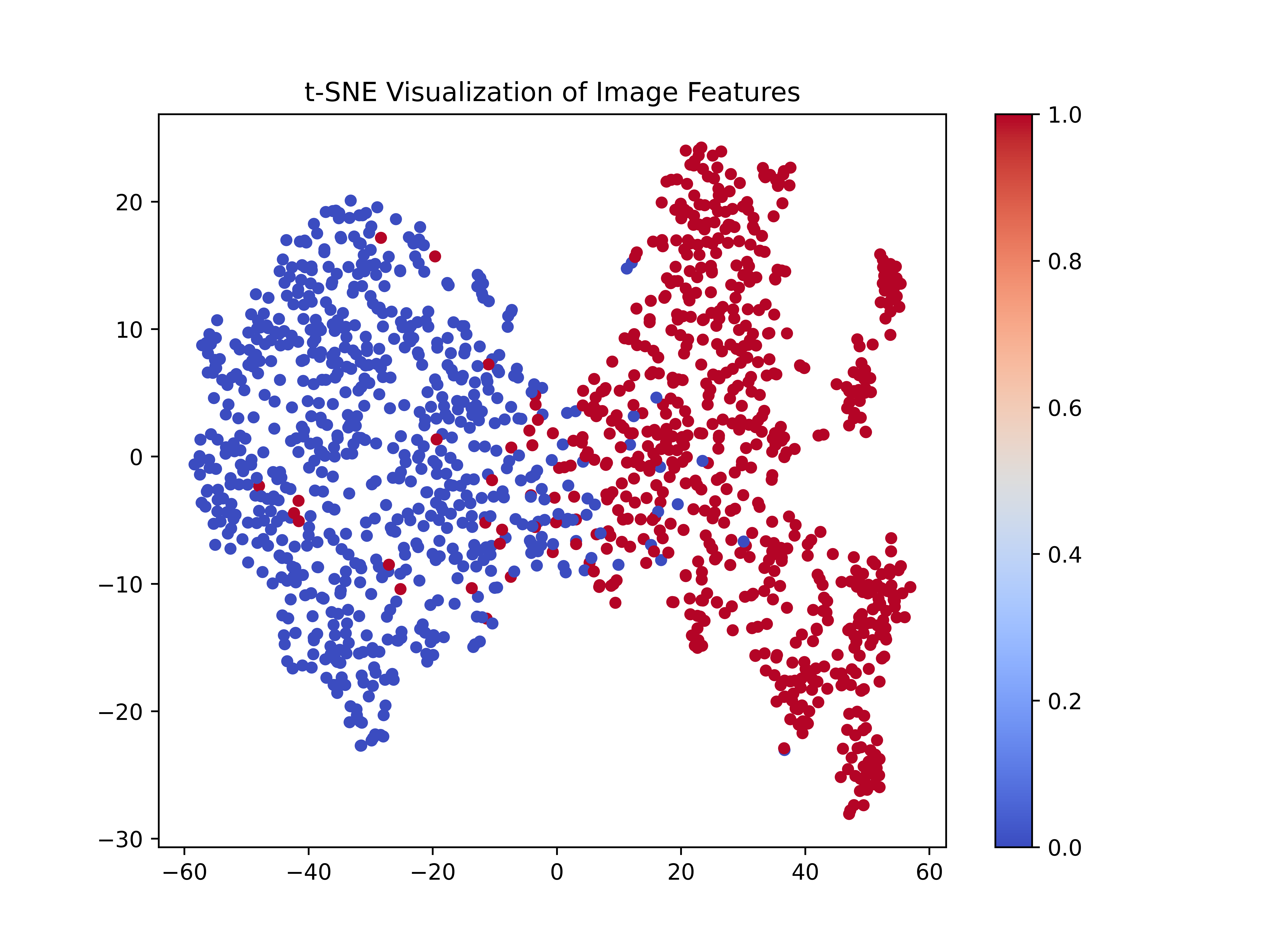}}
	
	\caption{The plot illustrate the t-SNE plots of the extracted features from the Swin Transformer for the JSSSTU dataset (D2) images for the RGB, YCbCr and HSV color space (from left to right)}
	\label{fig4}
\end{figure}

\begin{figure}[!htbp]
	\centering
	\subfigure[\label{fig5a}]{\includegraphics[width=4cm,height=3cm]{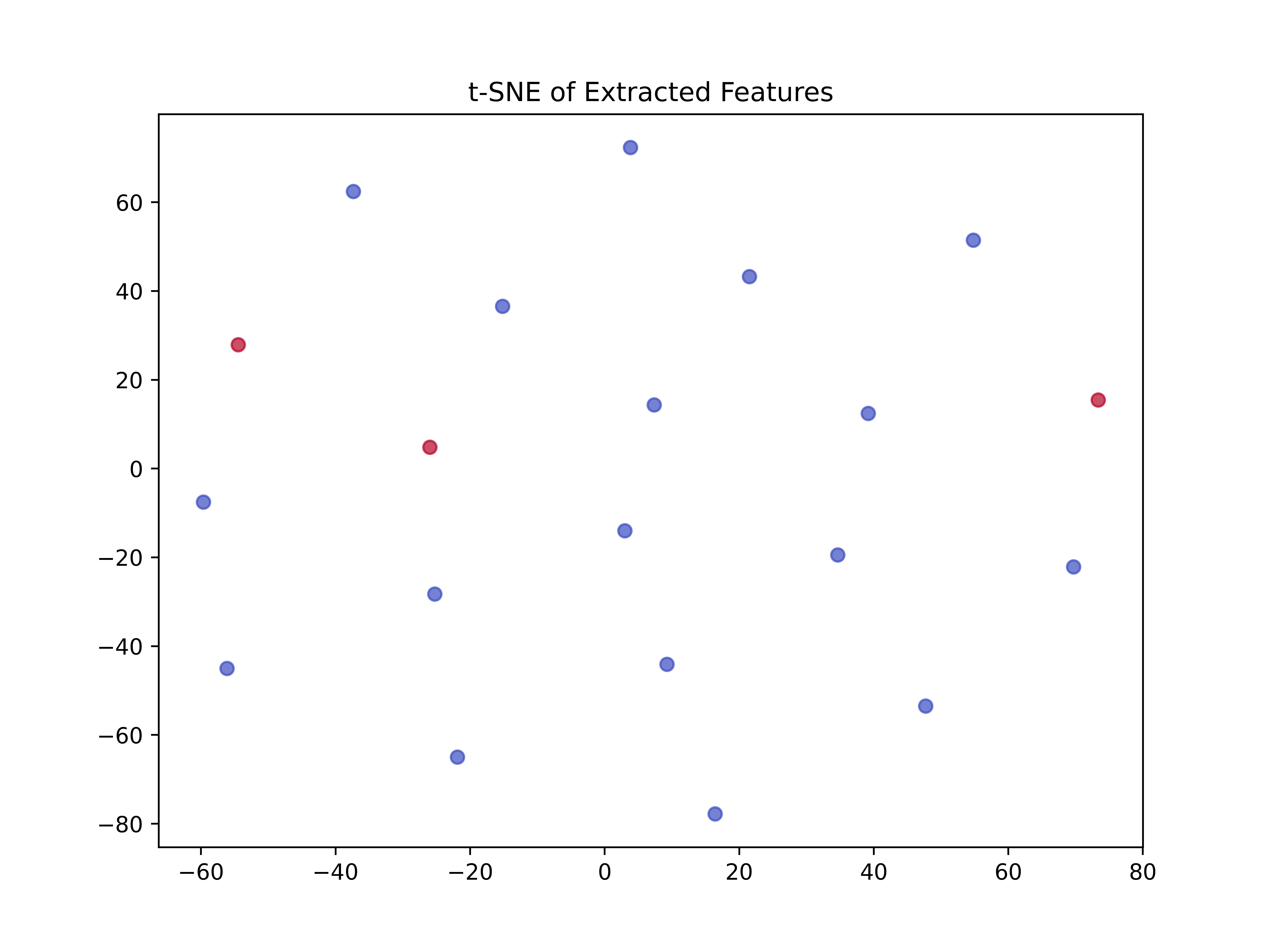}}
	\subfigure[\label{fig5b}]{\includegraphics[width=4cm,height=3cm]{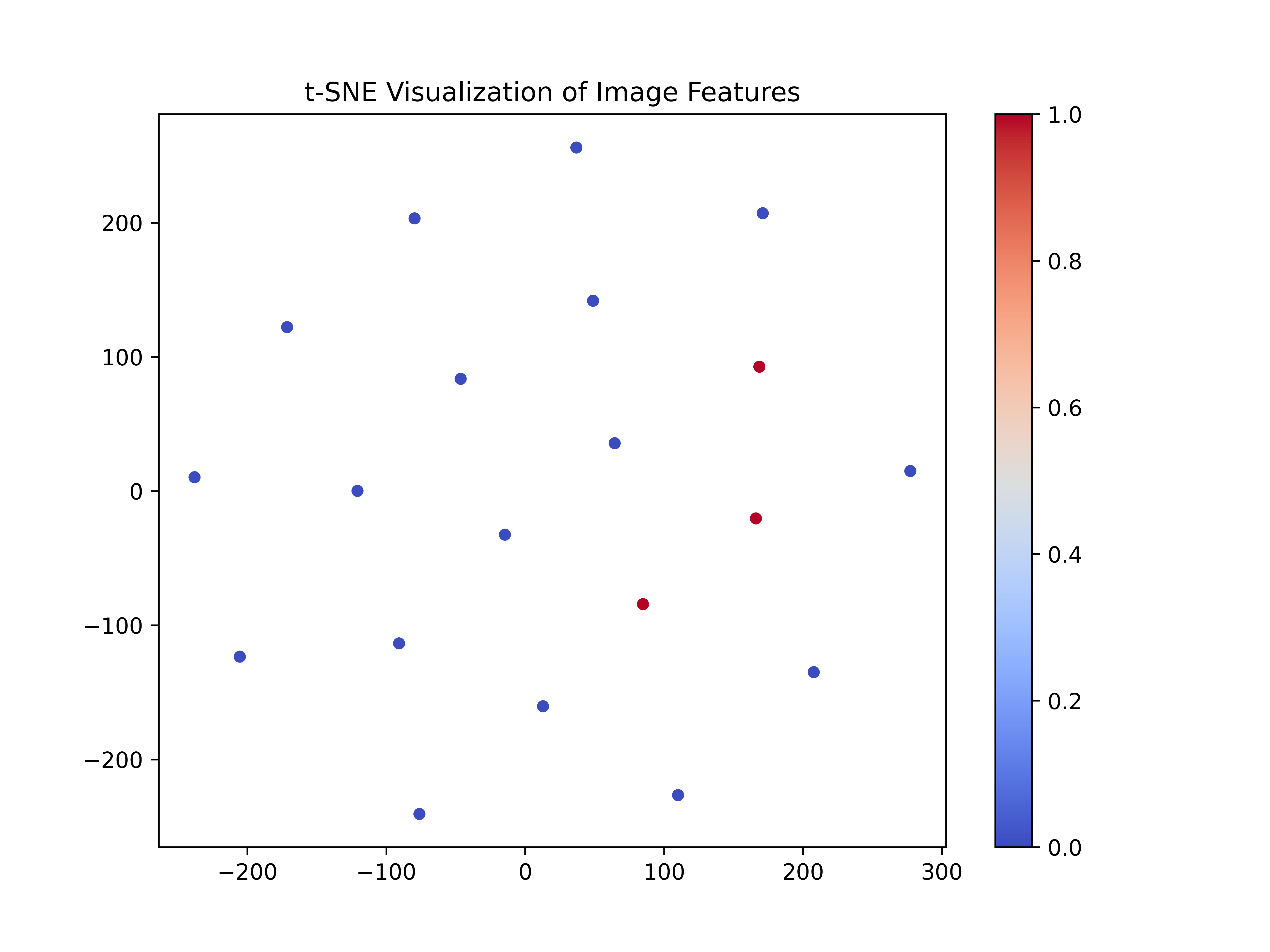}}
	\subfigure[\label{fig5c}]{\includegraphics[width=4cm,height=3cm]{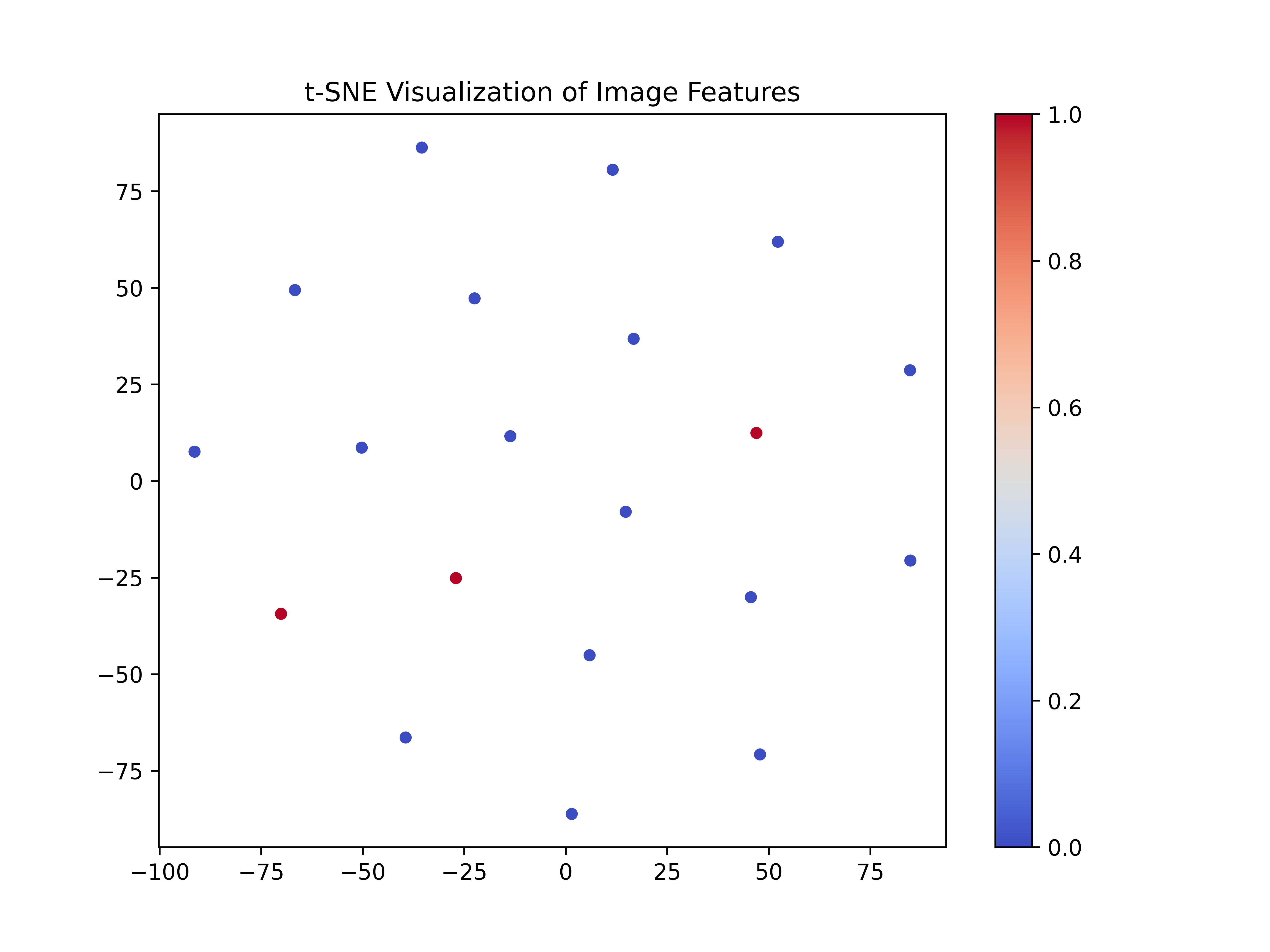}}
	
	\caption{The plot illustrate the t-SNE plots of the extracted features from the Swin Transformer for the Columbia PRCG and real images dataset (D3) images for the RGB, YCbCr and HSV color space (from left to right).}
	\label{fig5}
\end{figure}

The t-SNE visualizations in Figures \ref{fig3}, \ref{fig4}, and \ref{fig5} illustrate the feature separability achieved by the Swin Transformer for the CiFake (D1), JSSSTU (D2), and Columbia PRCG (D3) datasets across RGB, YCbCr, and HSV color spaces. In all three datasets, the RGB color space consistently demonstrates the most distinct separation between the CGI (red) and authentic image (blue) features, forming well-defined clusters with minimal overlap. This superior separability aligns with RGB’s highest classification performance across the datasets. Conversely, the YCbCr and HSV color spaces exhibit moderately overlapping clusters, leading to similar but slightly reduced classification accuracy compared to RGB.

The comparative analysis highlights RGB as the most effective color space for feature extraction and classification due to its ability to maximize inter-class separability. Although the YCbCr and HSV spaces perform comparably, they exhibit increased feature overlap, resulting in diminished separability and lower accuracy. These observations underscore the critical role of feature distribution in determining classification effectiveness and reaffirm the dominance of RGB in achieving optimal performance with the Swin Transformer.

\begin{figure}[!htbp]
	\centering
	\subfigure[\label{fig6a}]{\includegraphics[width=4cm,height=3cm]{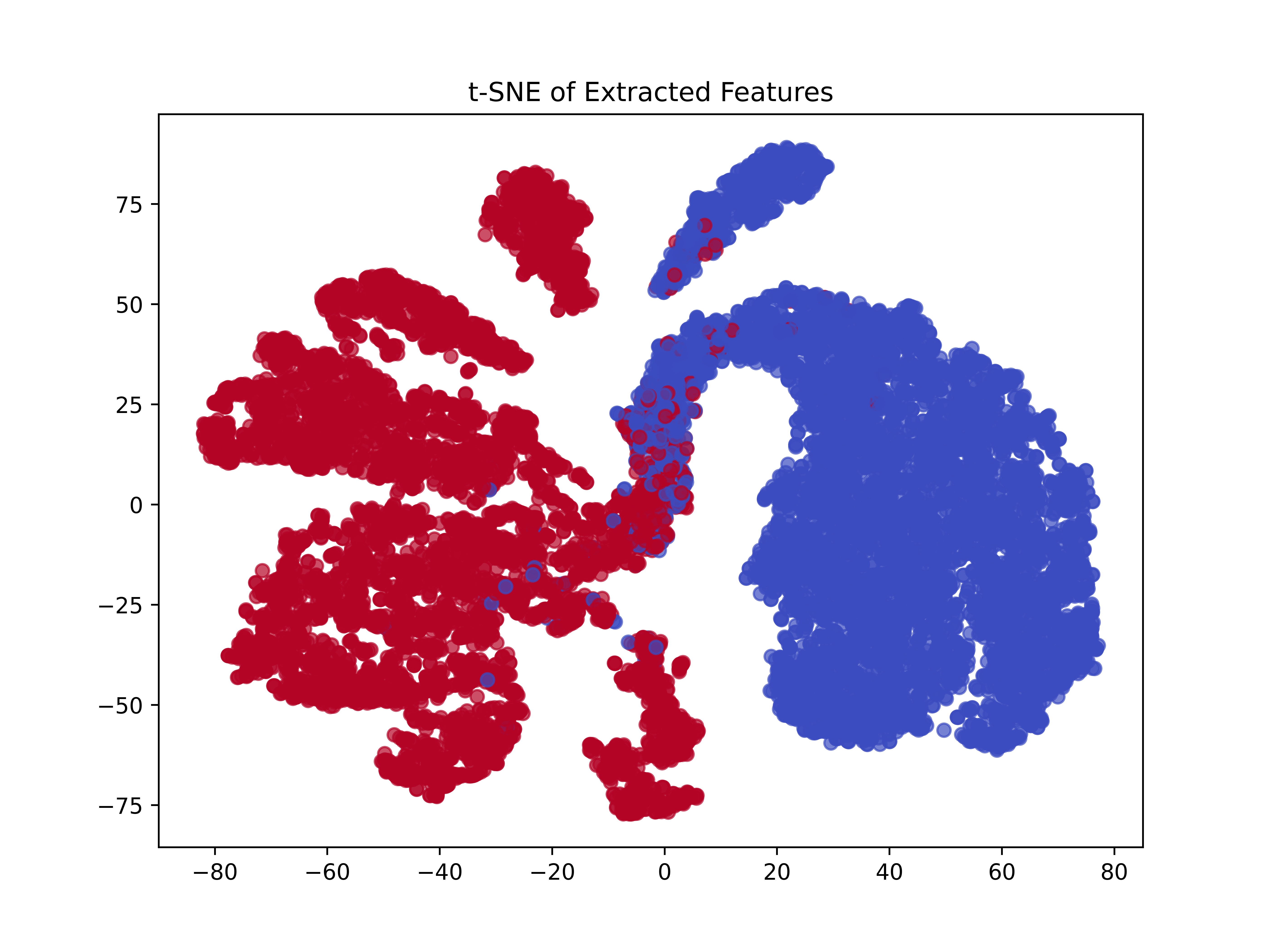}}
	\subfigure[\label{fig6b}]{\includegraphics[width=4cm,height=3cm]{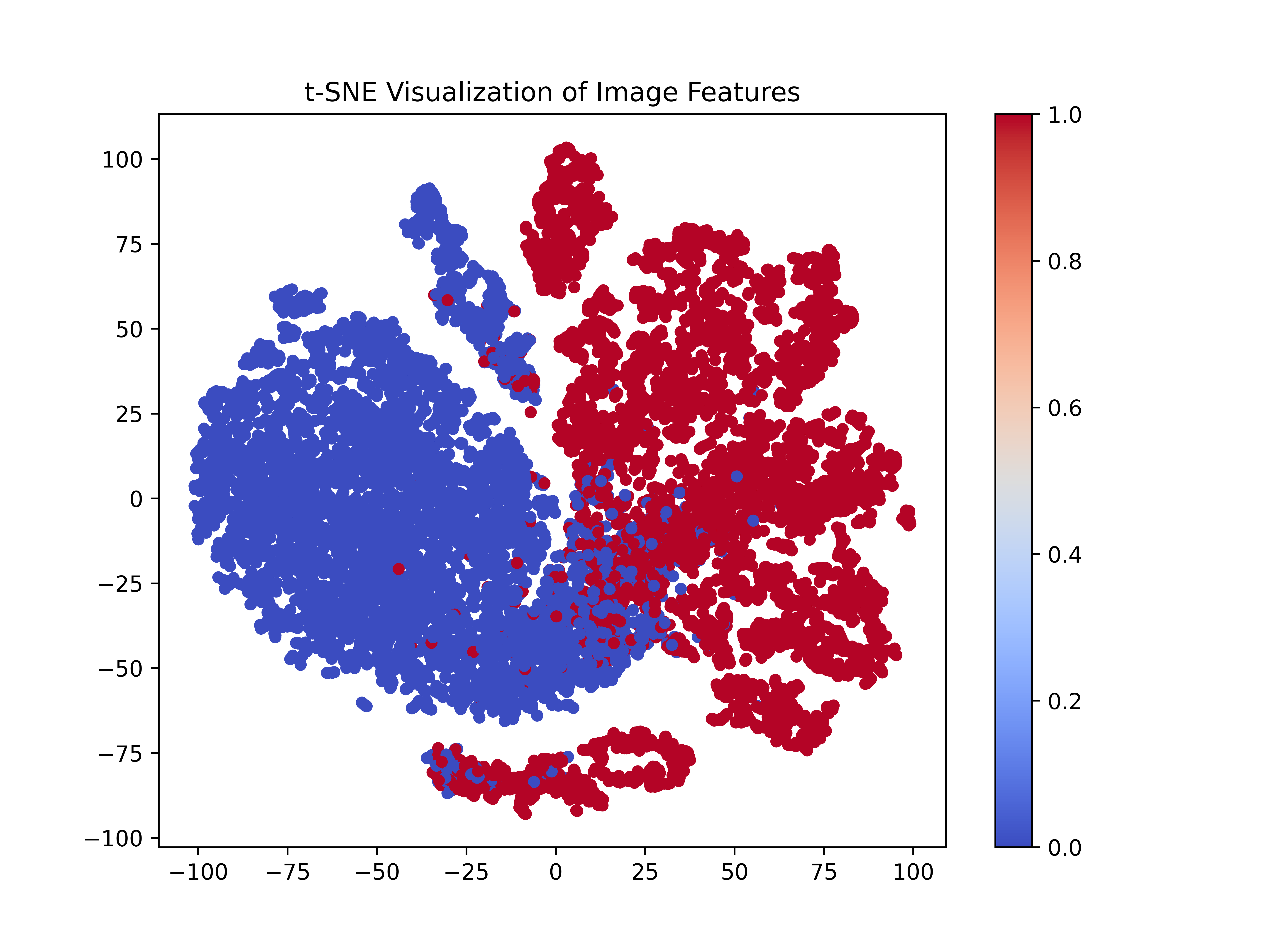}}
	\subfigure[\label{fig6c}]{\includegraphics[width=4cm,height=3cm]{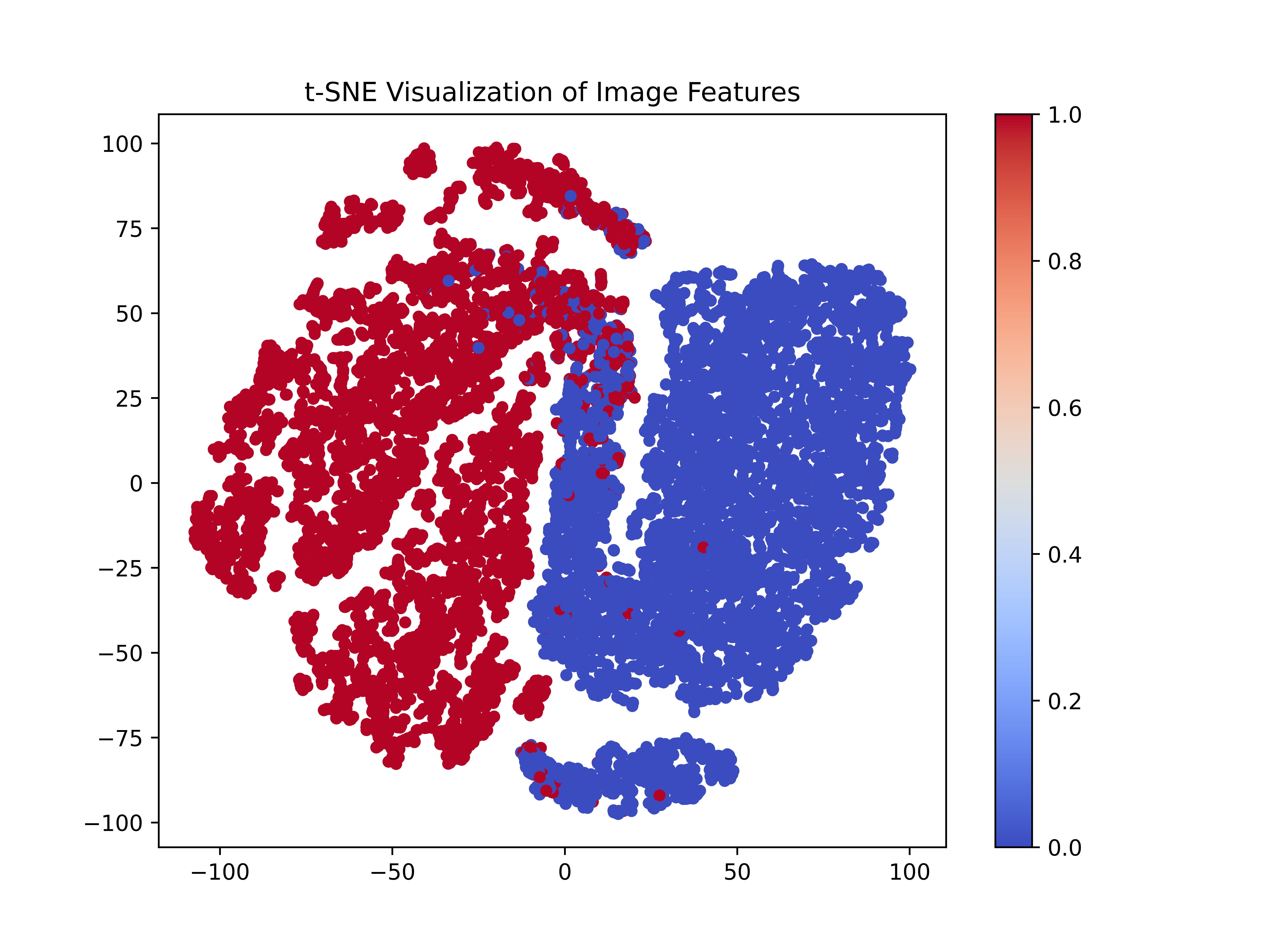}}
	
	\caption{The plot illustrate the t-SNE plots of the extracted features from the Swin Transformer for the combined datasets (D1+D2+D3) images for the RGB, YCbCr and HSV color space (from left to right)}
	\label{fig6}
\end{figure}
\section{Results and Discussion}
\label{result}

The Swin Transformer was trained utilizing features extracted from different color frames exclusively, owing to the characteristics of the datasets employed. The training objective centered on minimizing the cross-entropy loss function, defined as:
\[
\mathcal{L} = - \frac{1}{N} \sum_{i=1}^{N} \sum_{j=1}^{C} y_{ij} \log(p_{ij})
\]

where \( N \) is the total number of samples, \( C \) represents the number of classes, \( y_{ij} \) is the ground truth label, and \( p_{ij} \) is the predicted probability.

\subsection{Dataset Challenges and Curation}

The experimental setup involved three distinct datasets: CiFAKE (D1), Columbia (D2), and JSSSTU (D3). Each dataset comprised images classified into two categories: CGI and authentic images. However, we encountered significant challenges, particularly with the Columbia dataset. Numerous images from this dataset were corrupted or inaccessible due to broken URLs (HTTP 404 errors), substantially reducing the available image count. Specifically, only 43 CGI and 150 authentic images were retrievable from the intended 800 images per class.

\begin{table}[!htbp]
	\centering
	\caption{Comparison of Dataset Attributes used for evaluation of results.}
	\label{dataset_comparison}
	\resizebox{\columnwidth}{!}{
		\begin{tabular}{|l|p{4cm}|p{4cm}|p{4cm}|}
			\hline
			\textbf{Attribute} & \textbf{CiFAKE} & \textbf{Columbia} & \textbf{JSSSTU} \\ \hline
			\textbf{URL} & \href{https://www.kaggle.com/datasets/birdy654/cifake-real-and-ai-generated-synthetic-images}{Kaggle} & \href{https://www.ee.columbia.edu/~dvmmweb/dvmm/downloads/PIM\_PRCG\_dataset/dataset-download.htm}{Columbia University} & \href{https://sites.google.com/view/hrchennamma/research-activities/jssstu-data-sets}{JSSSTU} \\ \hline
			\textbf{Dataset Type} & Real \& AI-Generated Images & Photo-Realistic Computer Graphics (PRCG) \&Real & CGI \& Real Images \\ \hline
			\textbf{Image Count} & 1,000,000+ & 43 CGI, 150 Real (usable) & 14000 \\ \hline
			\textbf{Class Labels} & Fake, Real & PRCG, Google set & CG, PG \\ \hline
			\textbf{Image Resolution} & Varies & 1280x720 (typical) & Varies \\ \hline
			\textbf{Image Quality} & High-quality AI-generated and real images & Mixed quality, with some corrupted images & High-quality images \\ \hline
			\textbf{File Formats} & JPG & JPEG & JPG \\ \hline
			\textbf{Class Distribution} & Balanced & Imbalanced due to data loss & Balanced \\ \hline
	\end{tabular}}
\end{table}

To address the imbalance and limited number of images in the D3 dataset, which comprises only 43 CGI and 150 real images, data augmentation techniques were applied to increase the dataset size and enhance its representativeness. To ensure a balanced and comprehensive evaluation, we combined the images from all three datasets (D1, D2, and D3) to create a unified dataset (D1+D2+D3). This composite dataset contains 1500 images per class (CGI and authentic), providing a robust foundation for model training and evaluation. Table \ref{dataset_comparison} presents a detailed comparison of the attributes of the three datasets—CiFAKE, Columbia, and JSSSTU.

\subsection{Model Evaluation Settings}

The performance of the proposed methodology was assessed using standard classification metrics: accuracy, precision, recall, and F1-score. These metrics are critical in quantifying the model's effectiveness in distinguishing between CGI and authentic images based on the extracted features. The model was implemented in PyTorch and executed on a Dell Inspiron 5502, equipped with an Intel Core i5 processor and 16GB of RAM. Table \ref{table1} outlines the specific hyperparameters utilized during training.

\begin{table}[!htbp]
	\centering
	\caption{Swin Transformer Hyperparameters Values}
	\label{table1}
	\begin{tabular}{|l|c|}
		\hline
		\textbf{Hyperparameters} & \textbf{Values}\\
		\hline
		Learning Rate & 0.0001\\ \hline
		Optimizer & Adam\\ \hline
		Loss Function & Cross Entropy\\ \hline
		Batch Size & 32\\ \hline
		Input Image Size & 224$\times$224$\times$3\\ \hline
		Normalized Mean & [0.485, 0.456, 0.406]\\ \hline
		Normalized Std & [0.229, 0.224, 0.225]\\ \hline
		Epochs & 20\\
		\hline
	\end{tabular}
\end{table}

The composite dataset (D1+D2+D3) derived from CiFAKE, Columbia, and JSSSTU datasets was carefully curated to maintain class balance. Despite the initial variability in image counts across datasets due to the abovementioned challenges, a balanced dataset was assembled with 1500 images per class (CGI and authentic). This approach allowed for a comprehensive evaluation of the model's performance across diverse data sources, ensuring the reliability and generalizability of the results.

\subsection{Simulation Results}

The analysis of the results obtained from the Swin Transformer model, trained on the CiFAKE, Columbia, and JSSSTU datasets, reveals varying degrees of performance across these datasets. The training and validation accuracy curves, loss plots, and ROC curves provide insight into the model's effectiveness in distinguishing between CGI and authentic images. Each dataset presents unique challenges reflected in the model's performance metrics, such as accuracy, loss, and ROC AUC scores. The results demonstrate how the quality and quantity of data influence the model's learning process and its ability to generalize.

\subsubsection{Performance on CiFAKE Dataset}

\begin{figure}[!htbp]
	\centering
	\subfigure[\label{fig7a}]{\includegraphics[width=3cm,height=3cm]{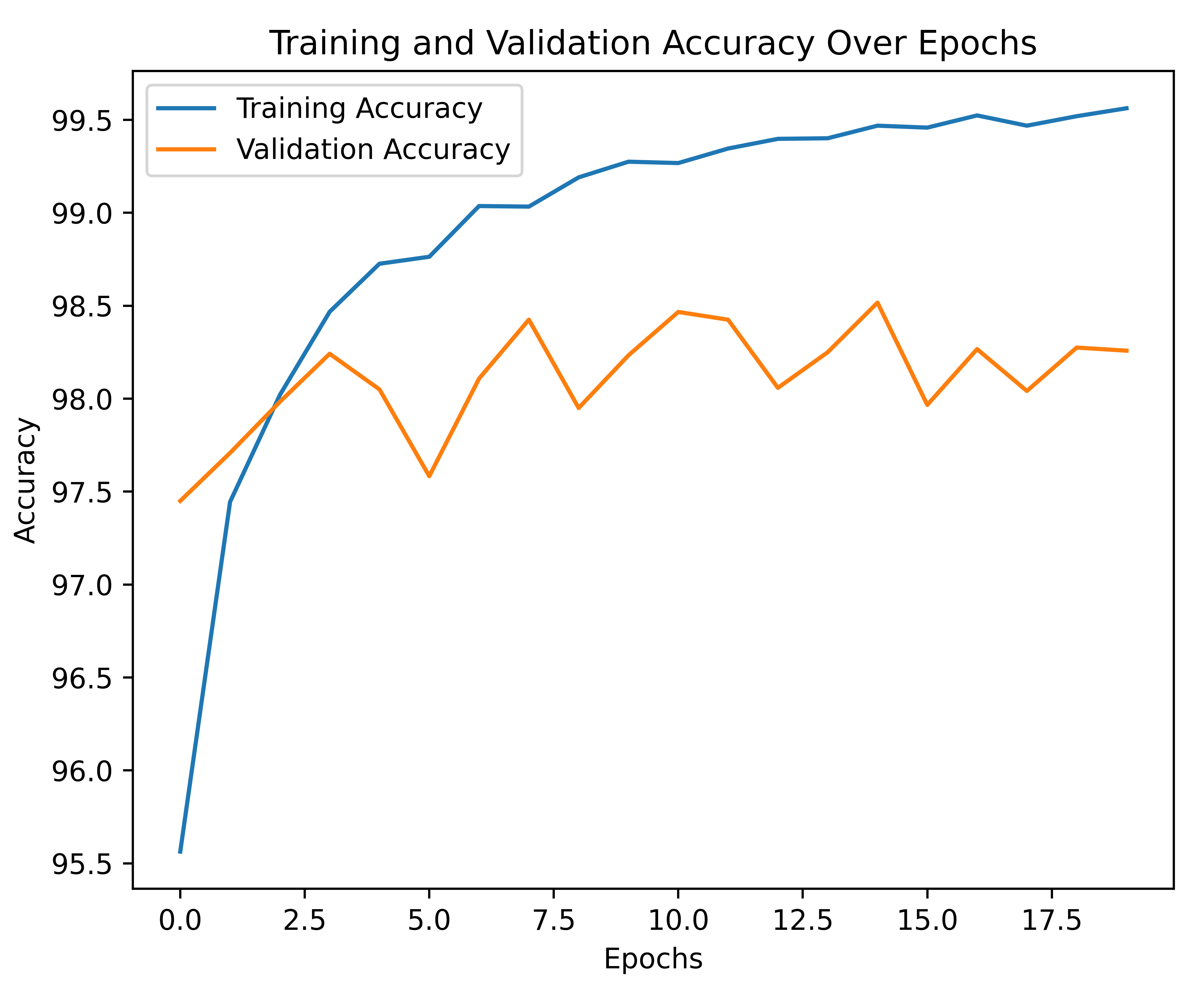}}
	\subfigure[\label{fig7b}]{\includegraphics[width=3cm,height=3cm]{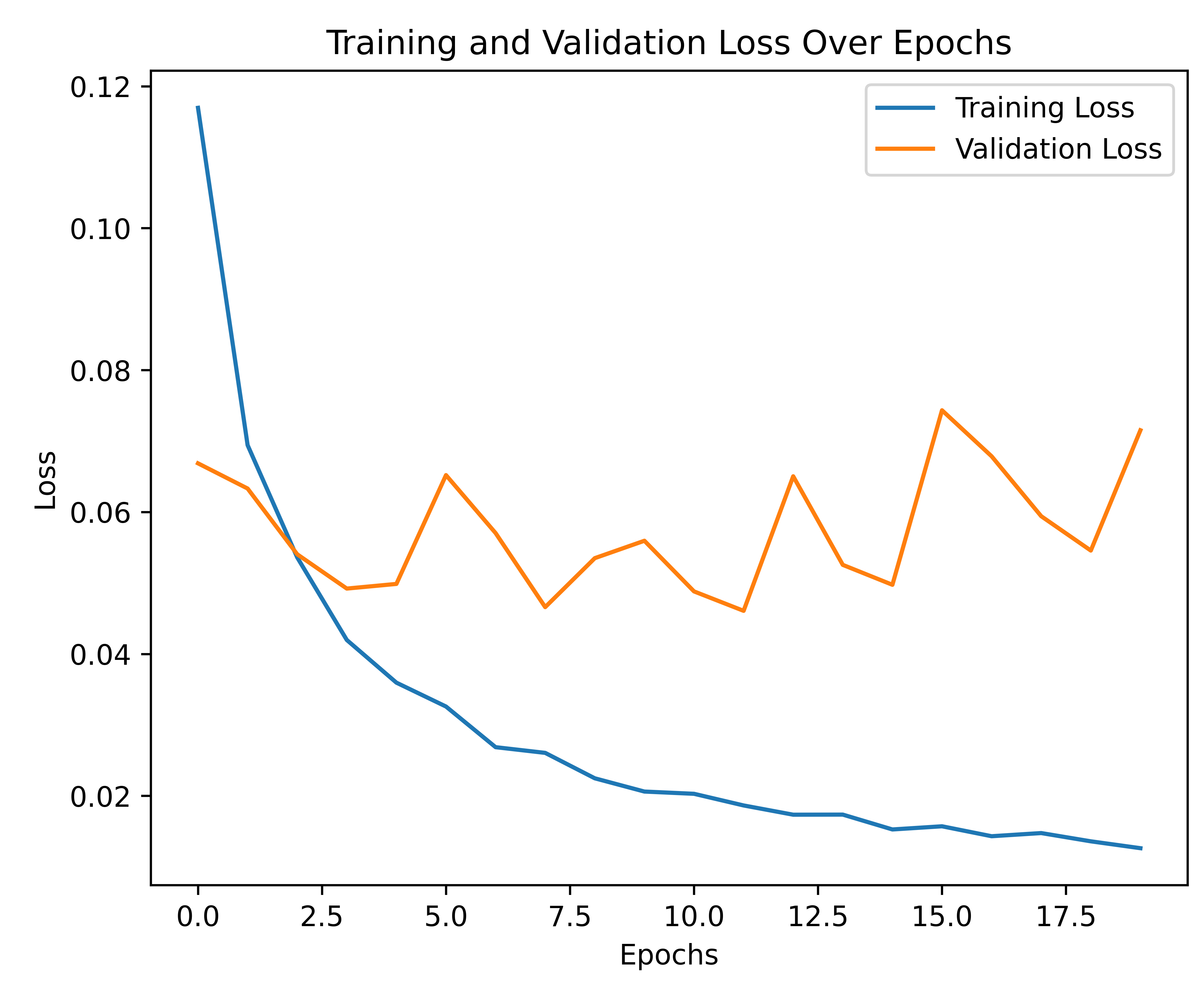}}
	\subfigure[\label{fig7c}]{\includegraphics[width=3cm,height=3cm]{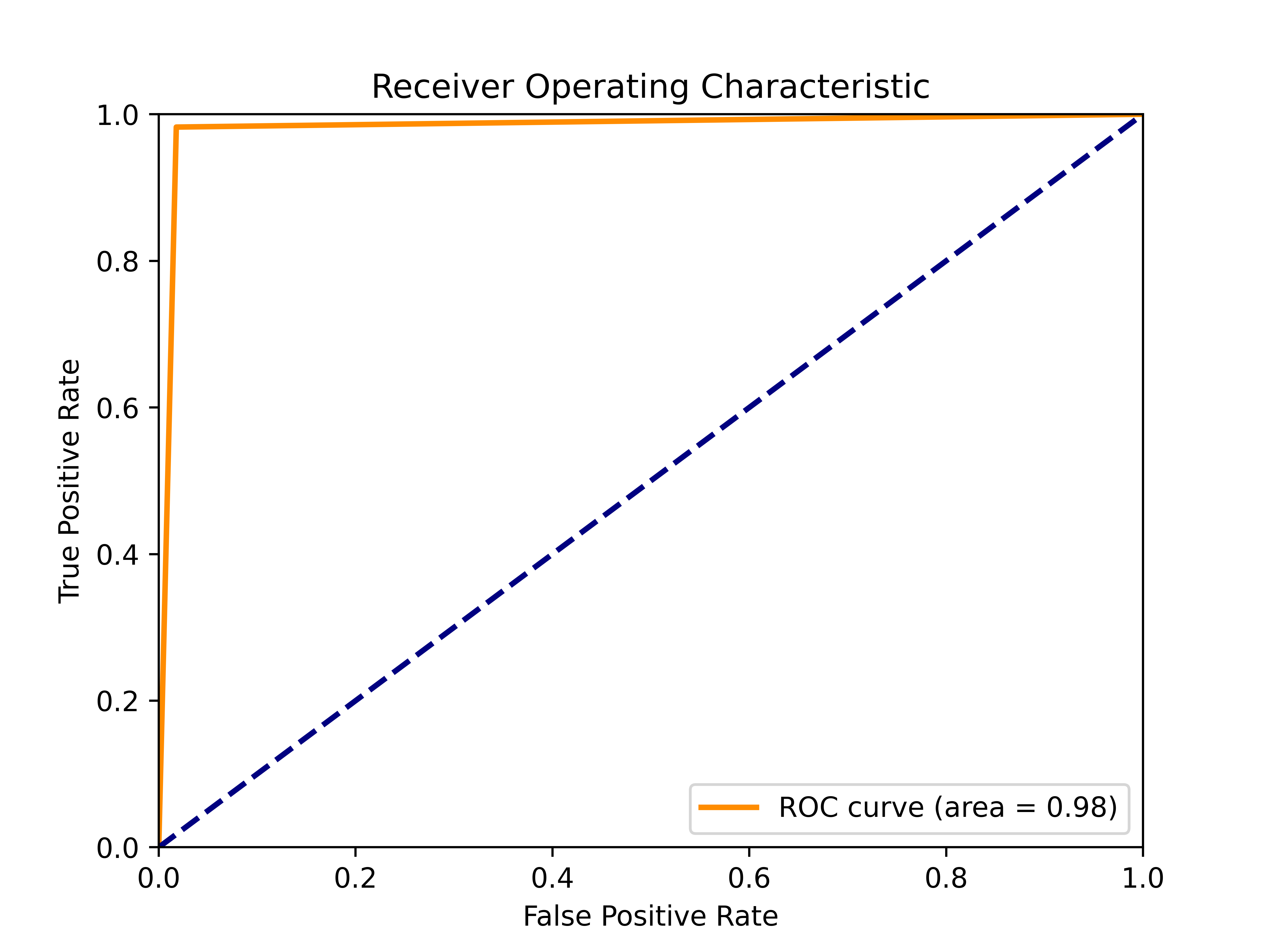}}\\
	
	\subfigure[\label{fig7d}]{\includegraphics[width=3cm,height=3cm]{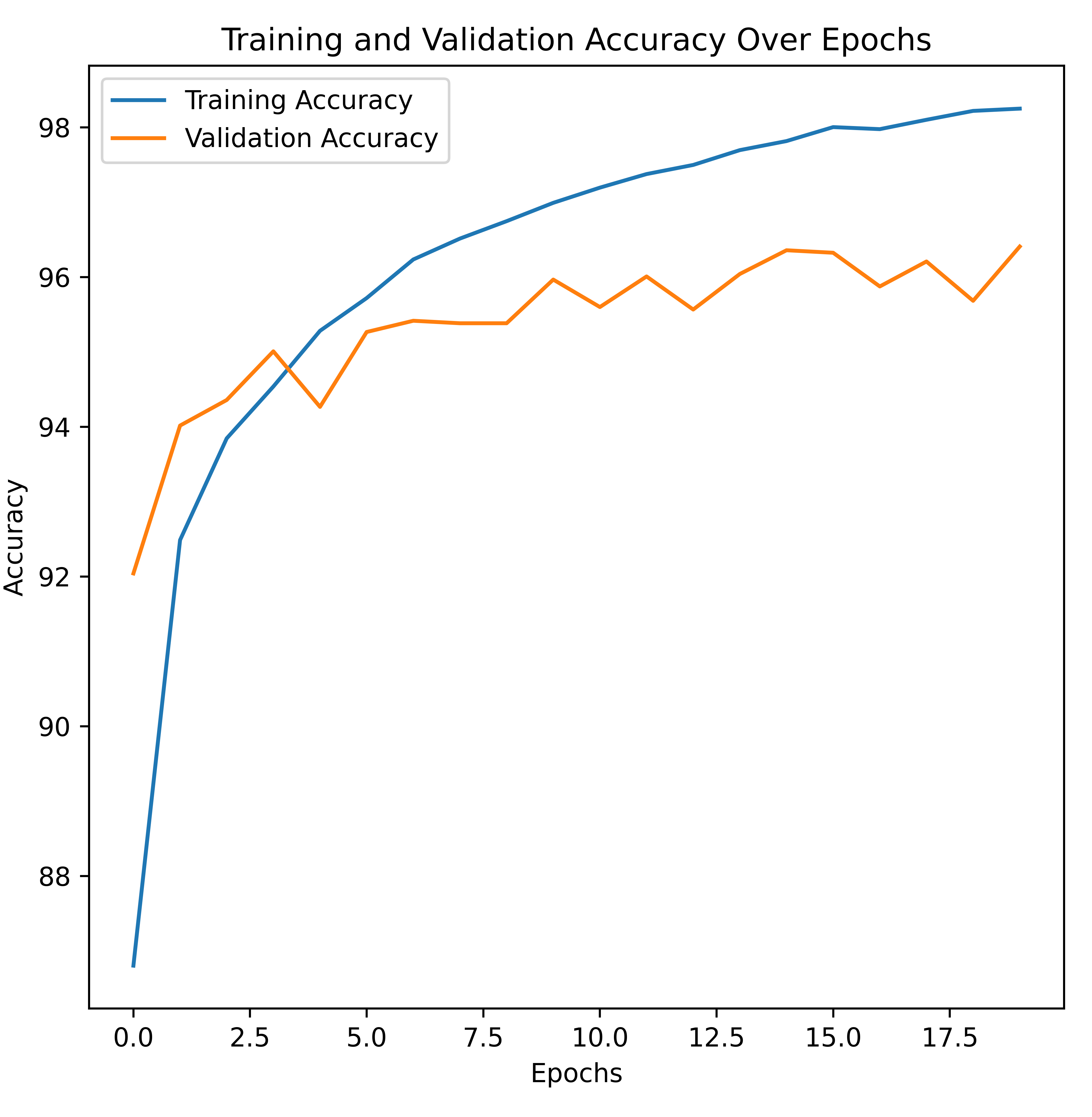}}
	\subfigure[\label{fig7e}]{\includegraphics[width=3cm,height=3cm]{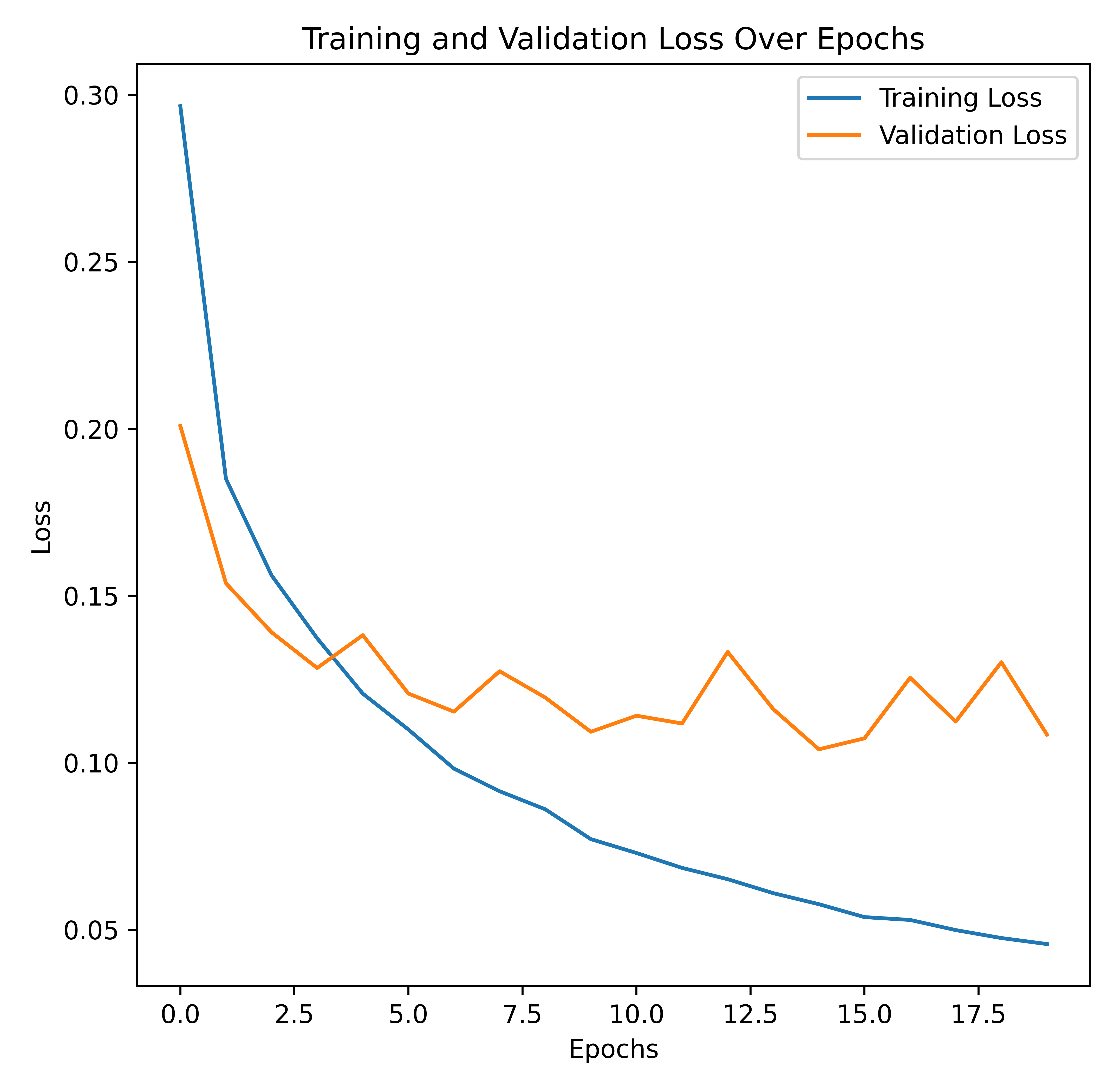}}
	\subfigure[\label{fig7f}]{\includegraphics[width=3cm,height=3cm]{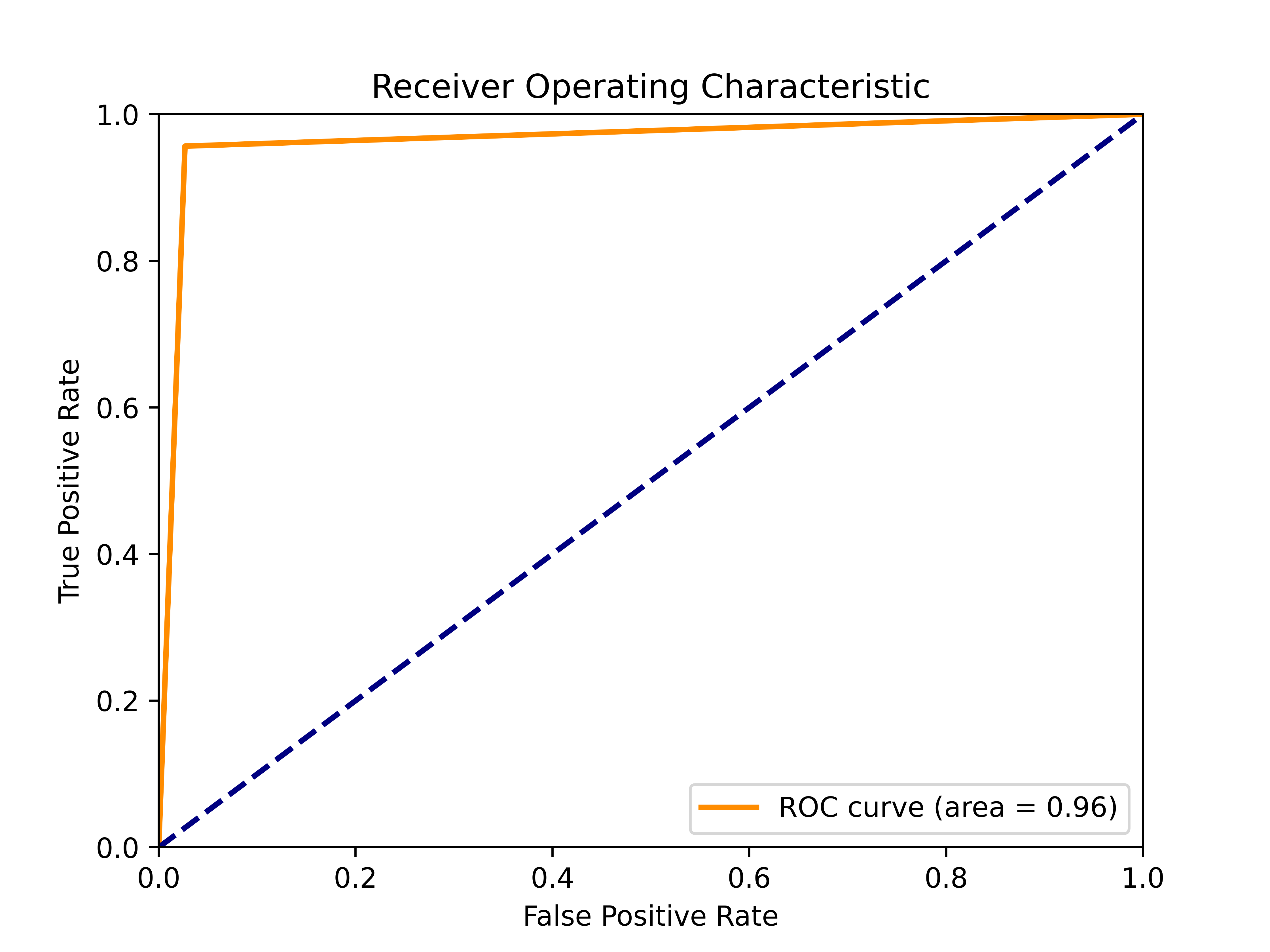}}\\
	
	\subfigure[\label{fig7g}]{\includegraphics[width=3cm,height=3cm]{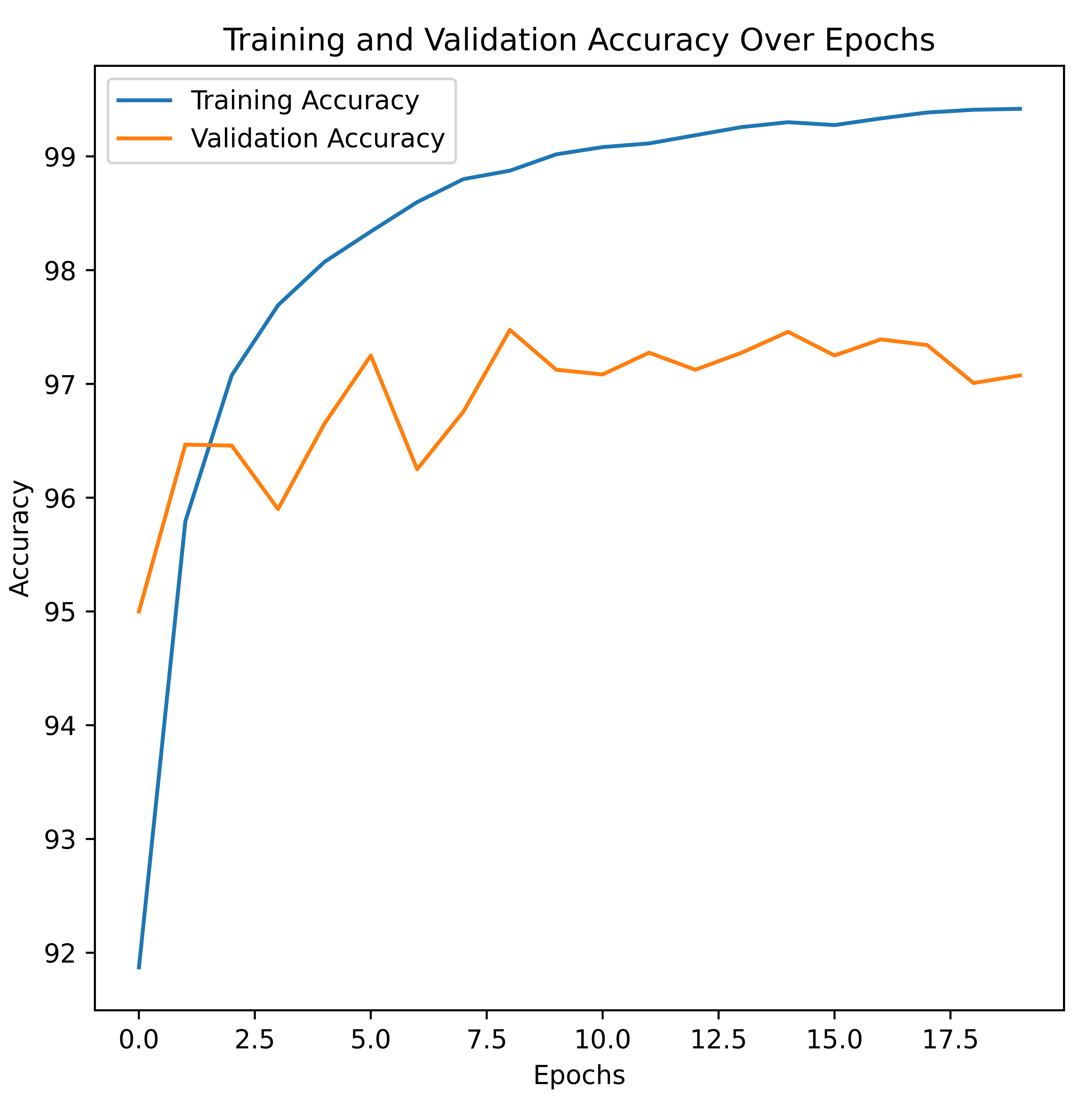}}
	\subfigure[\label{fig7h}]{\includegraphics[width=3cm,height=3cm]{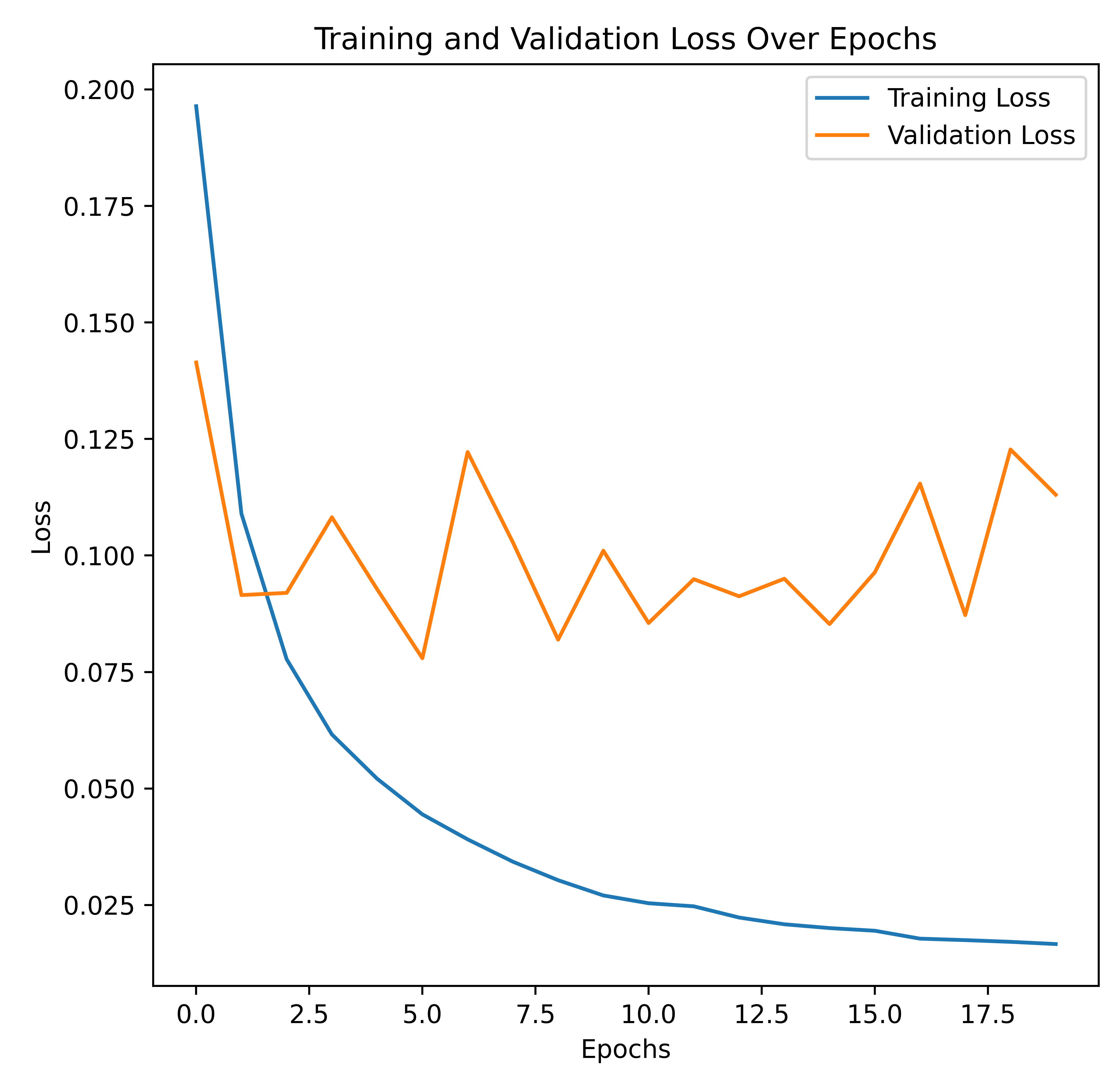}}
	\subfigure[\label{fig7i}]{\includegraphics[width=3cm,height=3cm]{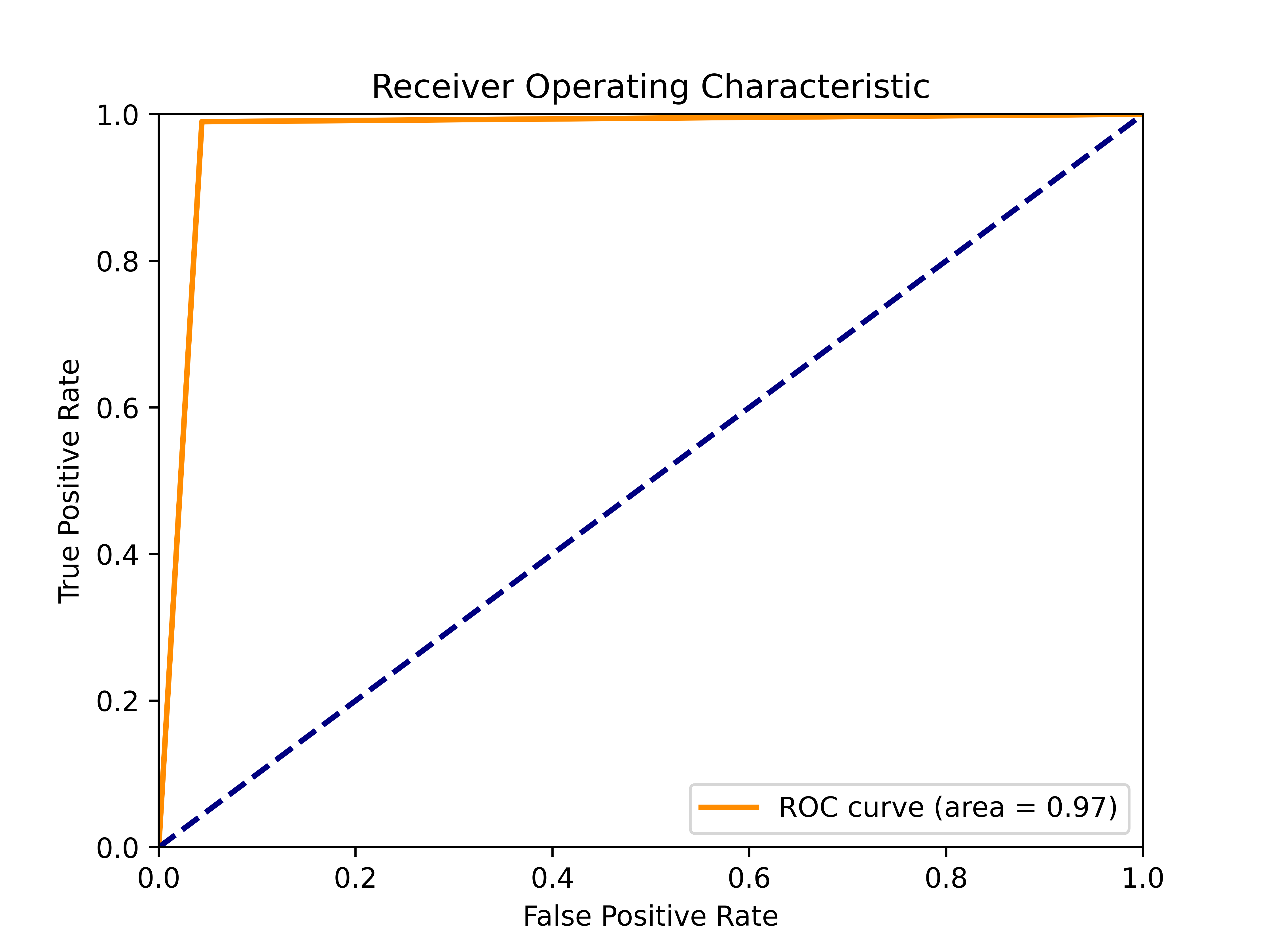}}
	
	\caption{The training and validation accuracy plot, loss plot and ROC (from left to right) for the CiFAKE dataset for the RGB, YCbCr and HSV color frames (top to bottom).}
	\label{fig7}
	
\end{figure}

\begin{table}[!htbp]
	\centering
	\caption{The evaluation parameters for the CiFake dataset for the different color spaces.}
	\label{table 4}
	\begin{tabular}{|l|c|c|c|c|c|}
		\hline
		\textbf{Color Space} & \textbf{Accuracy} &\textbf{Precision} &\textbf{Recall} & \textbf{F1-score} &\textbf{AUC}\\ \hline
		\textbf{RGB} &0.98 &0.98 &0.98 &0.98 &0.98\\
		\textbf{YCbCr} &0.97 &0.97 &0.96 &0.96 &0.96\\
		\textbf{HSV} &0.97 &0.96 &0.99 &0.97 &0.97\\
		\hline
	\end{tabular}
\end{table}

The performance analysis of the CiFAKE dataset for RGB, YCbCr, and HSV color spaces is illustrated in Fig. \ref{fig7} and summarized in Table \ref{table 4}. The training and validation accuracy plots (Figs. \ref{fig7a},\ref{fig7d},\ref{fig7g}) demonstrate consistently high convergence across all color spaces, with RGB achieving the fastest convergence and the highest stability. The loss plots (Figs. \ref{fig7b},\ref{fig7e},\ref{fig7h}) highlight that RGB exhibits the lowest loss values, further validating its superior performance. The ROC curves (Figs. \ref{fig7c},\ref{fig7f},\ref{fig7i}) show that RGB achieves an AUC of 0.98, indicating excellent classification capability, while YCbCr and HSV both attain slightly lower but comparable AUC values of 0.97. The quantitative evaluation in Table \ref{table 4} reinforces these findings, where RGB achieves the highest precision, recall, F1-score, and accuracy of 0.98 across all metrics. While YCbCr and HSV are effective alternatives with metrics closely aligned, the RGB color space consistently outperforms the others, establishing itself as the most effective for this dataset in terms of both feature separability and classification performance.

\subsubsection{Performance on JSSSTU Dataset}

\begin{figure}[!htbp]
	\centering
	\subfigure[\label{fig8a}]{\includegraphics[width=3cm,height=3cm]{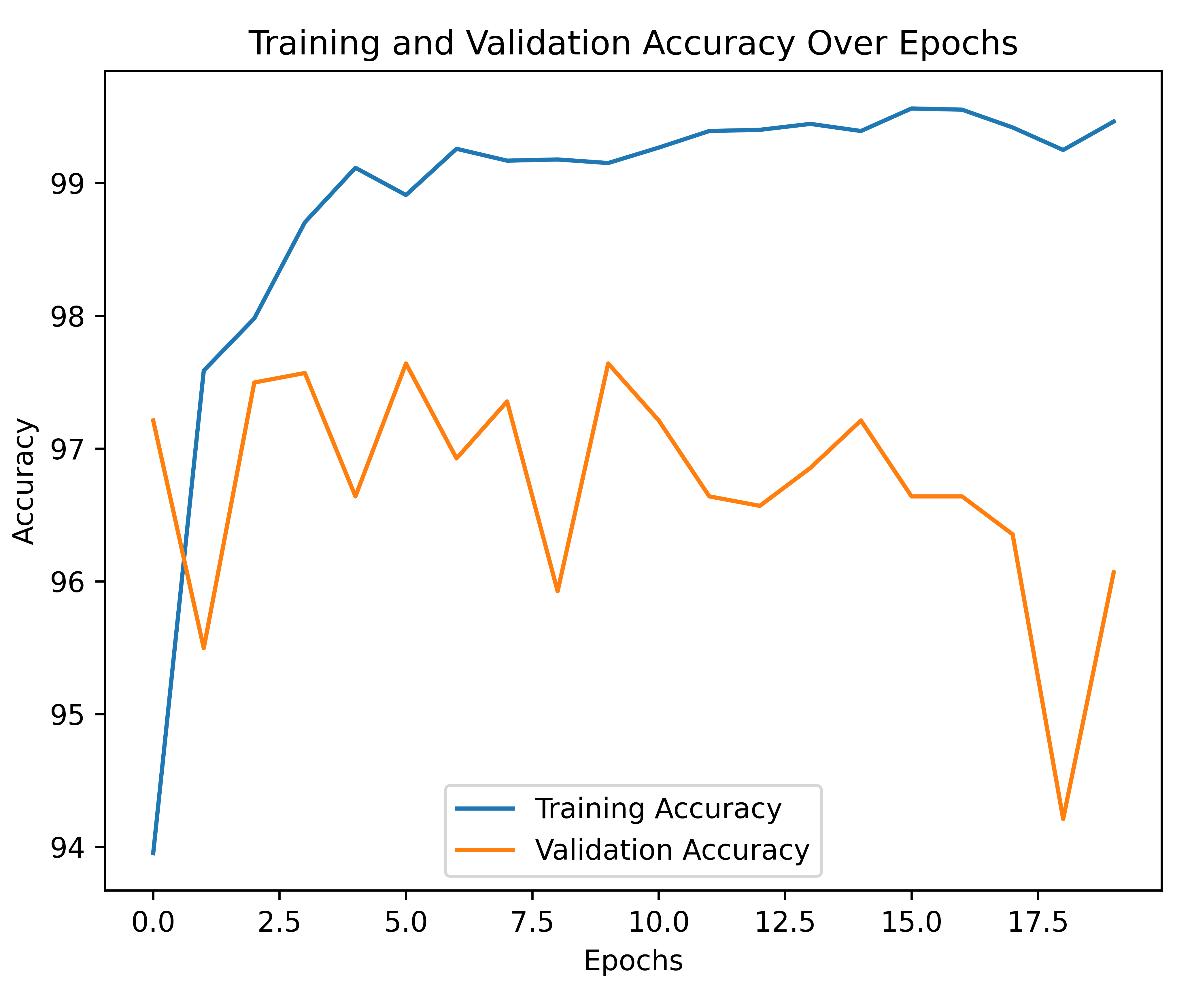}}
	\subfigure[\label{fig8b}]{\includegraphics[width=3cm,height=3cm]{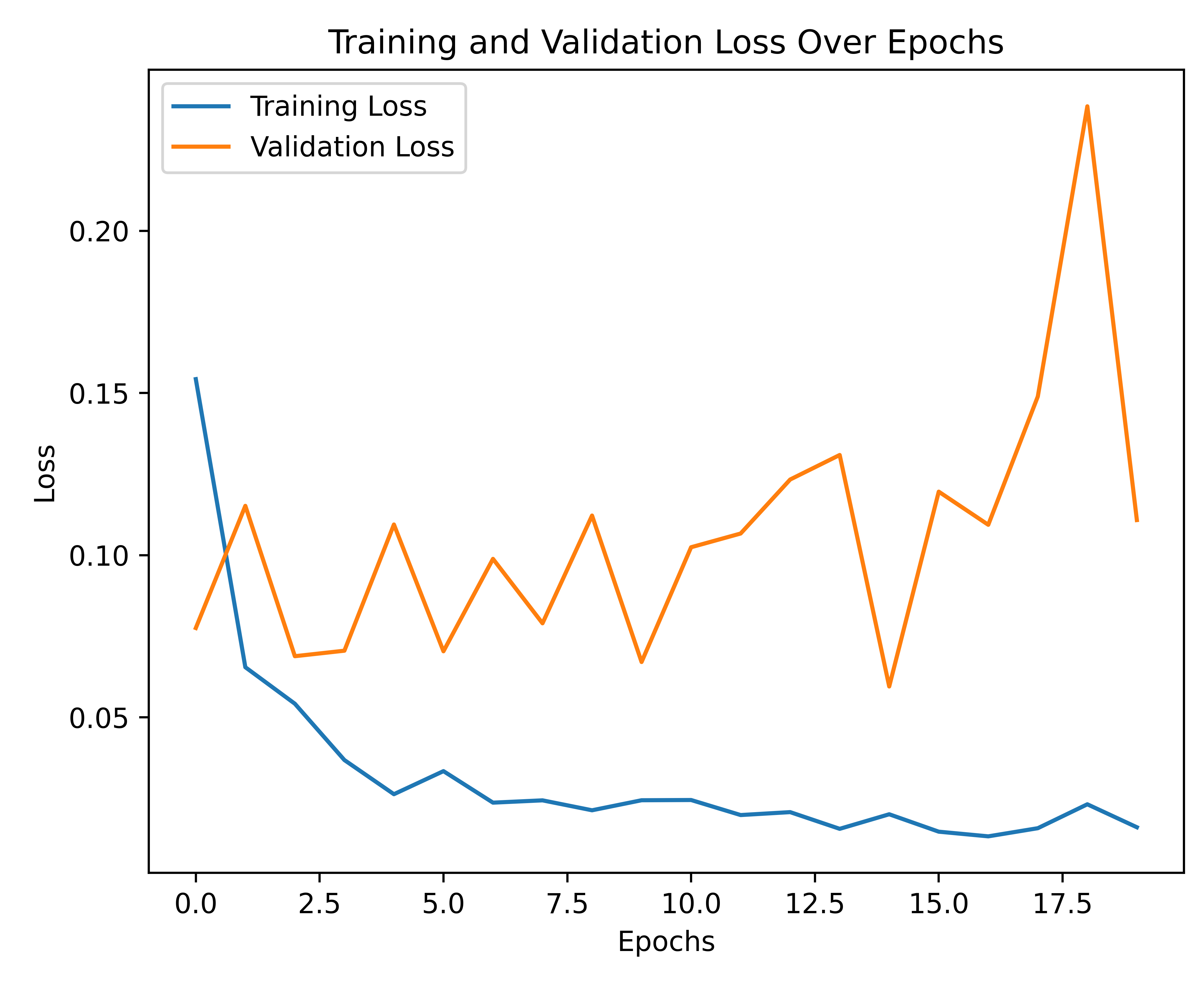}}
	\subfigure[\label{fig8c}]{\includegraphics[width=3cm,height=3cm]{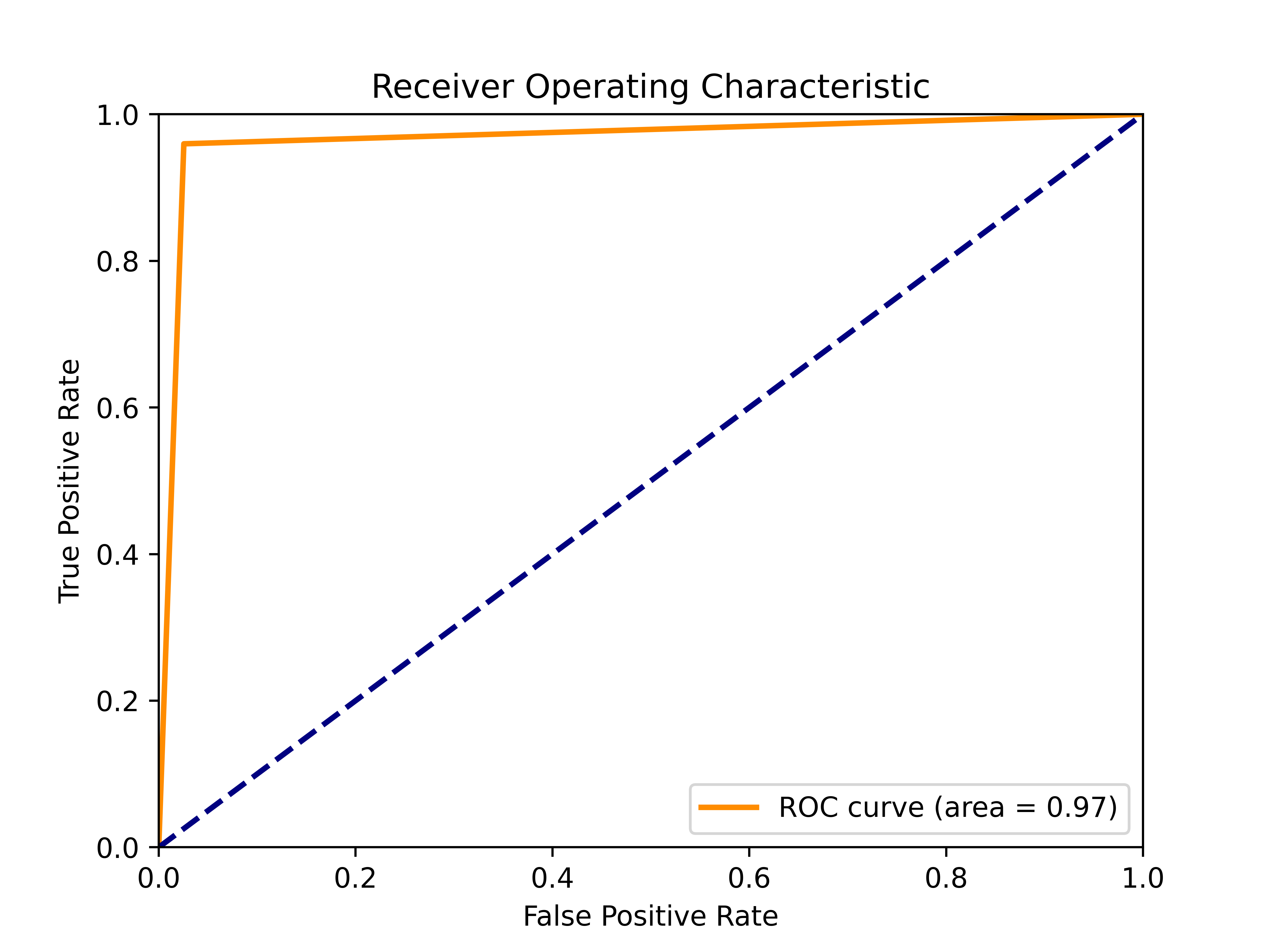}}\\
	
	\subfigure[\label{fig8d}]{\includegraphics[width=3cm,height=3cm]{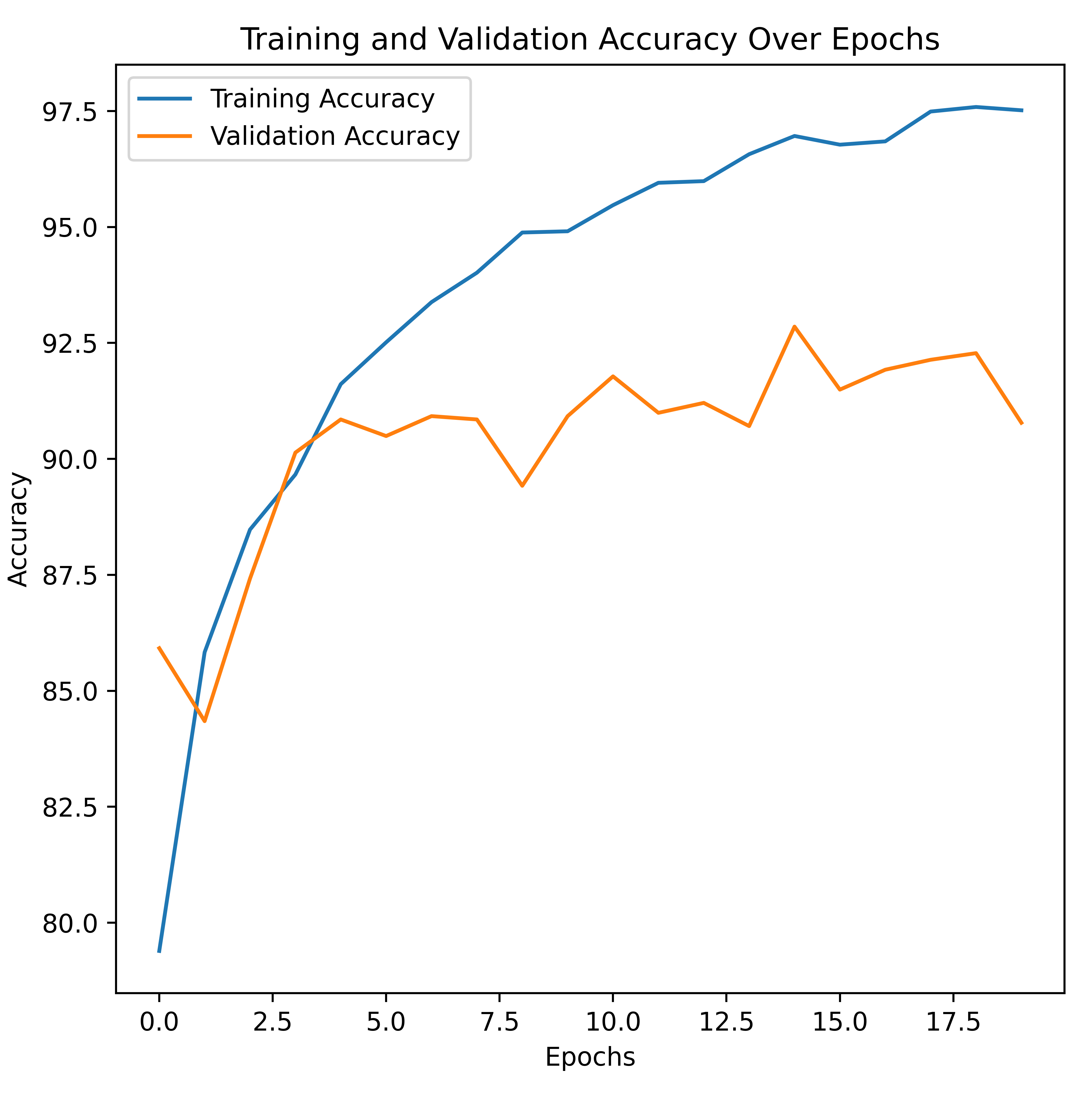}}
	\subfigure[\label{fig8e}]{\includegraphics[width=3cm,height=3cm]{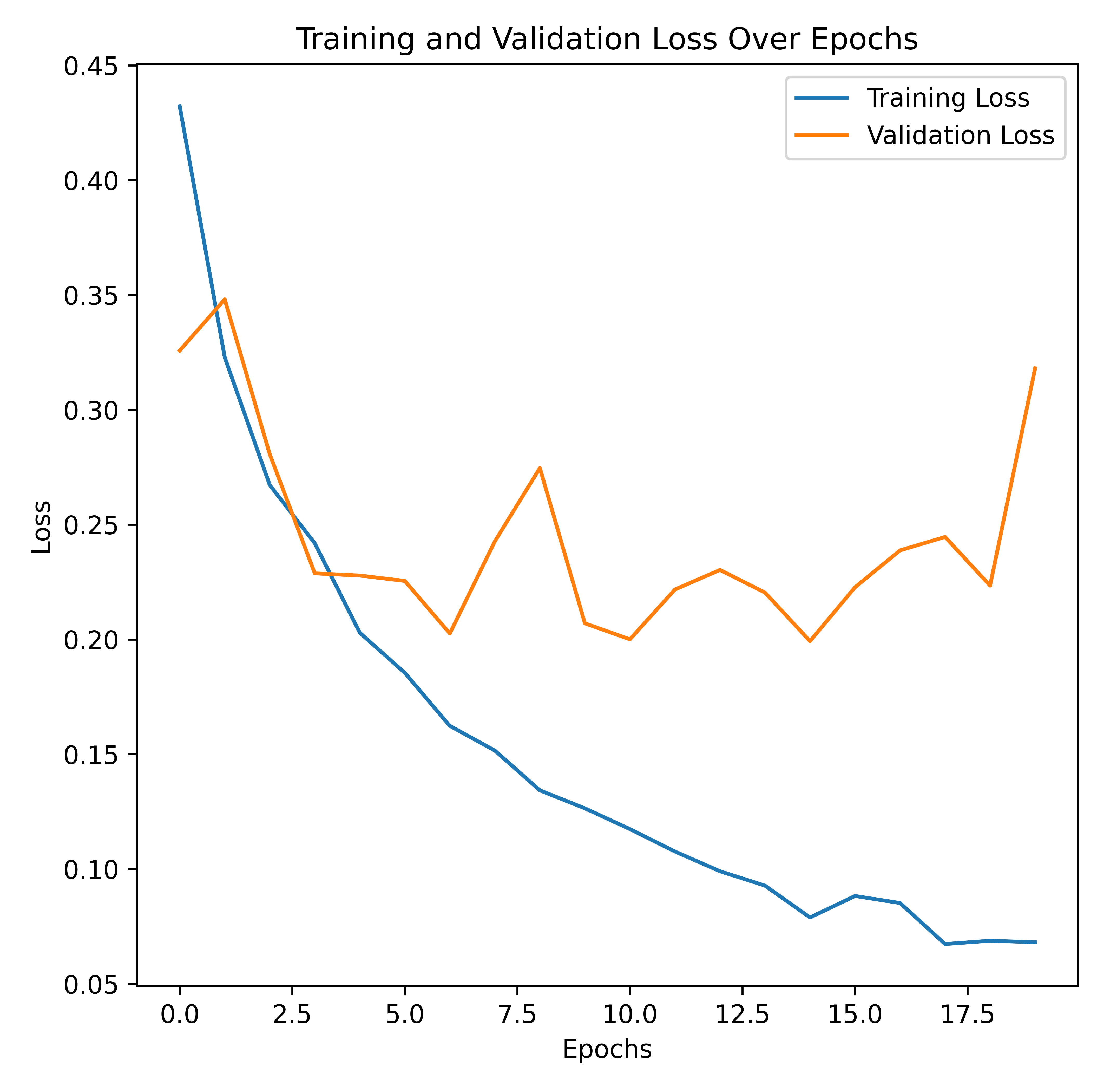}}
	\subfigure[\label{fig8f}]{\includegraphics[width=3cm,height=3cm]{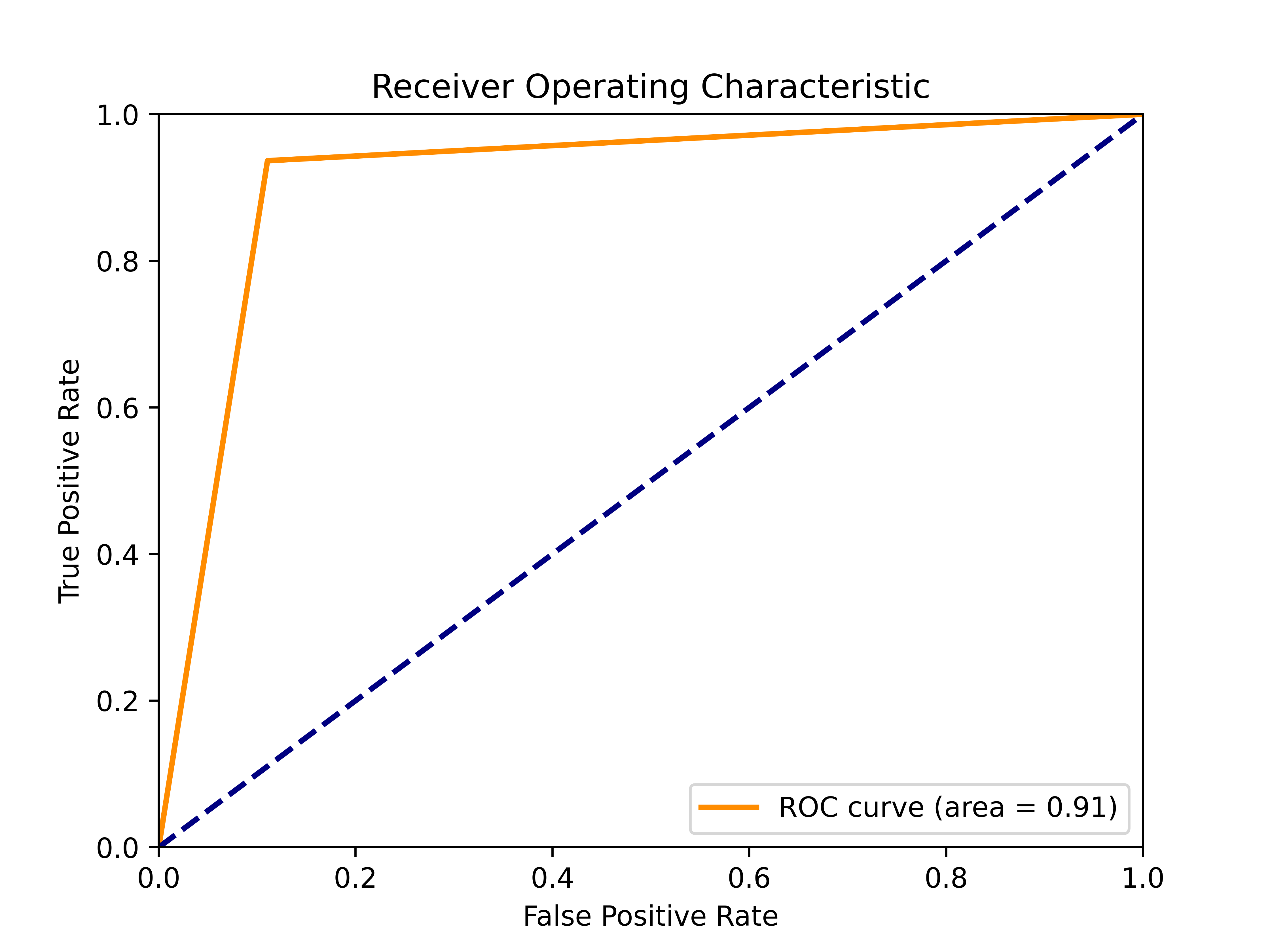}}\\
	
	\subfigure[\label{fig8g}]{\includegraphics[width=3cm,height=3cm]{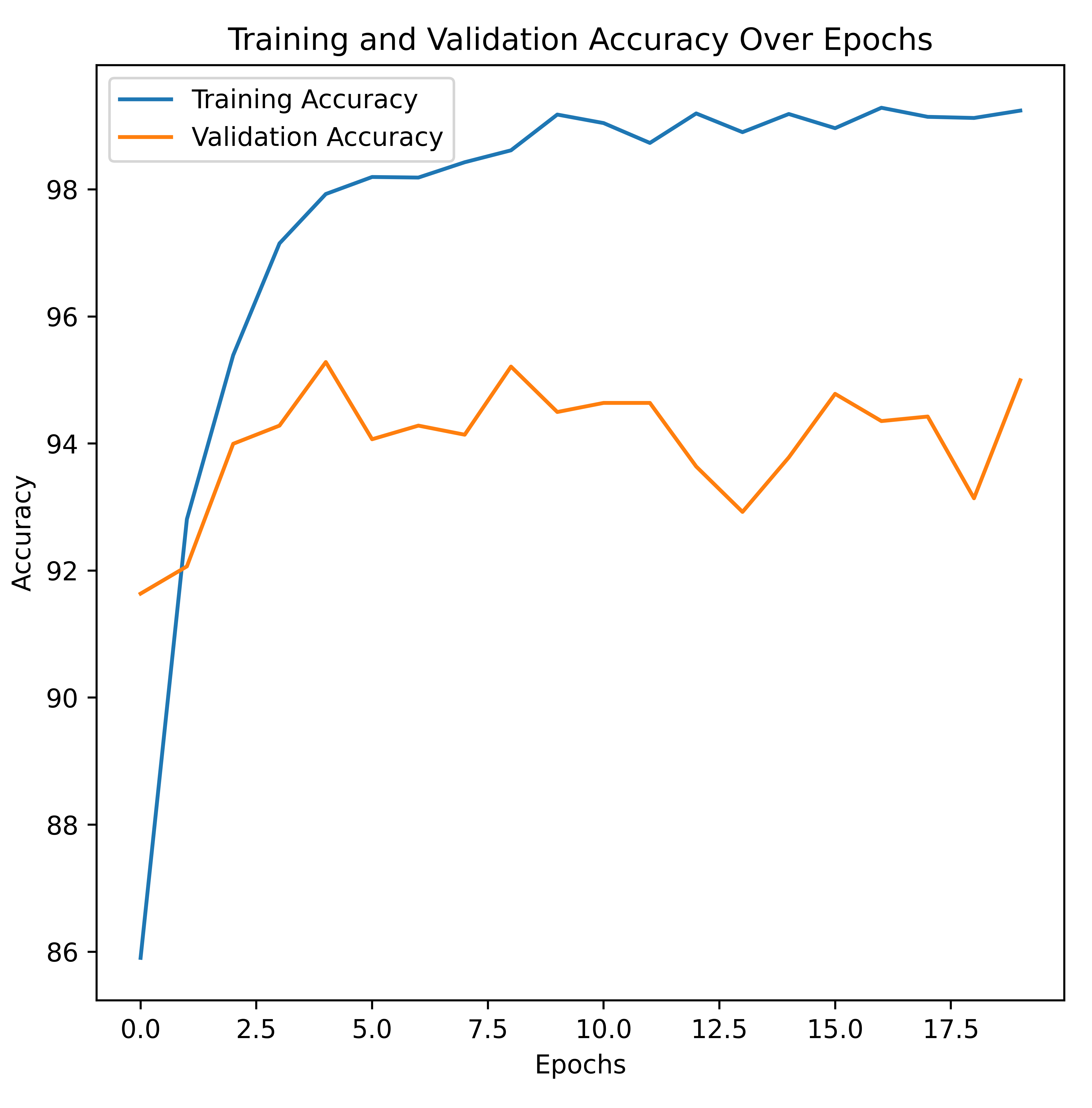}}
	\subfigure[\label{fig8h}]{\includegraphics[width=3cm,height=3cm]{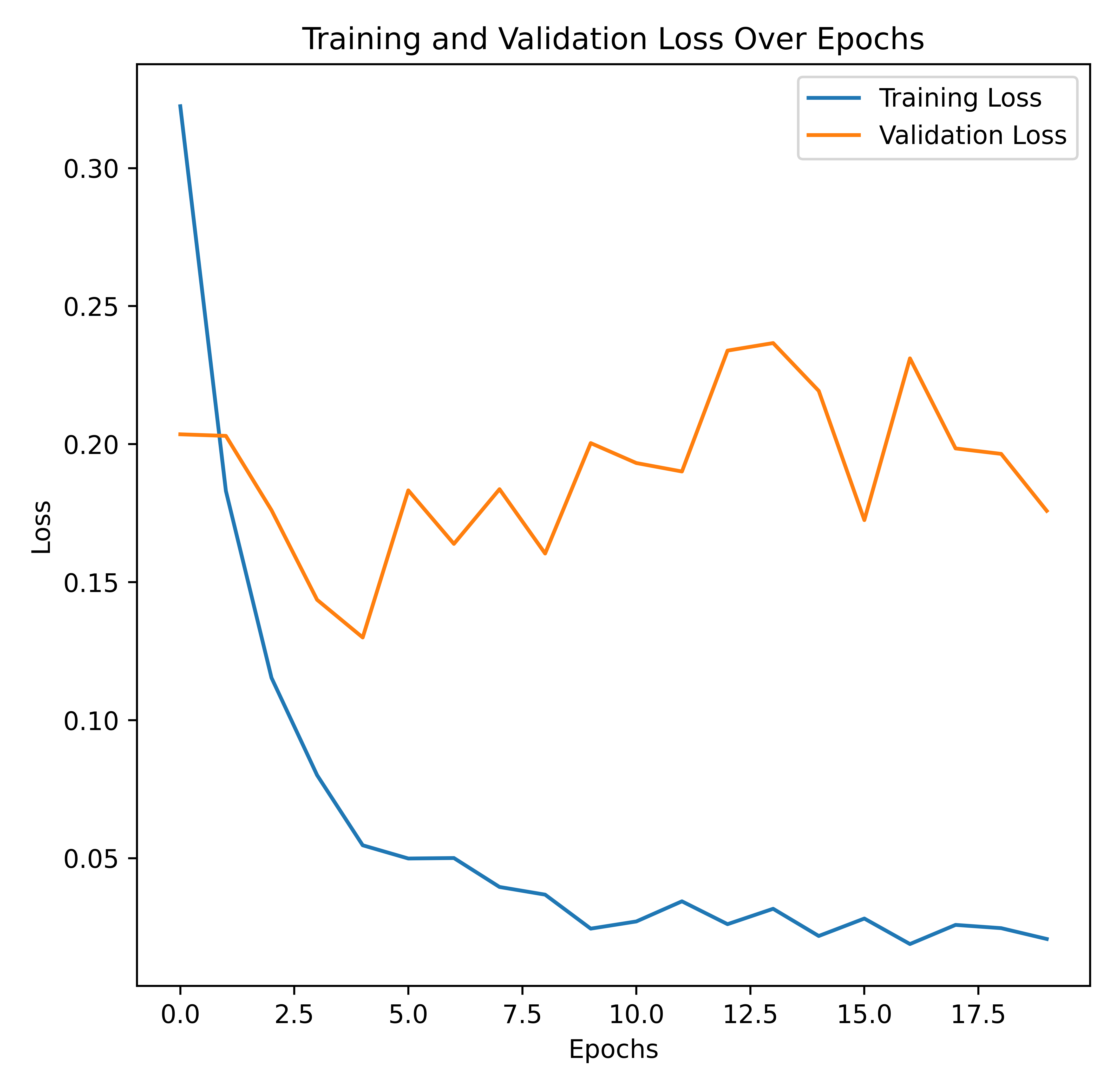}}
	\subfigure[\label{fig8i}]{\includegraphics[width=3cm,height=3cm]{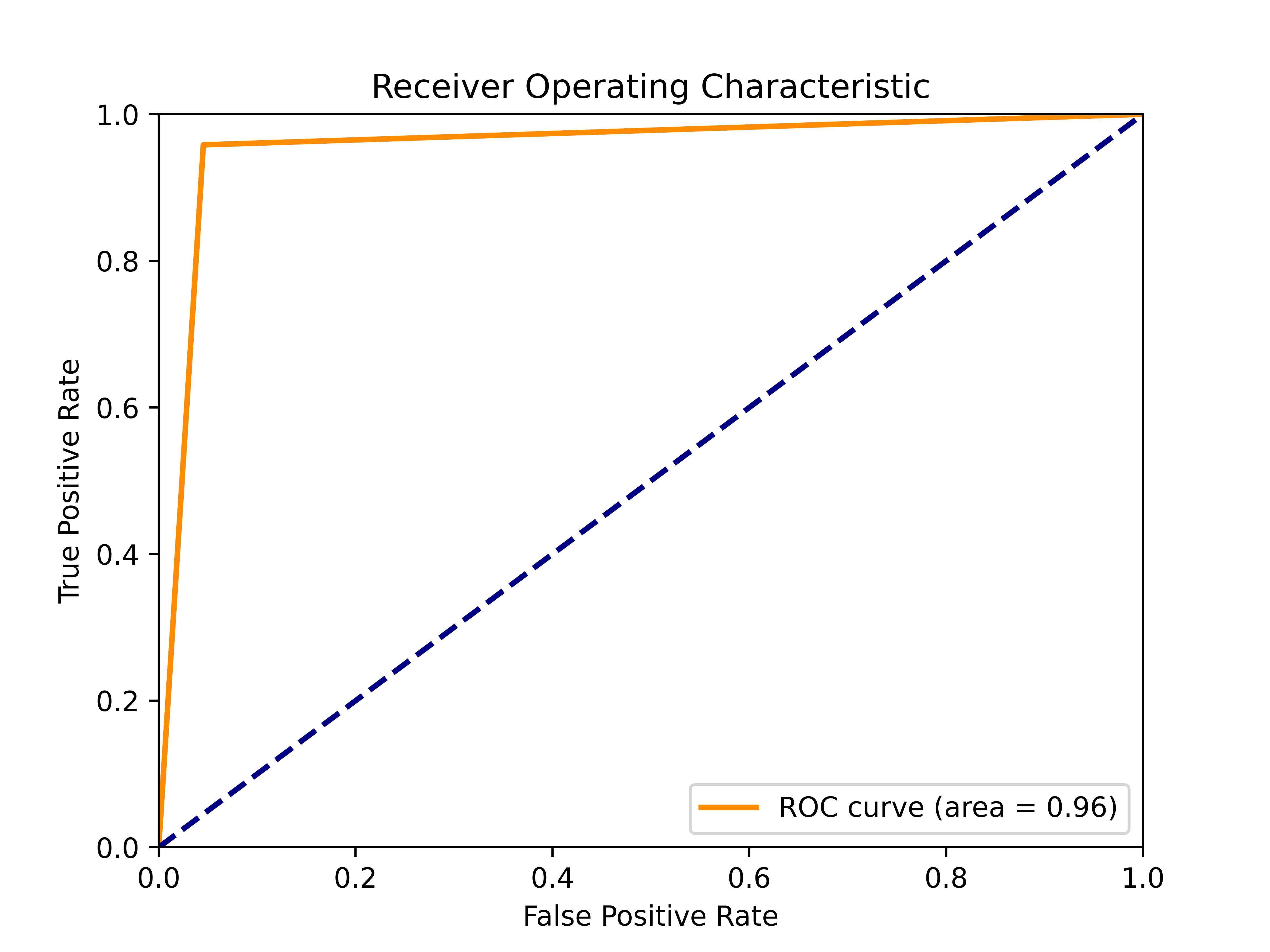}}
	
	\caption{The training and validation accuracy plot, loss plot and ROC (from left to right) for the JSSSTU dataset for the RGB, YCbCr and HSV color frames (top to bottom).}
	\label{fig8}
	
\end{figure}

\begin{table}[!htbp]
	\centering
	\caption{The evaluation parameters for the JSSSTU dataset for the different color spaces.}
	\label{table 5}
	\begin{tabular}{|l|c|c|c|c|c|}
		\hline
		\textbf{Color Space} & \textbf{Accuracy} &\textbf{Precision} &\textbf{Recall} & \textbf{F1-score} &\textbf{AUC}\\ \hline
		\textbf{RGB} &0.97 &0.97 &0.96 &0.97 &0.97\\
		\textbf{YCbCr} &0.91 &0.89 &0.94 &0.91 &0.91\\
		\textbf{HSV} &0.96 &0.95 &0.96 &0.96 &0.96\\
		\hline
	\end{tabular}
\end{table}

The performance analysis of the CiFAKE dataset for RGB, YCbCr, and HSV color spaces is illustrated in Fig. \ref{fig8} and summarized in Table \ref{table 5}. For the JSSSTU dataset, RGB color space outperforms others, achieving the highest accuracy (97\%), F1-score (0.97), and AUC (0.97), as reflected in its superior training and validation convergence in the plots. The HSV color space follows closely with an accuracy of 96\% and comparable F1-score (0.96) and AUC (0.96), making it a viable alternative for efficient color representation. However, the YCbCr color space lags behind, with a lower accuracy (91\%), F1-score (0.91), and AUC (0.91), despite a high recall (0.94), indicating challenges with false positives.

\subsubsection{Challenges with the Columbia Dataset}

\begin{figure}[!htbp]
	\centering
	\subfigure[\label{fig9a}]{\includegraphics[width=3cm,height=3cm]{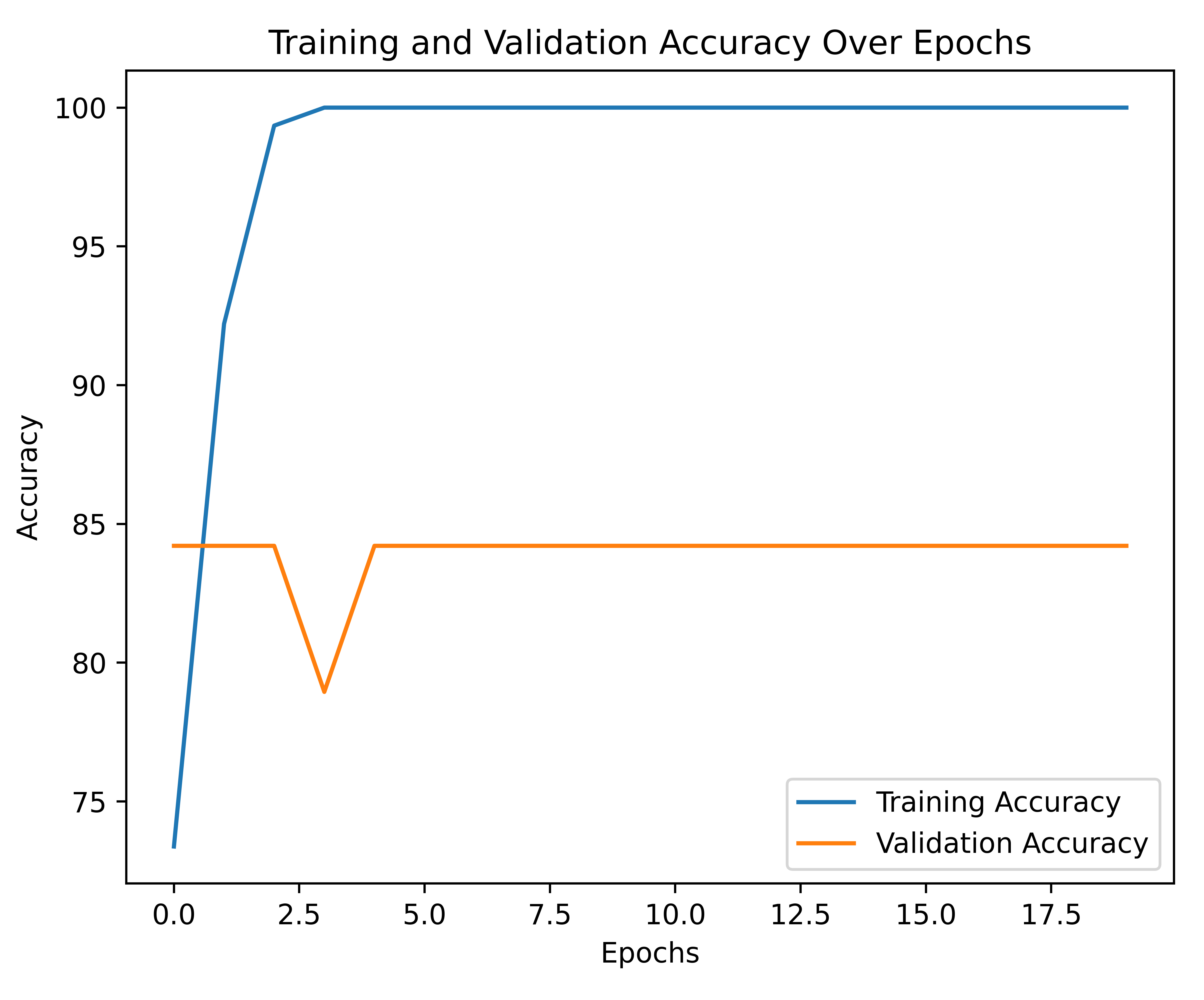}}
	\subfigure[\label{fig9b}]{\includegraphics[width=3cm,height=3cm]{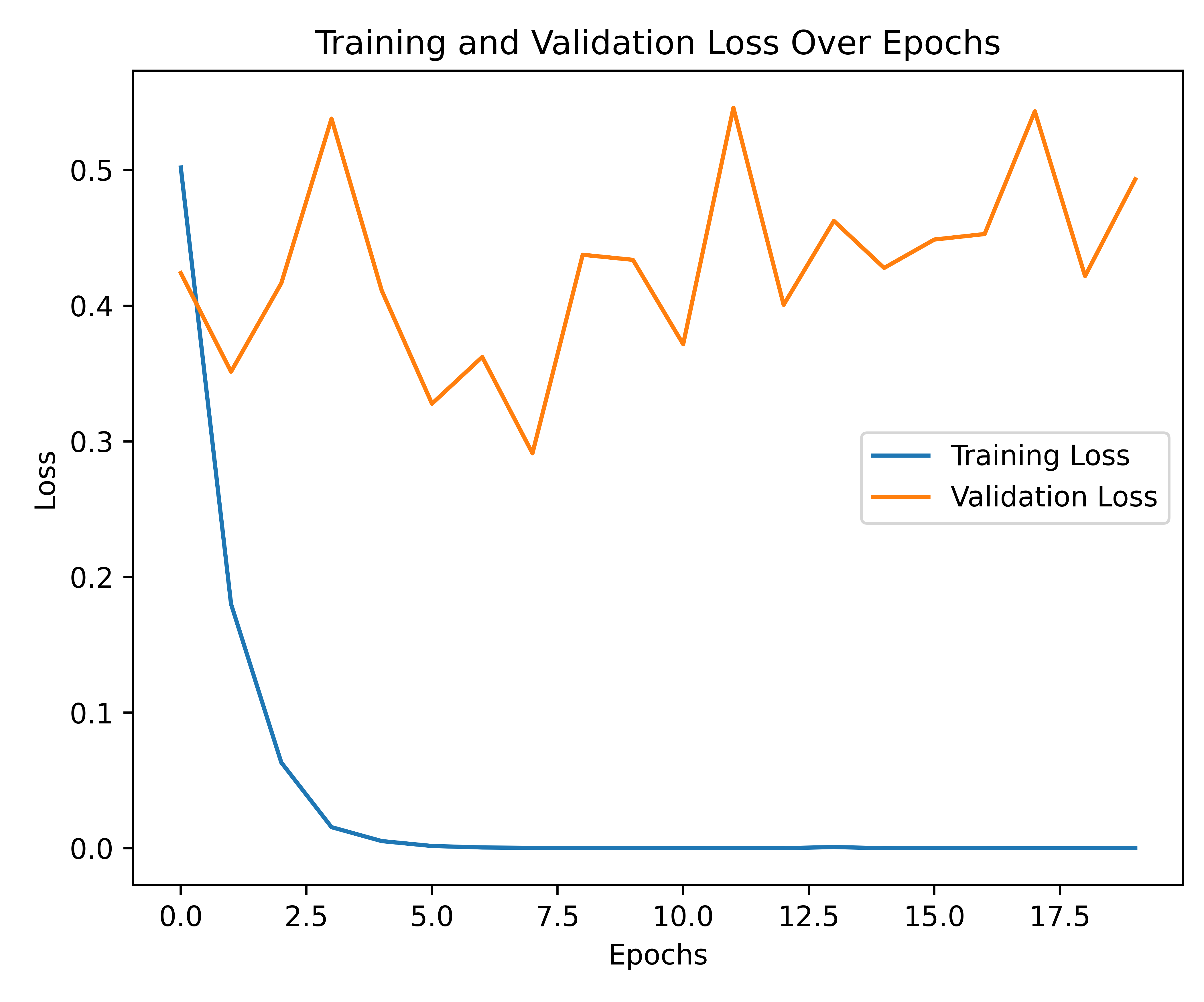}}
	\subfigure[\label{fig9c}]{\includegraphics[width=3cm,height=3cm]{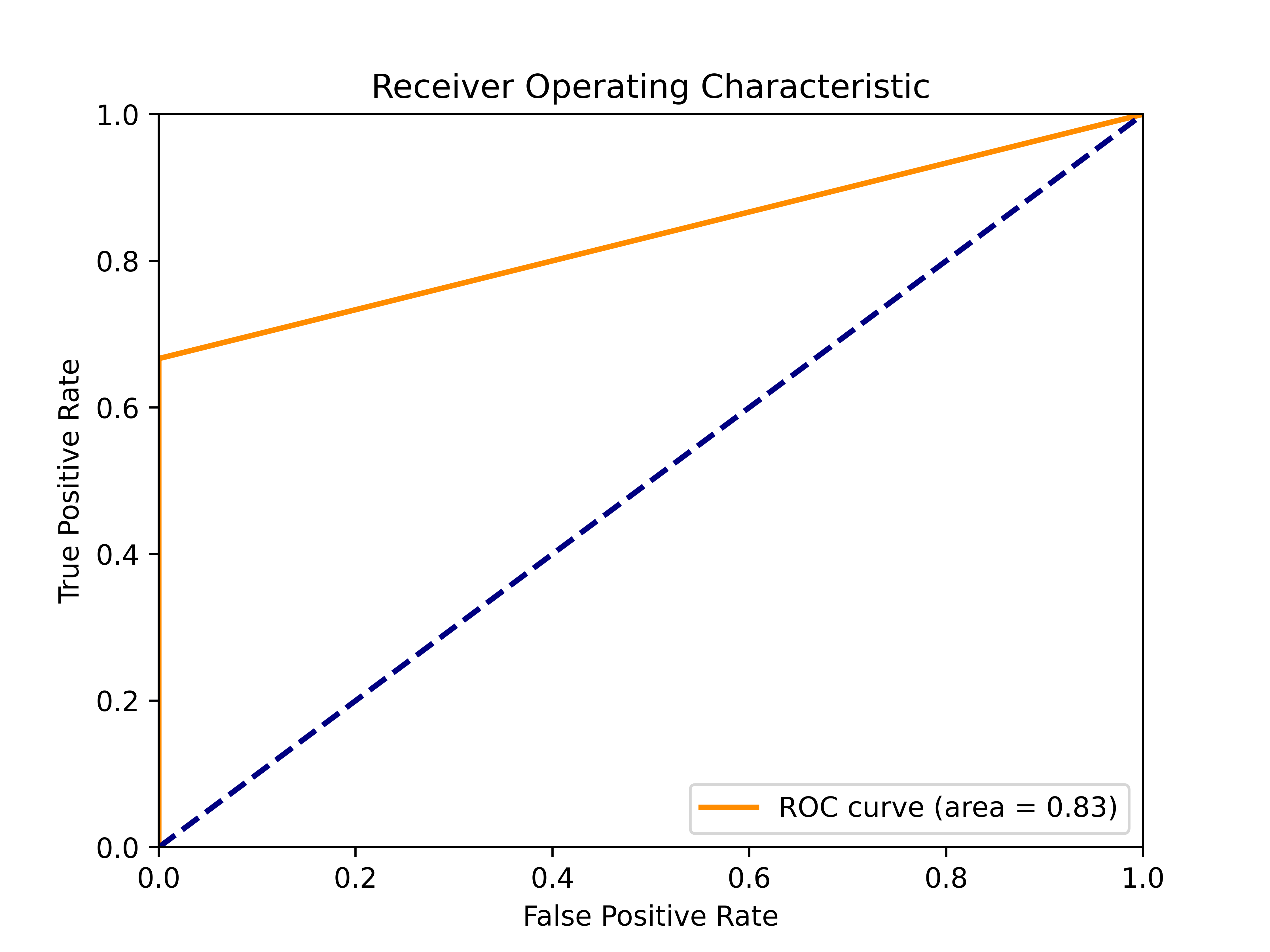}}\\
	
	\subfigure[\label{fig9d}]{\includegraphics[width=3cm,height=3cm]{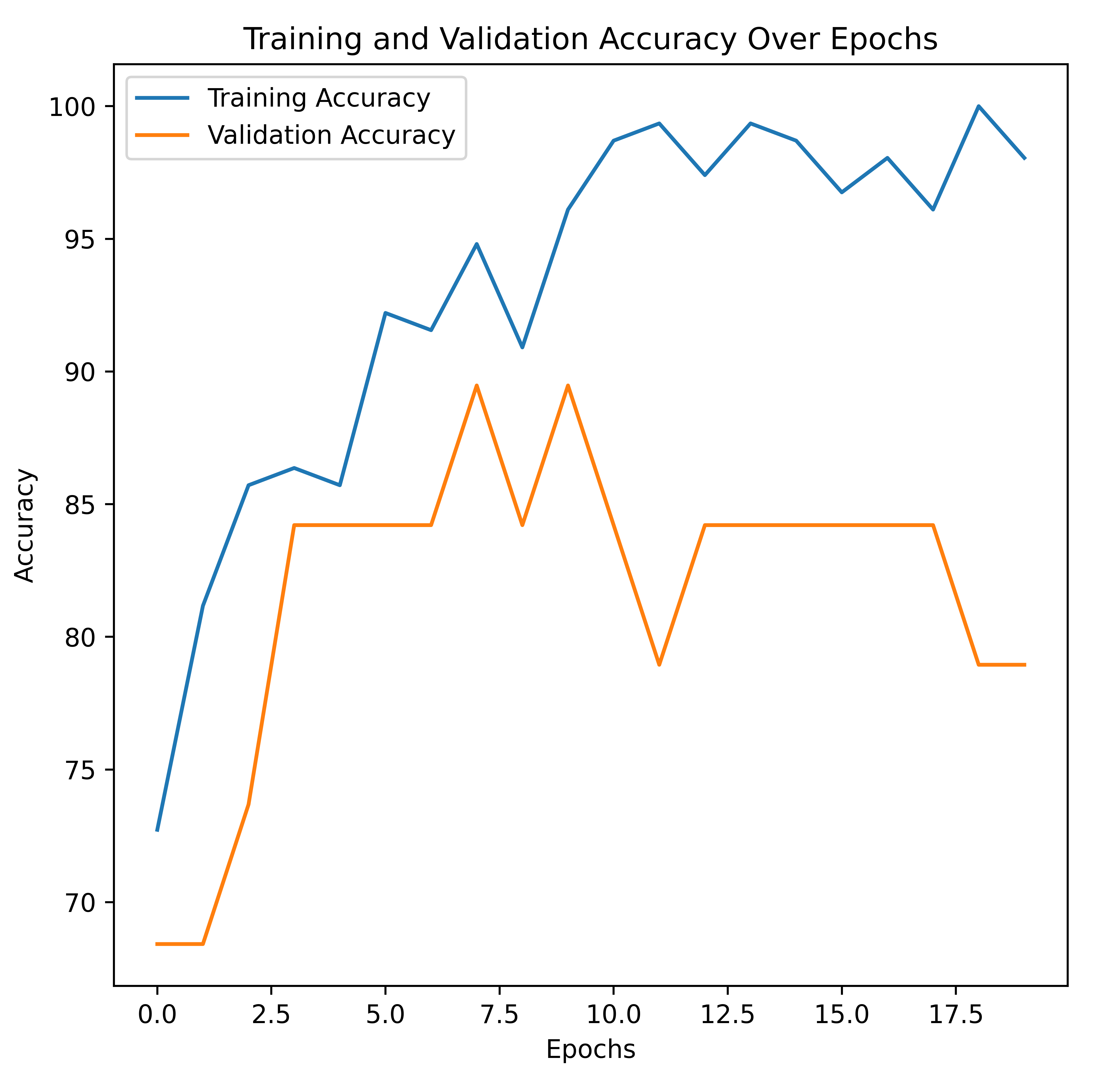}}
	\subfigure[\label{fig9e}]{\includegraphics[width=3cm,height=3cm]{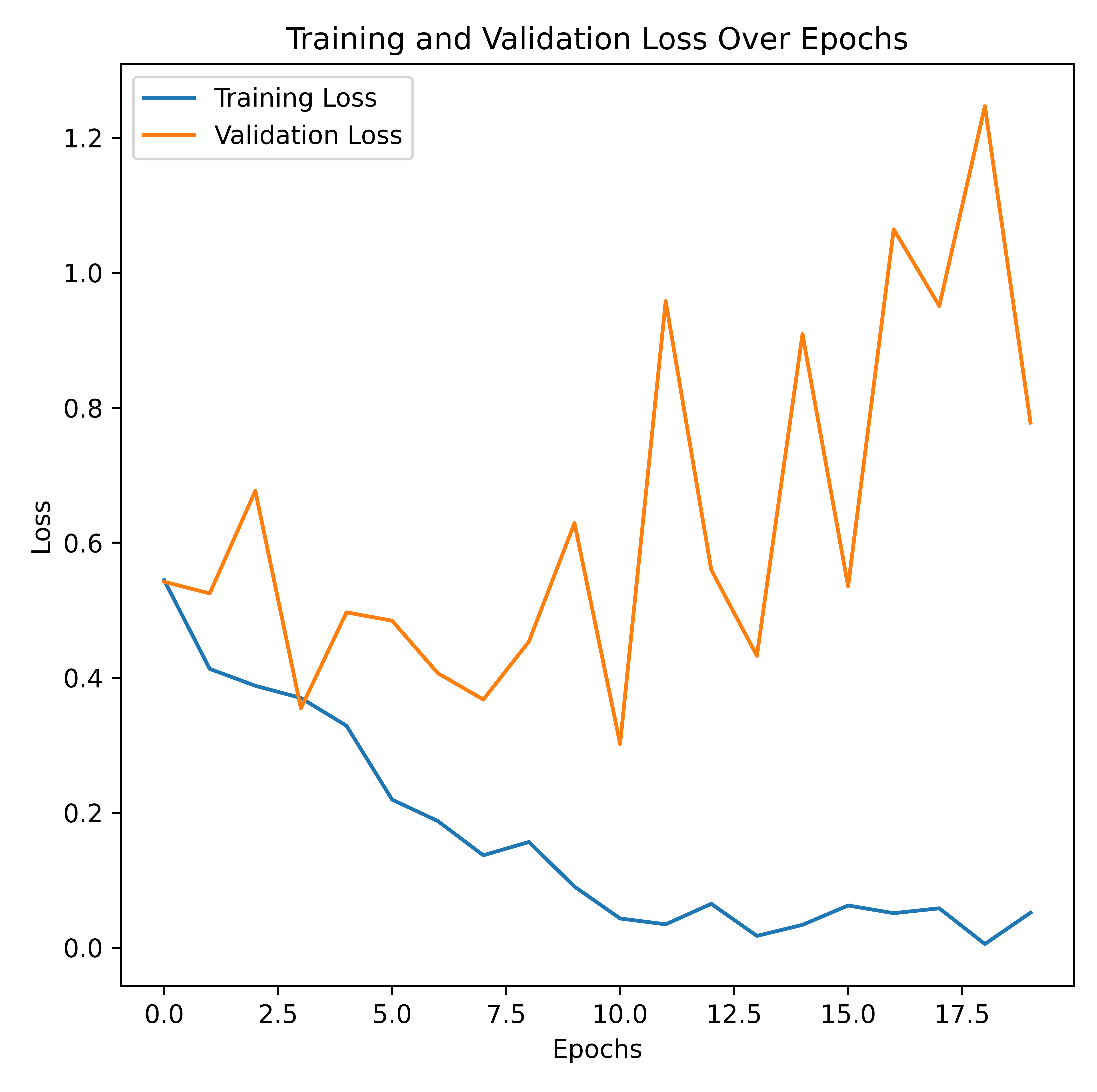}}
	\subfigure[\label{fig9f}]{\includegraphics[width=3cm,height=3cm]{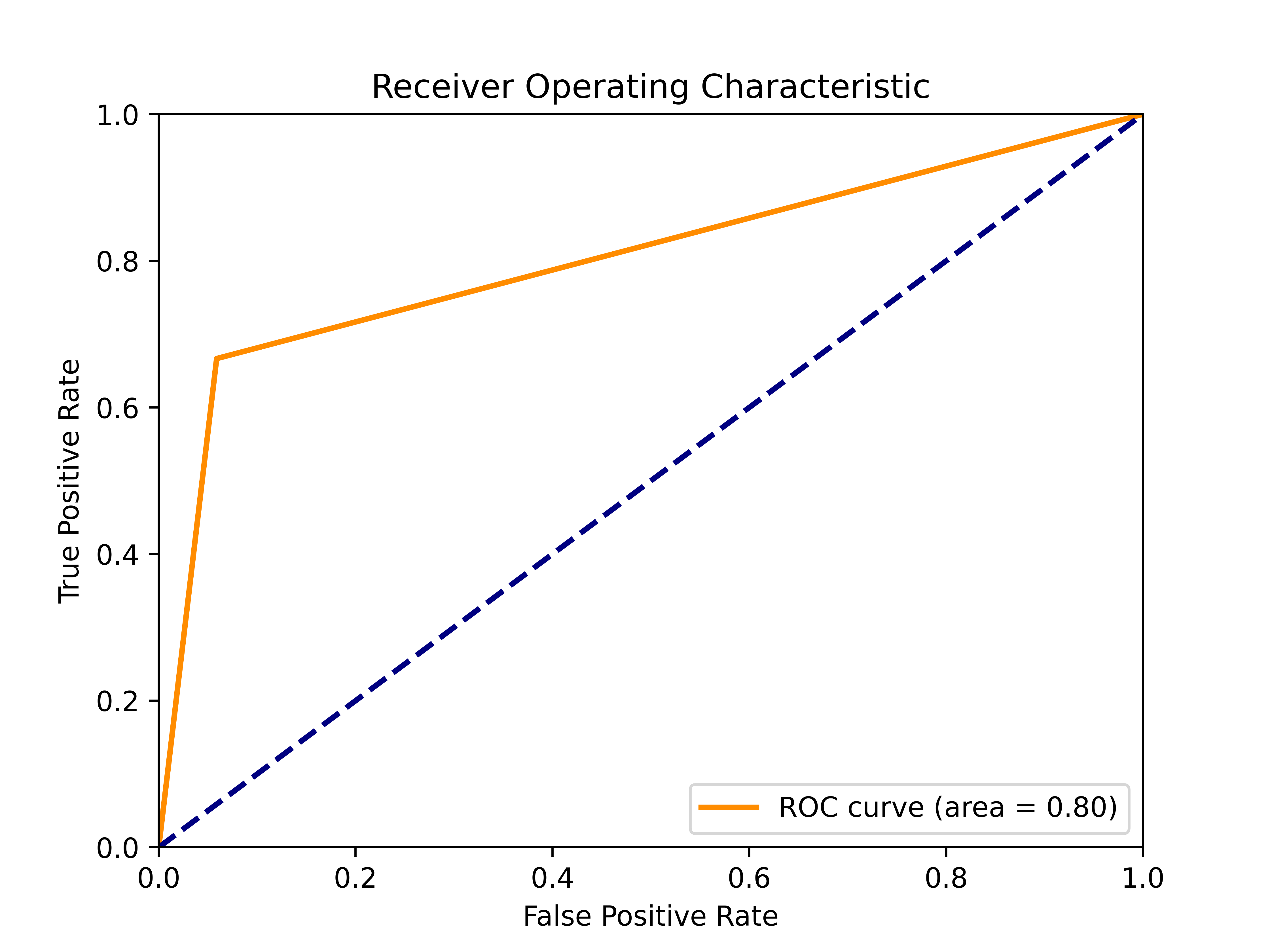}}\\
	
	\subfigure[\label{fig9g}]{\includegraphics[width=3cm,height=3cm]{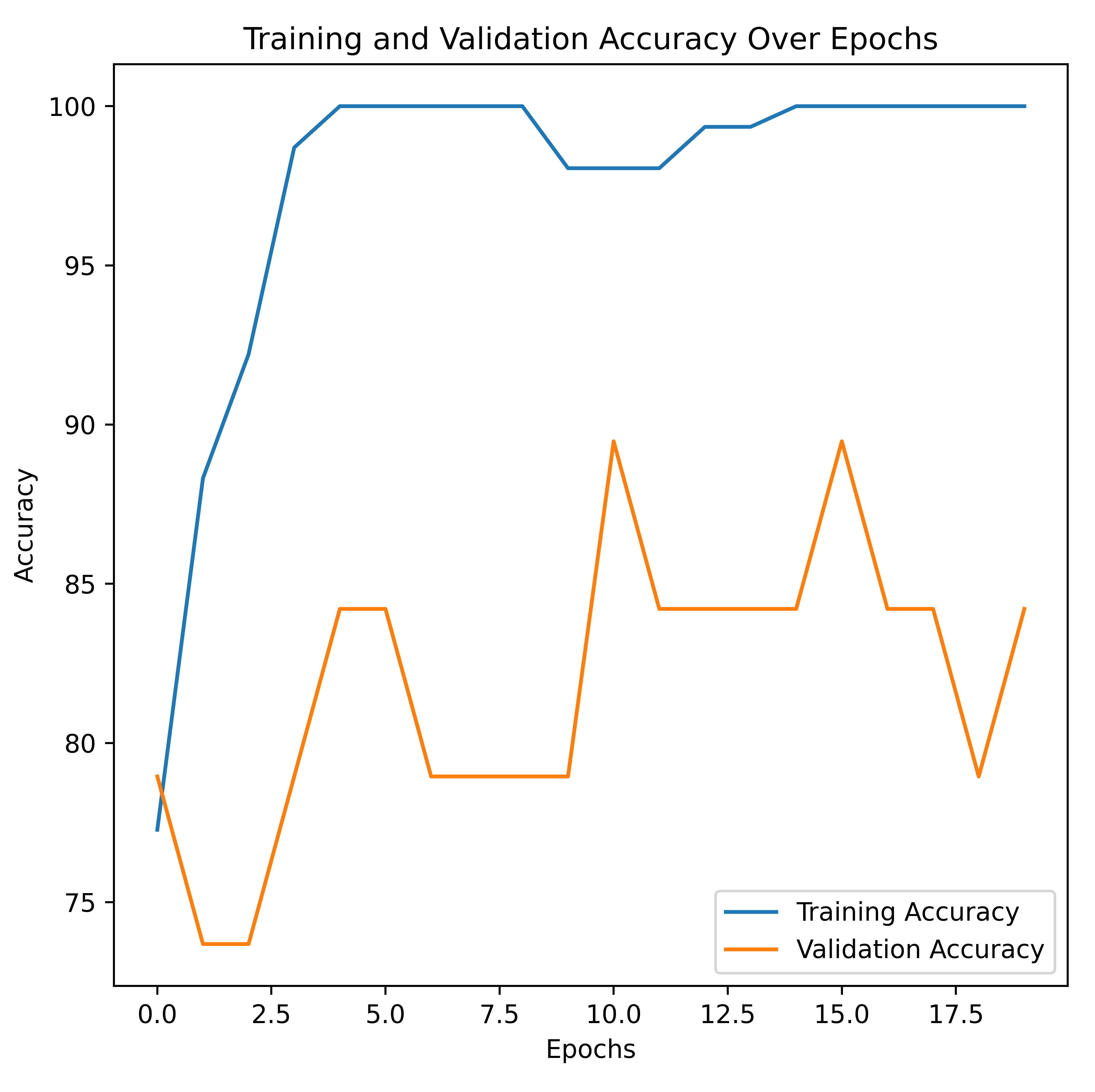}}
	\subfigure[\label{fig9h}]{\includegraphics[width=3cm,height=3cm]{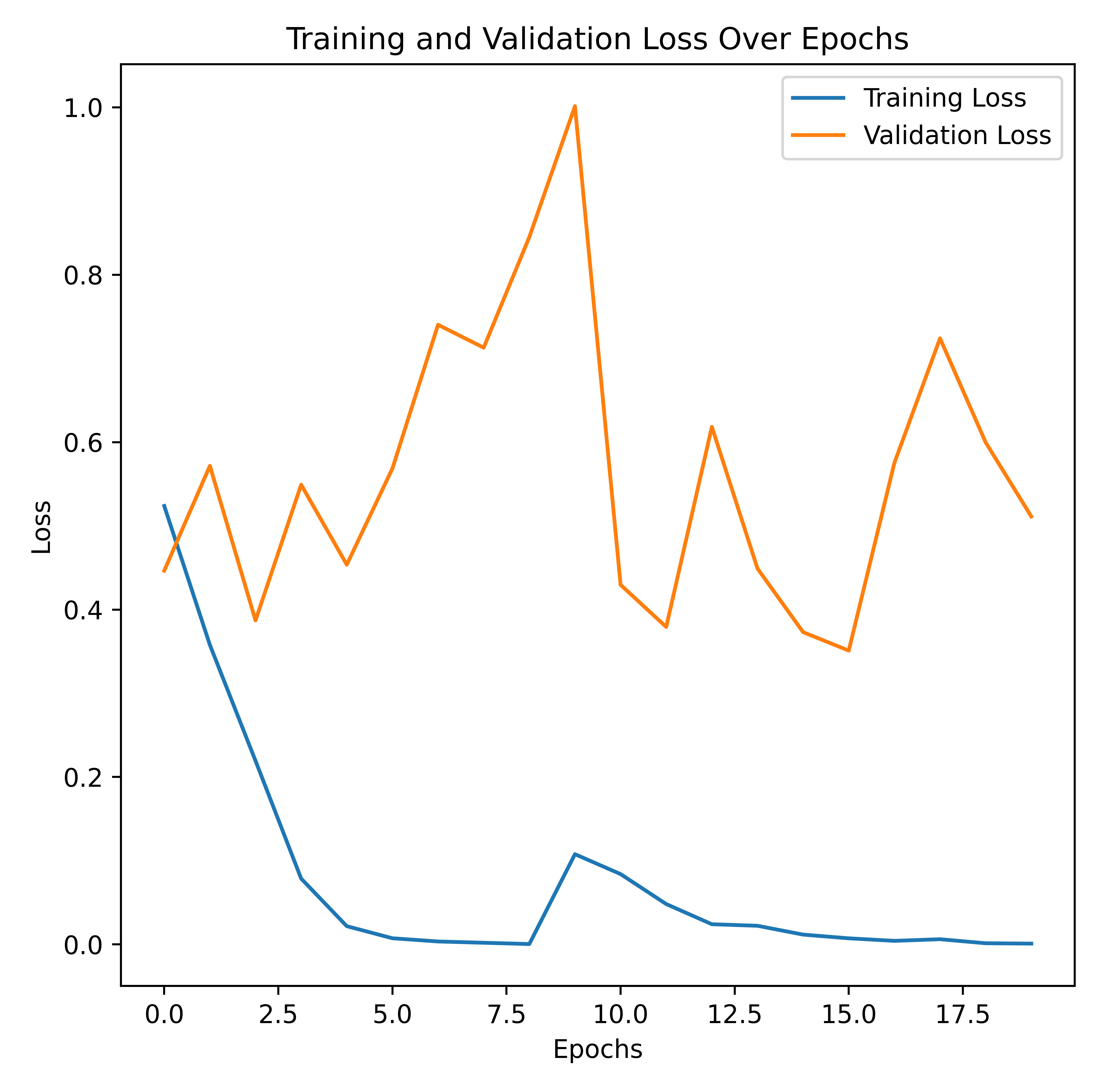}}
	\subfigure[\label{fig9i}]{\includegraphics[width=3cm,height=3cm]{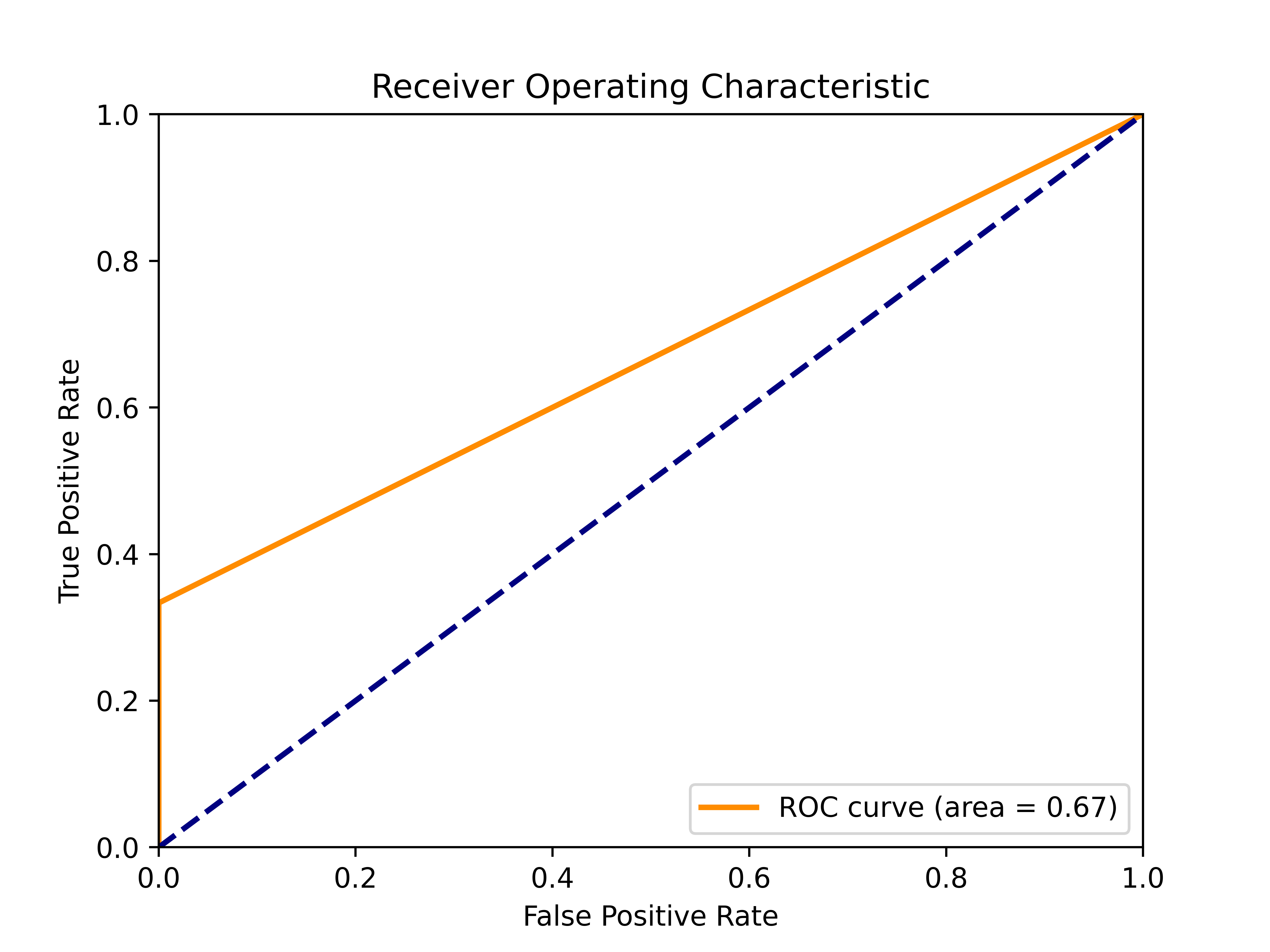}}
	
	\caption{The training and validation accuracy plot, loss plot and ROC (from left to right) for the Columbia dataset for the RGB, YCbCr and HSV color frames (top to bottom).}
	\label{fig9}
	
\end{figure}

The Columbia dataset poses considerable challenges, reflected in the fluctuating validation accuracy and suboptimal ROC performance. These issues stem from the reduced dataset size after removing corrupted images, leading to a less comprehensive representation of the data distribution. The smaller, imbalanced dataset impacts the model's ability to generalize effectively, resulting in inconsistent validation accuracy and difficulty distinguishing between classes. Among the evaluated color spaces, RGB outperforms others with an accuracy of 95\%, precision of 1.0, and an F1-score of 0.80, although its recall is relatively low at 0.67. The YCbCr and HSV color spaces achieve lower accuracies (both 90\%), with HSV showing the weakest recall (0.33) and F1-score (0.50). These results highlight the need for larger, more balanced datasets to enhance model performance and stability (refer to figure \ref{fig9} and Table \ref{table 6}).

\begin{table}[!htbp]
	\centering
	\caption{The evaluation parameters for the Columbia dataset for the different color spaces.}
	\label{table 6}
	\begin{tabular}{|l|c|c|c|c|c|}
		\hline
		\textbf{Color Space} & \textbf{Accuracy} &\textbf{Precision} &\textbf{Recall} & \textbf{F1-score} &\textbf{AUC}\\ \hline
		\textbf{RGB} &0.95 &1.00 &0.67 &0.80 &0.83\\
		\textbf{YCbCr} &0.90 &0.67 &0.67 &0.67 &0.80\\
		\textbf{HSV} &0.90 &1.00 &0.33 &0.50 &0.67\\
		\hline
	\end{tabular}
\end{table}

\subsubsection{Combined Dataset Performance}

\begin{figure}[!htbp]
	\centering
	\subfigure[\label{fig10a}]{\includegraphics[width=3cm,height=3cm]{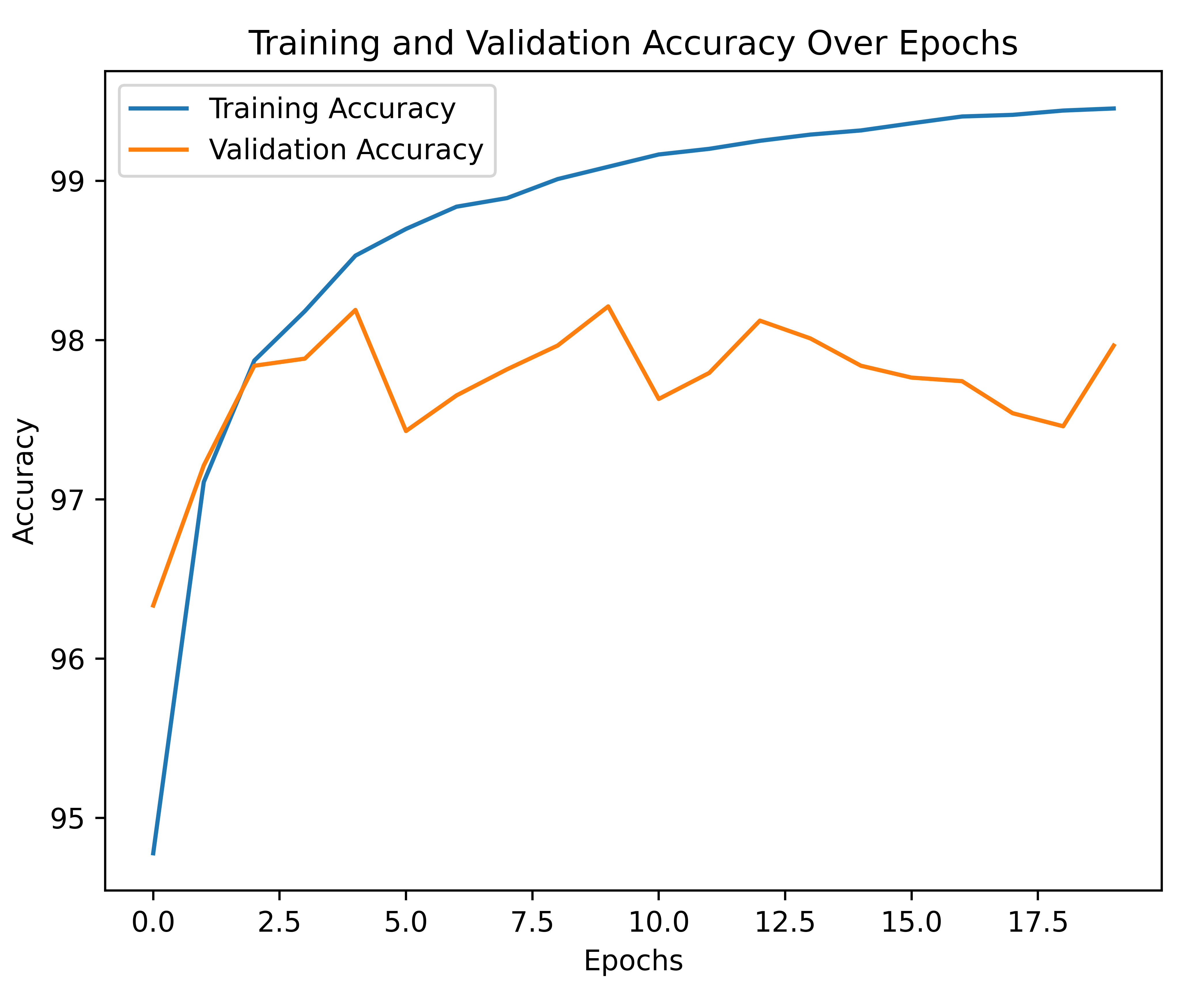}}
	\subfigure[\label{fig10b}]{\includegraphics[width=3cm,height=3cm]{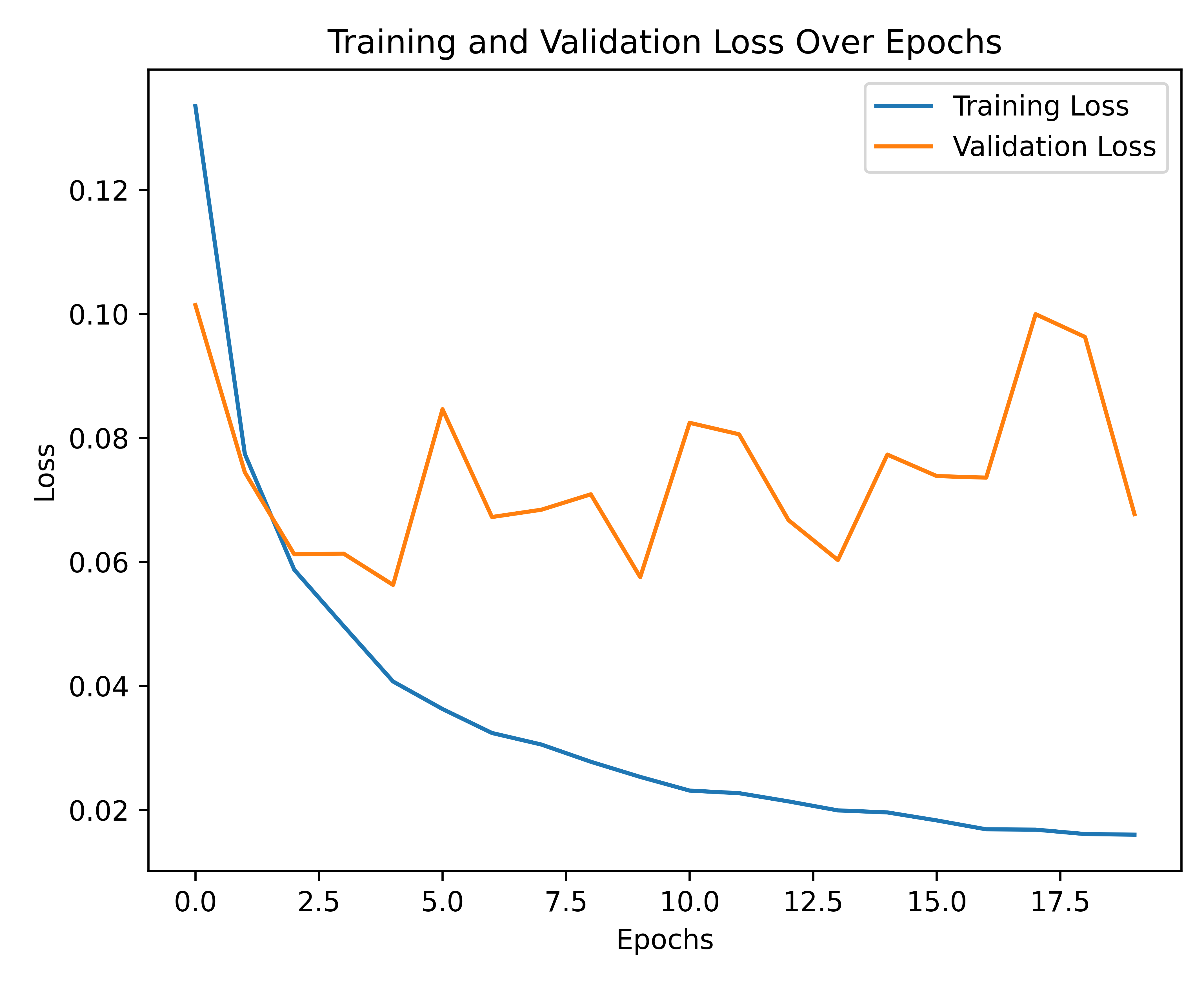}}
	\subfigure[\label{fig10c}]{\includegraphics[width=3cm,height=3cm]{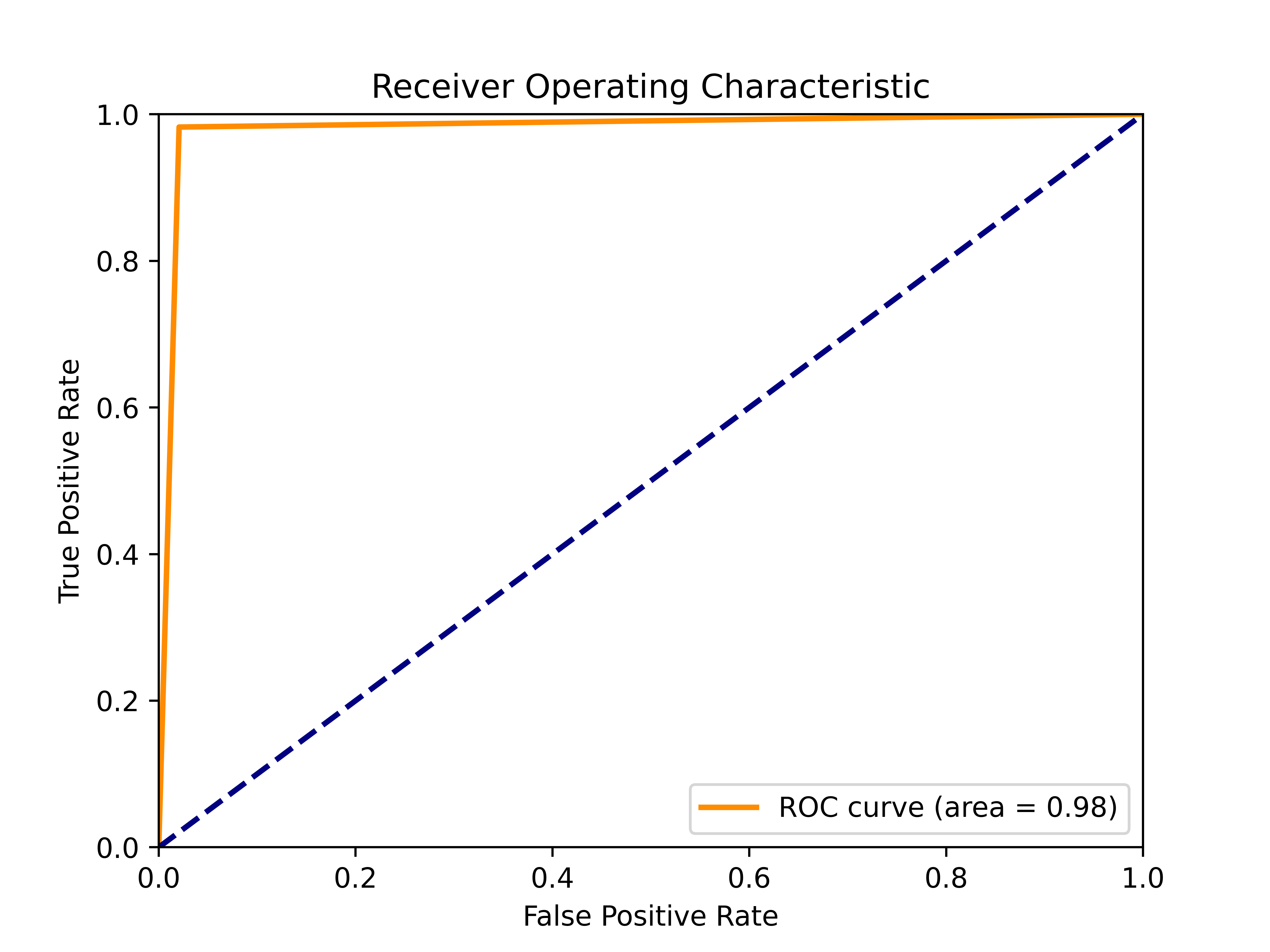}}\\
	
	\subfigure[\label{fig10d}]{\includegraphics[width=3cm,height=3cm]{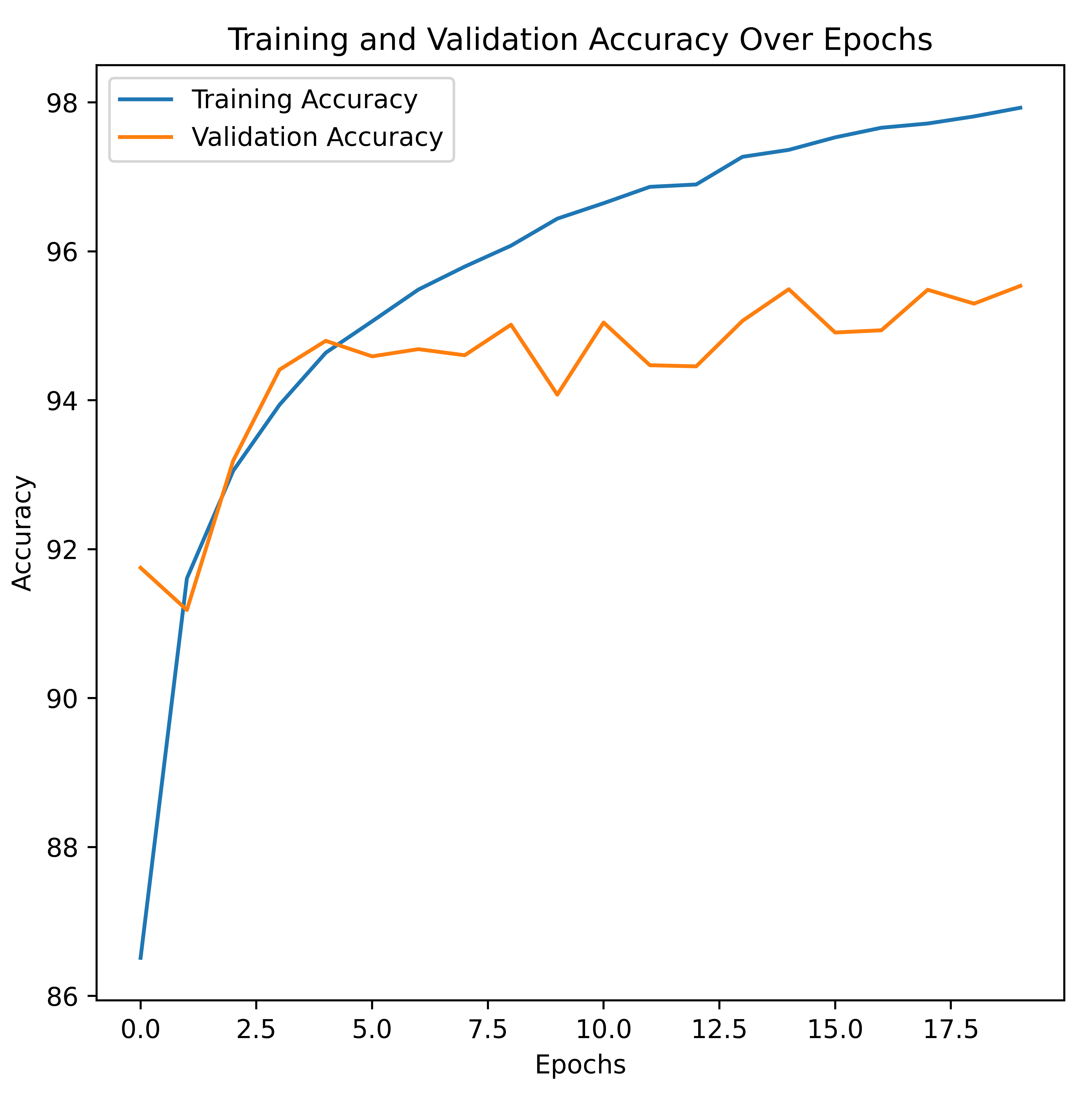}}
	\subfigure[\label{fig10e}]{\includegraphics[width=3cm,height=3cm]{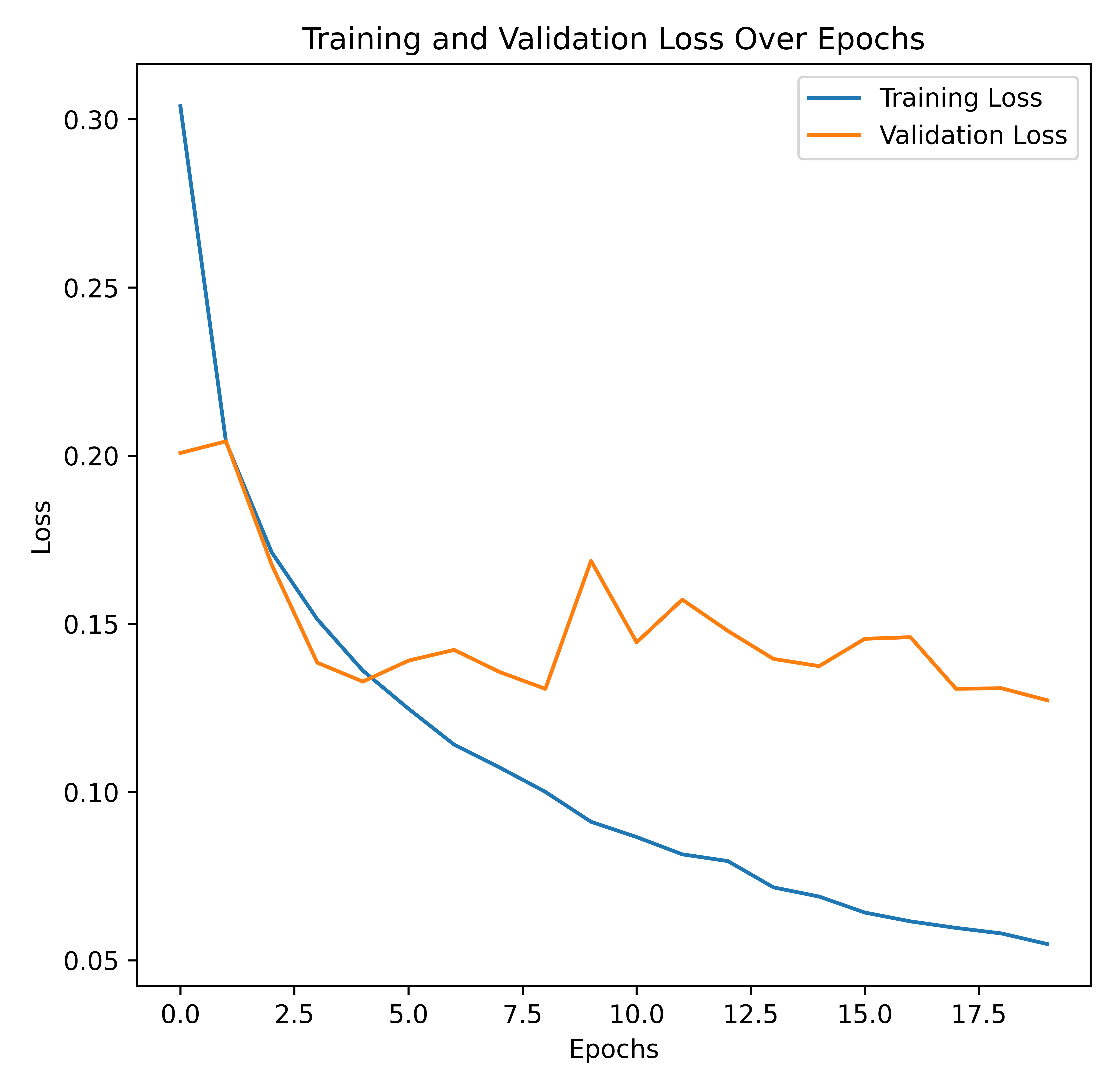}}
	\subfigure[\label{fig10f}]{\includegraphics[width=3cm,height=3cm]{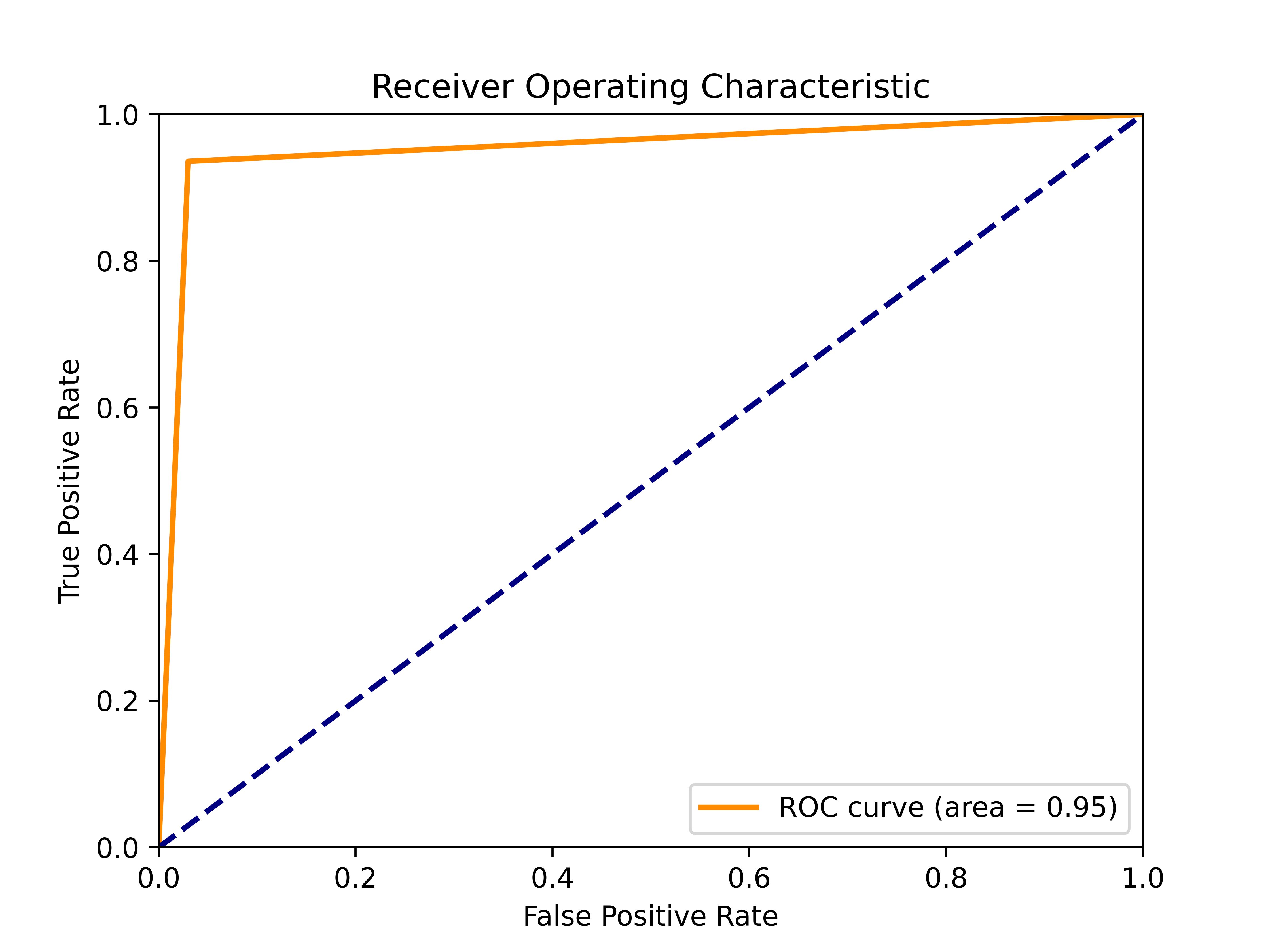}}\\
	
	\subfigure[\label{fig10g}]{\includegraphics[width=3cm,height=3cm]{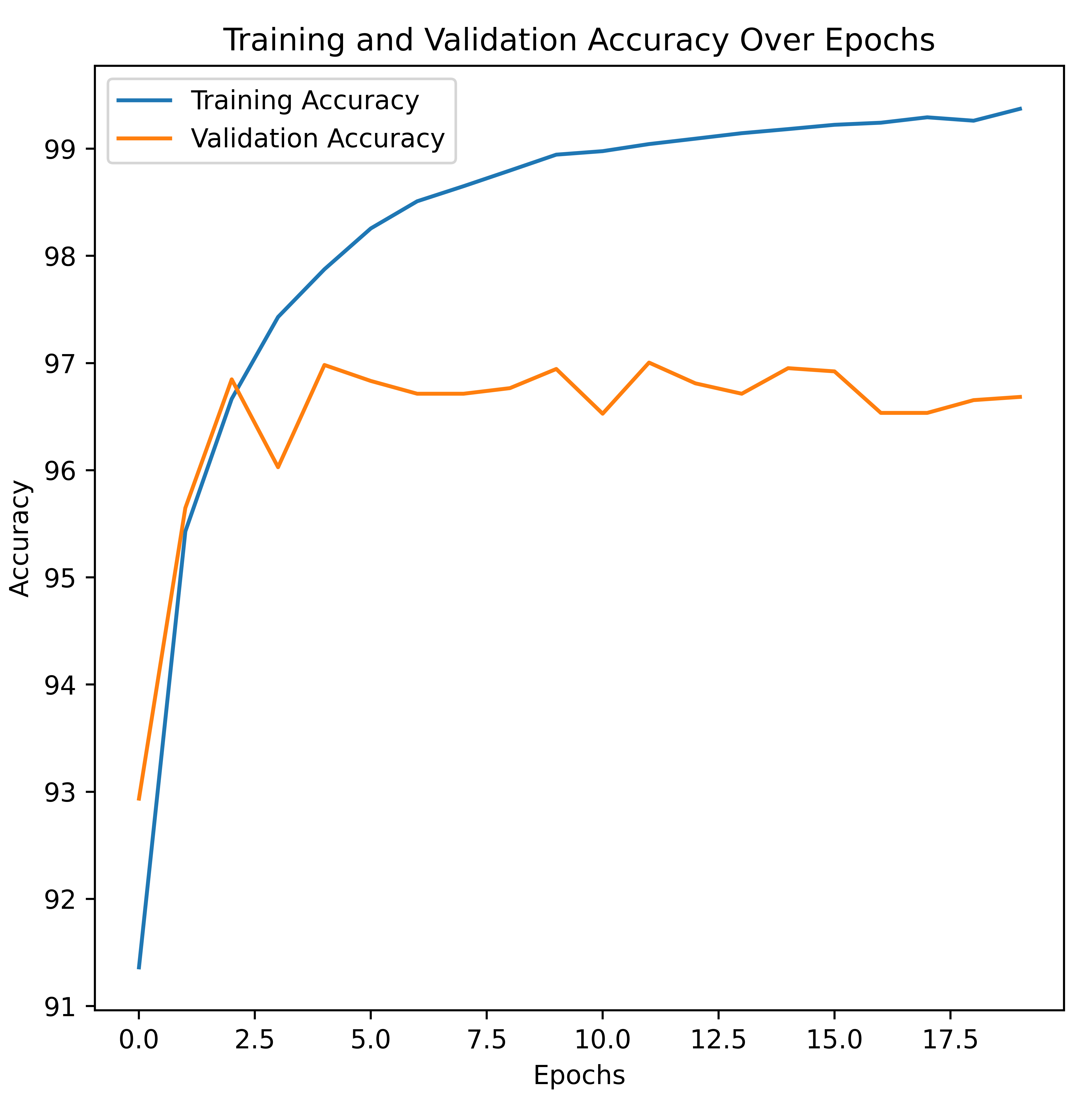}}
	\subfigure[\label{fig10h}]{\includegraphics[width=3cm,height=3cm]{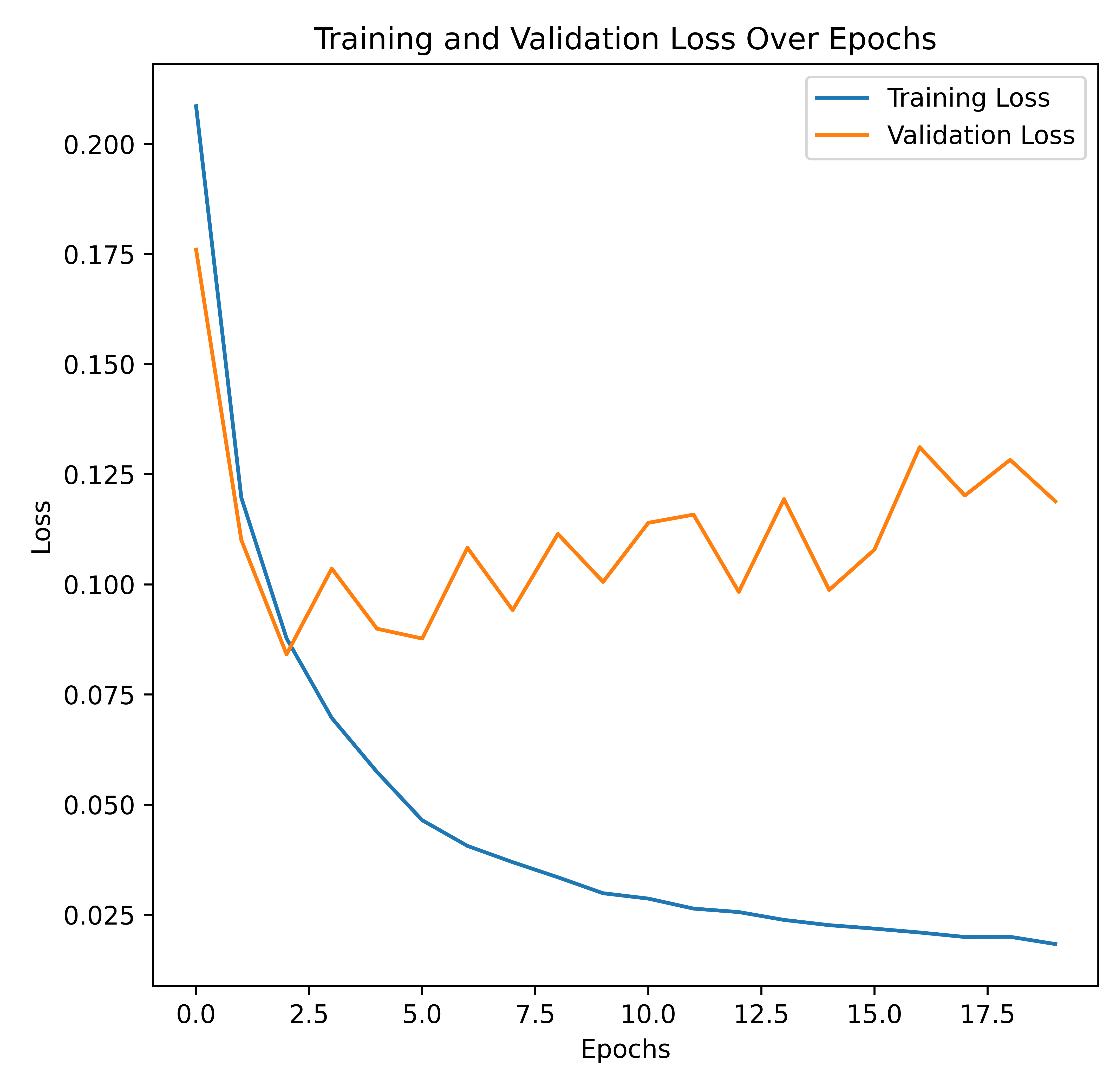}}
	\subfigure[\label{fig10i}]{\includegraphics[width=3cm,height=3cm]{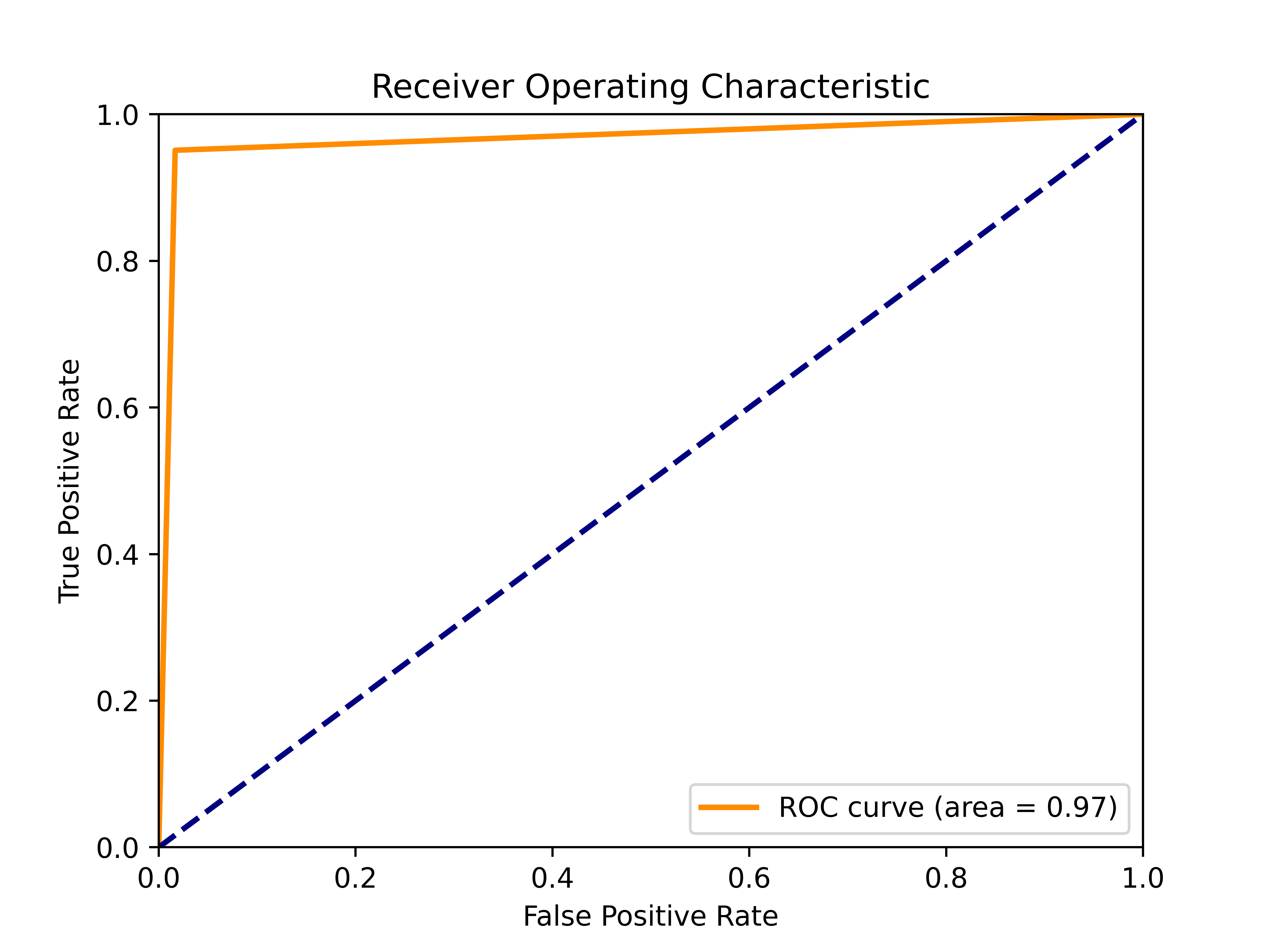}}
	
	\caption{The training and validation accuracy plot, loss plot and ROC (from left to right) for the Combined datasets (D1+D2+D3) for the RGB, YCbCr and HSV color frames (top to bottom).}
	\label{fig10}
	
\end{figure}

The evaluation of the combined dataset (D1+D2+D3) reveals that the model performs best in the RGB color space, achieving an accuracy, precision, recall, F1-score, and AUC of 0.98, indicating its robustness in handling diverse data. While the HSV color space also demonstrates competitive performance with an accuracy of 0.97, the YCbCr color space exhibits comparatively lower metrics, with an accuracy of 0.95, highlighting its limitations in feature representation (refer figs \ref{fig10} and table \ref{table 7}).

\begin{table}[!htbp]
	\centering
	\caption{The evaluation parameters for the Combined datasets (D1+D2+D3) for the different color spaces.}
	\label{table 7}
	\begin{tabular}{|l|c|c|c|c|c|}
		\hline
		\textbf{Color Space} & \textbf{Accuracy} &\textbf{Precision} &\textbf{Recall} & \textbf{F1-score} &\textbf{AUC}\\ \hline
		\textbf{RGB} &0.98 &0.98 &0.98 &0.98 &0.98\\
		\textbf{YCbCr} &0.95 &0.97 &0.94 &0.95 &0.95\\
		\textbf{HSV} &0.97 &0.98 &0.95 &0.97 &0.97\\
		\hline
	\end{tabular}
\end{table}

In conclusion, the Swin Transformer model demonstrates varying performance across the CiFAKE, Columbia, and JSSSTU datasets, with the Columbia dataset being the most challenging due to its reduced size and class imbalance. While the CiFAKE and JSSSTU datasets exhibit more stable training and promising ROC curves, the combined dataset (D1+D2+D3) highlights the complexities of generalizing across diverse data sources. Notably, the evaluation reveals the HSV color space as a competitive alternative, achieving high accuracy and efficiency, making it particularly suitable for applications requiring compact color representations. These findings emphasize the importance of dataset quality, balance, and appropriate color space selection in training effective machine learning models. Future work could explore advanced data augmentation, balancing techniques, and model tuning to further enhance performance across heterogeneous datasets, as reflected in the evaluation metrics summarized in Table \ref{table 7}.

\begin{table}[!htbp]
	\centering
	\caption{The comparative analysis of combined datasets(D1+D2+D3) for the RGB color space with other CNN models. }
	\label{table 8}
	\begin{tabular}{|l|c|c|c|c|c|}
		\hline
		\textbf{Models} & \textbf{Accuracy} &\textbf{Precision} &\textbf{Recall} & \textbf{F1-score} &\textbf{AUC}\\ \hline
		{VGG-19} &0.96 &0.97 &0.95 &0.96 &0.96\\
		{ResNet-50} &0.98 &0.97 &0.97 &0.97 &0.98\\
		\textbf{Proposed method} &0.98 &0.98 &0.98 &0.98 &0.98\\
		\hline
	\end{tabular}
\end{table}

The comparative analysis shown in Table  \ref{table 8} of the combined datasets (D1 + D2+ D3) across various CNN models, including VGG-19, ResNet-50, and the proposed Swin Transformer-based approach, reveals key insights. While VGG-19 delivers competitive performance with an accuracy of 96\% and an AUC of 0.96, ResNet-50 demonstrates a slight improvement with an accuracy of 98\% and an AUC of 0.98. The proposed method matches the performance of ResNet-50 in terms of accuracy, precision, recall, F1-score, and AUC, achieving 98\% across all metrics, indicating its robustness and efficiency.

\section{Conclusion and Future Work}
\label{conclusion}

The experiments conducted on the CiFAKE, Columbia, and JSSSTU datasets using the Swin Transformer model and an in-depth analysis of different color spaces revealed critical insights into the model's capacity to differentiate CGI from authentic images. The RGB color space yielded consistently high performance on the CiFAKE and JSSSTU datasets, achieving superior accuracy, precision, recall, and F1 scores, demonstrating the model's robustness in controlled environments. However, the Columbia dataset, characterized by limited and imbalanced data, posed significant challenges, resulting in lower recall and AUC values, reflecting the model's difficulty in generalizing under these constraints.
A comparative analysis of color spaces revealed that HSV offered competitive results, closely aligning with the performance of RGB, especially for applications prioritizing compact and discriminative representations of color information. This highlights the model's sensitivity to color space transformations and suggests HSV as a promising alternative for future studies. The combined dataset (D1+D2+D3) further emphasized the challenges of domain generalization, as integrating datasets with diverse distributions and characteristics exposed the complexities of achieving consistent performance across all evaluation metrics.

Future research should leverage these findings by incorporating multi-modal approaches that blend features from multiple color spaces, such as RGB and HSV, alongside texture and frequency-domain features. Transfer learning from large, diverse datasets and advanced balancing techniques for addressing data imbalances should also be explored to enhance model generalization. Combined with further analysis of alternative color spaces, these efforts could provide more comprehensive solutions for distinguishing CGI from authentic images across varied datasets.

\section*{Acknowledgment}
The article is an extended version of the article titled ``Swin Transformer for Robust Differentiation of Real and Synthetic Images: Intra- and Inter-Dataset Analysis" accepted in 9th International Conference on Computer Vision \& Image Processing, 2024 held on 19-21 December, 2024 at IIITDM Kancheepuram, Chennai - 600127, India.

%
%
%
%

\end{document}